%% file: paper.tex
\documentclass[sigconf]{aamas}

\usepackage{balance} %

\setcopyright{ifaamas}
\acmConference[AAMAS '21]{Proc.\@ of the 20th International Conference on Autonomous Agents and multi-agent Systems (AAMAS 2021)}{May 3--7, 2021}{Online}{U.~Endriss, A.~Now\'{e}, F.~Dignum, A.~Lomuscio (eds.)}
\copyrightyear{2021}
\acmYear{2021}
\acmDOI{}
\acmPrice{}
\acmISBN{}

\acmSubmissionID{573}

\input{math_commands.tex}

\usepackage{url}

\usepackage{amsthm}
\usepackage{multirow}
\usepackage{xargs}
\usepackage{subcaption}

\usepackage{lipsum}

\usepackage[inline]{enumitem}

\newtheorem{theorem}{Theorem}

\newtheorem{lemma}{Lemma}

\newtheorem{proposition}{Proposition}

\newcommand{\norm}[1]{\left \| #1 \right \|}
\newcommand{\abs}[1]{\left | #1 \right |}

\DeclareMathOperator{\var}{Var}

\newcommand{\joint}[1]{{\bm{#1}}}
\newcommand{\expect}[2]{ \E_{#1} \! \left[ #2 \right]}
\newcommand*{\tran}{^{\mkern-1.5mu\mathsf{T}}}

\title{Contrasting Centralized and Decentralized Critics in Multi-Agent Reinforcement Learning}

\author{Xueguang Lyu}
\affiliation{}
\email{lu.xue@northeastern.edu}

\author{Yuchen Xiao}
\affiliation{}
\email{xiao.yuch@northeastern.edu}

\author{Brett Daley}
\affiliation{}
\email{b.daley@northeastern.edu}

\author{Christopher Amato}
\affiliation{}
\email{c.amato@northeastern.edu}

\makeatletter
\def\blfootnote{\xdef\@thefnmark{}\@footnotetext}
\makeatother

\begin{abstract}

\blfootnote{Khoury College of Computer Sciences, Northeastern University. Boston, MA, USA.}

\textit{Centralized Training for Decentralized Execution},
where agents are trained offline using centralized information but execute in a decentralized manner online, has gained popularity in the multi-agent reinforcement learning community.
In particular, actor-critic methods with a centralized critic and decentralized actors are a common instance of this idea.
However, the implications of using a centralized critic in this context are not fully discussed and understood even though it is the standard choice of many algorithms.
We therefore formally analyze centralized and decentralized critic approaches,
providing a deeper understanding of the implications of critic choice.
Because our theory makes unrealistic assumptions, we also empirically compare the centralized and decentralized critic methods over a wide set of environments to validate our theories and to provide practical advice.
We show that there exist misconceptions regarding centralized critics in the current literature and show that the centralized critic design is not strictly beneficial, but rather both centralized and decentralized critics have different pros and cons that should be taken into account by algorithm designers.
\end{abstract}

\newcommand{\BibTeX}{\rm B\kern-.05em{\sc i\kern-.025em b}\kern-.08em\TeX}

\begin{document}

\pagestyle{fancy}
\fancyhead{}

\maketitle

\vspace{-0.1cm}
\section{Introduction}

Centralized Training for Decentralized Execution (CTDE), where agents are trained offline using centralized information but execute in a decentralized manner online, has seen widespread adoption in multi-agent reinforcement learning (MARL)~\cite{foerster2016learning, oliehoek2008optimal, jorge2016learning}.
In particular, actor-critic methods with centralized critics have become popular after being proposed by ~\citet{foerster2018counterfactual} and~\citet{lowe2017multi}, since the critic can be discarded once the individual actors are trained.
Despite the popularity of centralized critics, the choice is not discussed extensively
and its implications for learning remain largely unknown.

One reason for this lack of analysis is that recent state-of-the-art works built on top of a centralized critic focus on other issues such as multi-agent credit assignment~\cite{foerster2018counterfactual, SQDDPG}, multi-agent exploration~\cite{LIIR}, teaching~\cite{LeCTR} or emergent tool use~\cite{openAI}.
However, state-of-the-art methods with a centralized critic do not compare with decentralized critic versions of their methods.
Therefore, without precise theory or tests, previous works relied on intuitions and educated guesses.
For example, one of the pioneering works on centralized critics, MADDPG~\cite{lowe2017multi},
mentions that providing more information to critics eases learning and makes learning coordinated behavior easier.
Later works echoed similar viewpoints, suspecting that a centralized critic might speed up training~\cite{CM3},
or that it reduces variance~\cite{TarMAC}, is more robust~\cite{simoes2020multi}, improves performance~\cite{lee2019improved} or stabilizes training~\cite{M3DDPG}.

In short, previous works generally give the impression that a centralized critic is an obvious choice without compromises under CTDE, and there are mainly two advertised benefits:
\begin {enumerate*} [label=\itshape\alph*\upshape)]
  \item a centralized critic fosters ``cooperative behavior'',
  \item a centralized critic also stabilizes (or speeds up) training.
\end {enumerate*}
It makes intuitive sense because %
training a global value function on its own (i.e., joint learning~\cite{claus1998dynamics}) would help with cooperation issues and has much better convergence guarantees due to the stationary learning targets.
However, these intuitions have never been formally proven or empirically tested;
since most related works focus on additional improvements on top of a centralized critic,
the centralized critic is usually seen as part of a basic framework rather than an optional hyperparameter choice.\footnote{
    Methods are also typically implemented with state-based critics ~\cite{foerster2018counterfactual, CM3, SQDDPG, LIIR} instead of the history-based critics we use in this paper,
    which might be another reason for performance differences. However, we only consider history-based critics to fairly compare the use of centralized and decentralized critics with the same type of information.
} %
In this paper, we look into these unvalidated claims and point out that common intuitions turn out to be inaccurate.

First, we show theoretically that a centralized critic does not necessarily improve cooperation compared to a set of decentralized critics.
We prove that the two types of critics provide the decentralized policies with precisely the same gradients in expectation.
We validate this theory on classical cooperation games and more realistic domains and report results supporting our theory.

Second, we show theoretically that the centralized critic results in higher variance updates of the decentralized actors assuming converged on-policy value functions.
Therefore, we emphasize that stability of \textit{value function learning} does not directly translate to a reduced variance in \textit{policy learning}.
We also discuss that, in practice, this results in a bias-variance trade-off. 
We analyze straightforward examples and empirical evaluations, confirming our theory and showing that the centralized critic often makes the policy learning less stable, contrary to the common intuition. 

Finally, we test standard implementations of the methods over a wide range of popular domains and discuss our empirical findings where decentralized critics often outperform a centralized critic.
We further analyze the results and discuss possible reasons for these performance differences.
We therefore demonstrate room for improvement with current methods while laying theoretical groundwork for future work.

\section{Related Work}

Recent deep MARL works often use the CTDE training paradigm. 
Value function based CTDE approaches~\cite{QTRAN,MAVEN,ROMA, WeightedQMIX,COMIXandFacMADDPG, xiao_icra_2020, NDQ,QMIX,VDN} focus on how centrally learned value functions can be reasonably decoupled into decentralized ones and have shown promising results.
Policy gradient methods on CTDE, on the other hand, have heavily relied on centralized critics.
One of the first works utilizing a centralized critic was COMA~\cite{foerster2018counterfactual},
a framework adopting a centralized critic with a counterfactual baseline.
For convergence properties, COMA establishes that the overall effect on decentralized policy gradient with a centralized critic can be reduced to a single-agent actor-critic approach, which ensures convergence under similar assumptions~\cite{konda2000actor}.
In this paper, we take the theory one step further and show convergence properties for centralized and decentralized critics as well as their respective policies, while giving a detailed bias-variance analysis.

Concurrently with COMA, MADDPG~\cite{lowe2017multi} proposed to use a dedicated centralized critic for each agent in semi-competitive domains, demonstrating compelling empirical results in continuous action environments.
M3DDPG~\cite{M3DDPG} focuses on the competitive case and extends MADDPG to learn robust policies against altering adversarial policies by optimizing a minimax objective.
On the cooperative side, SQDDPG~\citep{SQDDPG} borrows the counterfactual baseline idea from COMA and extends MADDPG to achieve credit assignment in fully cooperative domains by reasoning over each agent's marginal contribution. 
Other researchers also use critic centralization for emergent communication with decentralized execution in TarMAC~\cite{TarMAC} and ATOC~\cite{ATOC}.
There are also efforts utilizing an attention mechanism addressing scalability problems in MAAC~\cite{MAAC}.
Besides, teacher-student style transfer learning LeCTR~\cite{LeCTR} also builds on top of centralized critics, which does not assume expert teachers. 
Other focuses include multi-agent exploration and credit assignment in LIIR~\cite{LIIR}, goal-conditioned policies with CM3~\cite{CM3}, and for temporally abstracted policies~\cite{chakravorty2020option}.
Based on a centralized critic, extensive tests on a more realistic environment using self-play for  hide-and-seek~\cite{openAI} have demonstrated impressive results showing emergent tool use.
However, as mentioned before, these works use centralized critics, but none of them specifically investigate the effectiveness of centralized critics, which is the main focus of this paper.

\section{Background}

This section introduces the formal problem definition of cooperative MARL with decentralized execution and partial observability. 
We also introduce the single-agent actor-critic method and its most straight-forward multi-agent extensions.

\subsection{Dec-POMDPs}

A decentralized partially observable Markov decision process (Dec-POMDP) is an extension of an MDP in decentralized multi-agent settings with partial observability~\cite{Book16}.
A Dec-POMDP is formally defined by the tuple
\(\langle \mathcal{I}, \mathcal{S}, \{\mathcal{A}_i\}, \mathcal{T}, \mathcal{R},
\{\Omega_i\}, \mathcal{O}  \rangle\), in which 

\begin{itemize}
    \item \(\mathcal{I}\) is the set of agents,
    \item \(\mathcal{S}\) is the set of states including initial state \(s_0\),
    \item \(\mathbf{A} = \times_i \mathcal{A}_i\) is the set of joint actions,
    \item \(\mathcal{T}: \mathcal{S} \times \mathbf{A} \rightarrow \mathcal{S}\) is the transition dynamics,
    \item \(\mathcal{R}: \mathcal{S} \times \mathbf{A} \times \mathcal{S} \rightarrow \mathbb{R}\) is the reward function,
    \item \(\mathbf{\Omega} = \times_i\Omega_i\) is the set of observations for each agent,
    \item \(\mathcal{O}: \mathcal{S} \times \mathbf{A} \rightarrow \mathbf{\Omega}\) is the observation probabilities.
\end{itemize}
At each timestep \(t\), a joint action \(\bm{a}
= \langle a_{1,t}, ..., a_{|\mathcal{I}|,t} \rangle\) 
is taken, each agent
receives its corresponding local observation \(\langle o_{1,t},\dots,o_{|\mathcal{I}|,t} \rangle \sim \mathcal{O}(\bm{s}_t,\bm{a}_t)\) 
and a global reward \(r_t = \mathcal{R}(\mathbf{s}_t,\bm{a}_t,
\mathbf{s}_{t+1})\). 
The joint expected discounted return is
\(G = \sum_{t=1}^T\gamma^t r_t\) where \(\gamma\) is a discount factor.
Agent $i$'s \textit{action-observation history} for timestep $t$ is defined as $\langle o_{i,1},a_{i,1},o_{i,2},\dots,a_{i,t-1},o_{i,t}\rangle$, and we define history recursively as $h_{i,t} = \langle h_{i,t-1}, a_{i,t-1}, o_{i,t}\rangle$,
likewise, a joint history is $\bm{h}_t=\langle \bm{h}_{t-1},\bm{a}_{t-1},\bm{o}_t\rangle$.\footnote{
    For Proposition~\ref{prop:ss_history} and all following results, we employ a fixed-memory history where the history only consists of the past $k$ observations and actions.
    Formally, a history at timestep $t$ with memory length $k$ is defined as $\bm{h}_{t,k}=\langle\bm{h}_{t-1,k-1},\bm{a}_{t-1},\bm{o}_t\rangle$ when $k>0$ and $t>0$, otherwise $\varnothing$.
}
To solve a Dec-POMDP is to find a set of  policies $\bm{\pi} = \langle\pi_1,\dots,\pi_{|\mathcal{I}|}\rangle$ where
$\pi_i: h_{i,t} \rightarrow a_{i,t}$ such that the joint expected discounted  return $G$ is maximized.

\paragraph{Notation}
For notational readability, we denote the parameterized decentralized policy 
\(\pi_{\theta_i}\) as \(\pi_i\), on-policy $G$ estimate \(Q^\pi_i\) as \(Q_i\), and $Q^{\bm{\pi}}$ as $Q$.
We denote the objective (discounted expected return) for any agent with a
decentralized critic as \(J_d\) and the objective with a centralized critic as \(J_c\).
In addition, a timestep \(t\) is implied for \(s\), \(\bm{h}\), $h_i$, $\bm{a}$ or \(a_i\).

\subsection{Actor Critic method (AC)}

Actor critic (AC)~\cite{konda2000actor} is a widely used policy gradient (PG) architecture and is the basis for many single-agent policy gradient approaches.
Directly optimizing a policy,
policy gradient (PG) algorithms perturb policy parameters $\theta$ in the direction of the gradient of the expected return $\nabla_{\theta}\E[G_{\pi_\theta}]$,
which is conveniently given by the \textit{policy gradient theorem}~\cite{sutton2000policy,konda2000actor}:
\begin{equation}\label{eq:regular_ac_grad}
    \nabla_{\theta}\E[G_{\pi_\theta}] 
    = \E_{h,a} [\nabla_\theta \log\pi_\theta(a \mid h)Q^{\pi_\theta}(h,a)]
\end{equation}
Actor-critic (AC) methods~\citep{konda2000actor} directly implement the policy gradient theorem by learning the value function $Q^{\pi_\theta}$, i.e., the critic, commonly through TD learning~\citep{sutton1985temporal}.
The policy gradient for updating the policy (i.e., the actor) then follows the return estimates given by the critic (Equation~\ref{eq:regular_ac_grad}).

\subsection{Multi-Agent Actor Critic methods}
We introduce three extensions of single-agent Actor Critic methods to multi-agent settings,
which are highlighted in Table~\ref{table:ac_method_naming}.

The first AC multi-agent extension, Joint Actor Critic (JAC)~\cite{bono2018cooperative,wang2019achieving},
treats the multi-agent environment as a single-agent environment and learns in the joint observation-action space;
JAC learns a centralized actor, $\bm{\pi}(\bm{a}\mid\bm{h};\theta)$, and a centralized value function (critic), $Q^{\bm{\pi}}(\bm{h},\bm{a};\phi)$.
The policy gradient for JAC follows that of single-agent actor critic:
\begin{equation}
    \nabla J(\theta) = \E_{\bm{a},\bm{h} } [\nabla \log 
    \bm{\pi}(\bm{a}\mid \bm{h};\theta)
    Q^{\bm{\pi}}(\bm{h},\bm{a};\phi)].
\end{equation}

The second AC multi-agent extension, called Independent Actor Critic
(IAC)~\citep{tan1993multi,foerster2018counterfactual}, learns a
decentralized policy and critic \(\langle\pi_i(a_i \mid h_i;\theta_i), Q_i(h_i,a_i; \phi_i)\rangle\) for each of the agents locally. 
At every timestep $t$, a local experience $\langle h_{i,t},a_{i,t}\rangle$ is generated for agent $i$.
The policy gradient learning for agent $i$ is defined as
\begin{equation}
 \label{decentral_critic_policy_update}
   \nabla_{\theta_i} J_d(\theta_i) = \E_{\bm{a},\bm{h}} [\nabla \log \pi_i(a_i \mid h_i;\theta_i)Q_i(h_i,a_i;\phi_i)].
\end{equation}

Finally, we define Independent Actor with Centralized
Critic (IACC), a class of centralized critic methods where a joint value function
\(Q^{\bm{\pi}}(\bm{h},\bm{a};\phi)\)
is used to update each decentralized policy \(\pi_{\theta_i}\)~\cite{foerster2018counterfactual,bono2018cooperative}.\footnote{
    For example, COMA~\citep{foerster2018counterfactual} is then considered as an IACC approach with a variance reduction baseline.
}
Naturally, the policy gradient for decentralized policies with a centralized critic is defined as
\begin{equation}
 \label{central_critic_policy_update}
   \nabla_{\theta_i} J_c(\theta_i) = \E_{\bm{a},\bm{h}} [\nabla \log \pi_i(a_i \mid h_i;\theta_i)Q^{\bm{\pi}}(\bm{h},\bm{a};\phi)].
\end{equation}
At any timestep, the joint expected return estimate $Q^{\bm{\pi}}(\bm{h},\bm{a};\phi)$ is used to update the decentralized policy $\pi(a_i \mid h_i;\theta_i)$.
Notice that the centralized critic $Q^{\bm{\pi}}(\bm{h},\bm{a};\phi)$ estimates the return based on joint information (all agents' action-histories) that differs from the decentralized case in Eq.~\ref{decentral_critic_policy_update}.
In the following section, we shall show that from a local viewpoint of agent $i$, for each joint action-history $\langle\bm{h}_t,\bm{a}_t\rangle$,
$Q^{\bm{\pi}}(\bm{h}_t,\bm{a}_t;\phi)$ is a sample from the return distribution given local action-histories $\Pr(G_{t:T} \mid h_{i,t},a_{i,t})$,
while the decentralized critic $Q_i(h_{i,t},a_{i,t})$ provides an expectation.

\begin{table}[t]
   \centering
   \renewcommand{\arraystretch}{1.2} %
   \begin{tabular}{l|l l}
    Method  & Critic & Actor \\
    \hline
    JAC~\cite{bono2018cooperative,wang2019achieving}  & Centralized  & Centralized \\
    IAC~\cite{tan1993multi,foerster2018counterfactual} & Decentralized  & Decentralized \\
    IACC~\cite{foerster2018counterfactual,lowe2017multi} & Centralized & Decentralized \\
   \end{tabular}
   \vspace{0.3cm}
   \caption{Our Multi-agent Actor Critic Naming Scheme.}
   \vspace{-0.6cm}
   \label{table:ac_method_naming}
\end{table}

\section{Bias Analysis}
\label{sec:bias_analysis}
In this section, we prove policies have the same expected gradient whether using centralized or decentralized critics.
We prove that the centralized critic provides unbiased and correct on-policy return estimates,
but at the same time makes the agents suffer from the same action shadowing problem~\cite{matignon27independent} seen in decentralized learning.
It is reassuring that the centralized critic will not encourage a decentralized policy to pursue a joint policy that is only achievable in a centralized manner, but also calls into question the benefits of a centralized critic.
We provide a theoretical proof on bias equivalence, then analyze a classic example for intuitive understanding.

\subsection{Equivalence in Expected Gradient}
\label{nobias_subsec}

We first show the gradient updates for IAC and IACC are the same in expectation.
We assume the existence of a limiting distribution $\Pr(\joint{h})$ over fixed-length histories, analogous to a steady-state assumption.
That is, we treat fixed-memory trajectories as nodes in a Markov chain for which the policies induce a stationary distribution.
\begin{proposition}
    \label{prop:ss_history}
    Suppose that agent histories are truncated to an arbitrary but finite length.
    A stationary distribution on the set of all possible histories of this length is guaranteed to exist under any collection of agent policies.
\end{proposition}
We provide formal justification for Proposition~\ref{prop:ss_history} in Appendix~\ref{appendix:ss_history},
necessary to derive Bellman equations in our following theoretical results.
With this, we begin by establishing novel convergence results for the centralized and decentralized critics in Lemmas~\ref{lemma:central_convergence} and~\ref{lemma:decentral_convergence} below, which culminate in Theorem~\ref{theorem:unbiased_gradient} regarding the expected policy gradient.\footnote{
    While it appears that we analyze the case with two agents, agent \(i\) and \(j\),
    the result holds for arbitrarily many agents by letting \(j\) represent all agents except $i$.
}
\begin{lemma}
    \label{lemma:central_convergence}
    Given the existence of a steady-state history distribution (Proposition~\ref{prop:ss_history}),
    training of the \textbf{centralized} critic is characterized by the Bellman operator $B_c$ which admits a unique fixed point
    $Q^\joint{\pi} (h_i,h_j,a_i,a_j)$ where $Q^\joint{\pi}$ is the true expected return under the joint policy $\joint{\pi}$.
\end{lemma}
\begin{lemma}
    \label{lemma:decentral_convergence}
    Given the existence of a steady-state history distribution (Proposition~\ref{prop:ss_history}),
    training of the $i$-th \textbf{decentralized} critic is characterized by a Bellman operator $B_d$ which admits a unique fixed point
    $\expect{h_j,a_j}{Q^{\bm{\pi}} (h_i,h_j,a_i,a_j)}$ where $Q^\joint{\pi}$ is the true expected return under the joint policy $\joint{\pi}$.
\end{lemma}
\begin{theorem}
    \label{theorem:unbiased_gradient}
    After convergence of the critics' value functions, the expected policy gradients from the centralized critic and the decentralized critics are equal.
    That is,
    \begin{equation}
        \expect{}{\nabla_\theta J_c(\theta)} = \expect{}{\nabla_{\theta} J_d(\theta)}
    \end{equation}
    where $J_c$ and $J_d$ are the respective objective functions for the actors with central and decentral critics, and the expectation is taken over all joint histories and joint actions.
    All actors are assumed to have the same policy parameterization.
\end{theorem}
\paragraph{Proof sketch:}
We derive Bellman equations for the centralized and decentralized critics and express them as Q-function operators.
We show that these operators are contraction mappings and admit fixed points at
$Q^\joint{\pi} (h_i,h_j,a_i,a_j)$
and
$\expect{h_j,a_j}{Q^\joint{\pi} (h_i,h_j,a_i,a_j)}$,
respectively.
These convergence results reveal that the decentralized critic becomes the marginal expectation of the centralized critic after training for an infinite amount of time.
Under the total expectation over joint histories and joint actions, these fixed points are identically equal, implying that gradients computed for the centralized and decentralized actors are the same in expectation and therefore unbiased.
The full proofs for Lemmas~\ref{lemma:central_convergence} and \ref{lemma:decentral_convergence} and Theorem~\ref{theorem:unbiased_gradient}
are respectively provided in Appendices~\ref{appendix:central_convergence},~\ref{appendix:decentral_convergence}, and~\ref{appendix:unbiased_gradient}.

Our theory assumes that the critics are trained sufficiently to converge to their true on-policy values.
This assumption often exists in the form of infinitesimal step sizes for the actors~\citep{konda2000actor, singh2000nash, zhang2010multi,bowling2001convergence,foerster2018counterfactual} for convergence arguments of AC, since the critics are on-policy return estimates and the actors need an unbiased and up-to-date critic.
Although this assumption is in line with previous theoretical works, it is nevertheless unrealistic;
we discuss the practical implications of relaxing this assumption in Section~\ref{sec:value_function_learning}.

\subsection{Climb Game as an Intuitive Example}
We use a classic matrix game as an example to intuitively highlight that IAC and IACC give the same policy gradient in expectation.
The Climb Game~\cite{claus1998dynamics}, whose reward function is shown in Table~\ref{climb_game_return},
is a matrix game (a state-less game) in which agents are supposed to cooperatively try to achieve the highest reward of \(11\) by taking the joint action \(\langle u_1,u_1 \rangle\), facing the risk of being punished by \(-30\) or \(0\) when agents miscoordinate.
It is difficult for independent learners to converge onto the optimal \(\langle u_1,u_1 \rangle\) actions due to low expected return for agent 1 to take \(u_1\) when agent 2's policy is not already favoring \(u_1\) and vise versa.

This cooperation issue arises when some potentially good action $a$ has low ("shadowed") on-policy values because yielding a high return depends on other agents' cooperating policies,
but frequently taking action $a$ is essential for other agents to learn to adjust their policies accordingly, creating a dilemma where agents are unwilling to frequently take the low-value action and are therefore stuck in a local optimum.
In the case of the Climb Game, the value of $u_1$ is often shadowed, because $u_1$ does not produce a satisfactory return unless the other agent also takes $u_1$ frequently enough.
This commonly occurring multi-agent local optimum is called a \textit{shadowed equilibrium}~\citep{fulda2007predicting, panait2008theoretical}, a known difficulty in independent learning which usually requires an additional cooperative mechanism (e.g., some form of centralization) to overcome.

Solving the Climb Game independently with IAC, assume the agents start with uniformly random policies. In expectation, IAC's decentralized critic would estimate $Q_1(u_1)$ at $(1/3)\times11+(1/3)\times(-30)+0\approx -6.3$, making $u_3$ (with $Q_1(u_3)\approx3.7$) a much more attractive action, and the same applies for agent 2.
Naturally, both agents will update towards favoring $u_3$;
continuing this path, agents would never favor $u_1$ and never discover the optimal value of $u_1$: $Q^*_i(u_1)=11$.

In the case of an IACC method, unsurprisingly, the centralized critic can quickly learn the correct values, including the value of the optimal joint action
\(Q(u_1,u_1) = 11\) since there is no environmental stochasticity.
However, consider at timestep $t$ agent 1 takes optimal action \(u_1\), the (centralized) $Q$ value estimate used in 
policy gradient $\nabla Q(u_1,a_{2,t}) \pi_1(u_1)$ actually depends on what action agent 2 chooses to take according to its policy $a_{2,t} \sim \pi_2$.
Again assuming uniform policies,
consider a rollout where action $a_{2,t}$ is sampled from the policy of agent 2; then with sufficient sampling, we expect the mean policy gradient (given by centralized critic) for updating $\pi_1(u_1)$ would be
\begin{equation}
\begin{aligned}
    & \E_{\pi_2}\nabla J_C(\theta\mid a_{1,t}=u_1) \\
    & = \pi_2(u_1)\nabla Q(u_1,u_1) \pi_1(u_1) + \pi_2(u_2)\nabla Q(u_1,u_2) \pi_1(u_1)
    \\&\quad+ \pi_2(u_3)\nabla Q(u_1,u_3) \pi_1(u_1)   \\
    & = (1/3)\nabla 11 \pi_1(u_1) +  (1/3)\nabla -30 \pi_1(u_1) + (1/3)\nabla 0 \pi_1(u_1) \\
    & \approx -6.3\cdot \nabla \pi_1(u_1)
\end{aligned}
\end{equation}
That is, with probabilities \(\pi_2(u_1)\), \(\pi_2(u_2)\) and \(\pi_2(u_3)\),
the joint Q-values \(11\),\(-30\) and \(0\) are sampled for generating the policy gradient of action 1 for agent 1, $\pi_1(u_1)$.
It implies that the sample-average gradient for \(\pi_1(u_1)\) is high only when agent 2 takes action $u_1$ frequently.
If agent 2 has a uniform policy, the sample-average gradient $-6.3\nabla\pi_1(u_1)$ cannot compete with the gradient for $u_3$ at $1.7\nabla\pi_1(u_3)$.
Therefore, in the IACC case with a centralized critic, we see the rise of an almost identical action shadowing problem we described for IAC, even though the centralized critic trained jointly and has the correct estimate of the optimal joint action.

\begin{figure}
  \centering
  \includegraphics[width=0.3\textwidth]{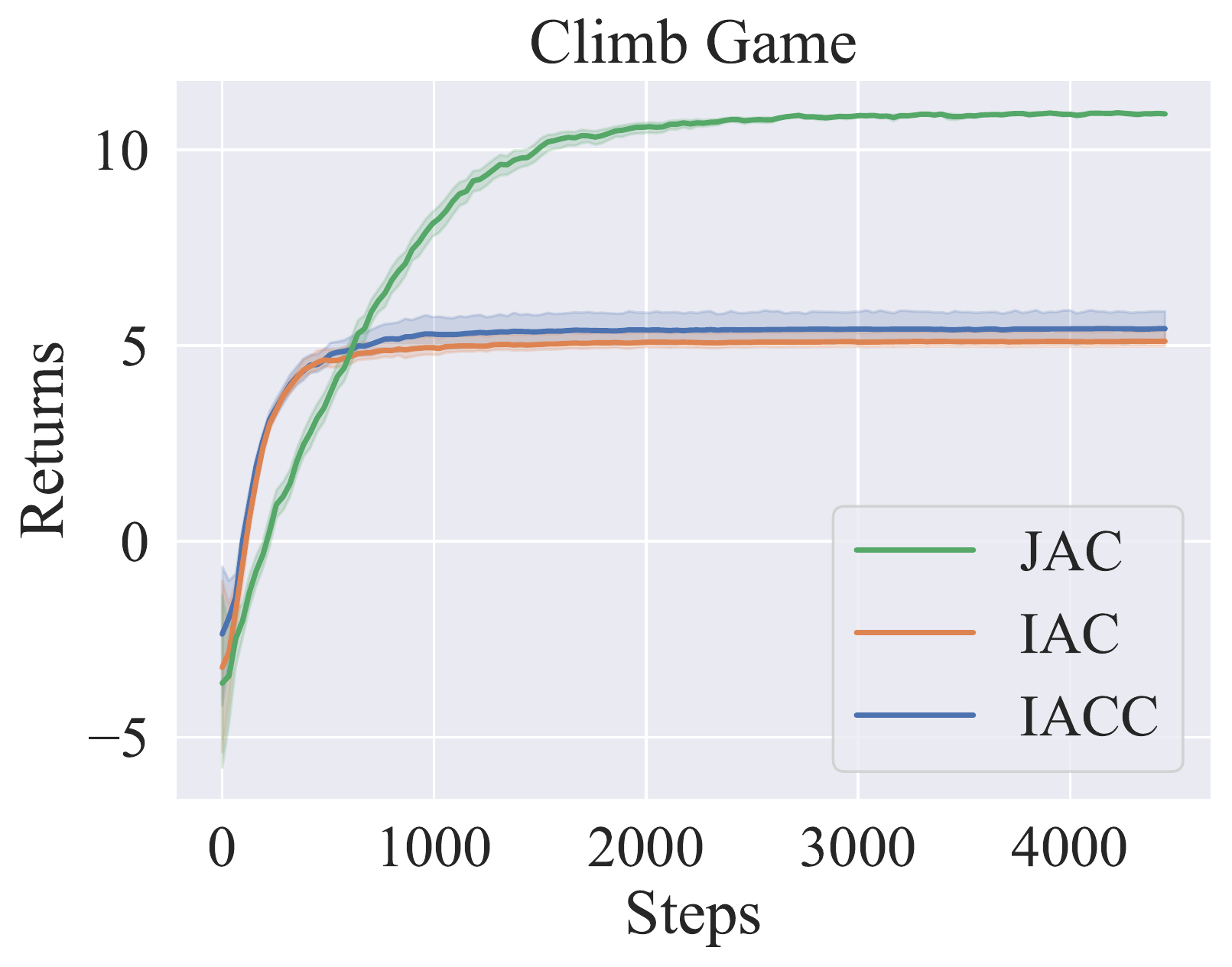}
  \vspace{-0.2cm}
  \caption{Climb Game empirical results (50 runs per method) showing both decentralized and centralized critic methods succumb to the \textit{shadowed equilibrium} problem.}
  \label{climb_game1}
  \vspace{-0.4cm}
\end{figure}

Empirical evaluation on the Climb Game (shown in Figure~\ref{climb_game1})
conforms to our analysis, showing both methods converge to the suboptimal
solution $\langle a_3, a_3 \rangle$. At the same time, unsurprisingly, a centralized controller always gives the optimal solution $\langle a_1,a_1\rangle$.
In general, we observe that the centralized critic has the information of the optimal solution, information that is only obtainable in a centralized fashion and is valuable for agents to break out of their cooperative local optima.
However, this information is unable to be effectively utilized on the individual actors to form a cooperative policy.
Therefore, contrary to the common intuition, in its current form, the centralized critic is unable to foster cooperative behavior more easily than the decentralized critics.

\begin{table}[t]
  \vspace{.1in}
  \centering
  \renewcommand{\arraystretch}{1.3} %
  \begin{tabular}{l l|l l l}
    & & \multicolumn{3}{c}{agent 1} \\
    & & \(u_1\) & \(u_2\) & \(u_3\) \\
    \hline

    \multirow{3}{*}{agent 2} &

      \(u_1\) & 11  & -30 & 0 \\
    & \(u_2\) & -30 & 7   & 6 \\
    & \(u_3\) & 0   & 0   & 5 \\

  \end{tabular}
  \vspace{.1in}
  \caption{Return values for Climb Game~\citep{claus1998dynamics}.}
  \vspace{-0.4cm}
  \label{climb_game_return}
\end{table}

\section{Variance Analysis}
\label{sec:variance_analysis}
In this section, we first show that with true on-policy value functions, the centralized critic formulation can increase policy gradient variance.
More precisely, we prove that the policy gradient variance using a centralized critic is at least as large as the policy gradient variance with decentralized critics. 
We again assume that the critics have converged under fixed policies, thus ignoring the variance due to value function learning;
we discuss the relaxation of this assumption in Section~\ref{sec:discussion}.

We begin by comparing the policy gradient variance between centralized and decentralized critics:
\begin{theorem}
    \label{theorem:variance}
    Assume that all agents have the same policy parameterization.
    After convergence of the value functions, the variance of the policy gradient using a centralized critic is at least as large as that of a decentralized critic along every dimension.
\end{theorem}
\paragraph{Proof sketch:}
As with our proof of Theorem~\ref{theorem:unbiased_gradient}, Lemmas~\ref{lemma:central_convergence} and~\ref{lemma:decentral_convergence} show that a decentralized critic's value estimate is equal to the marginal expectation of the central critic's $Q$-function after convergence.
This implies that the decentralized critics have already averaged out the randomness caused by the other agents' decisions.
Since each agent has the same policy parameterization, their policy gradient covariance matrices are equal up to the scale factors induced by the critics' value estimates.
By Jensen's inequality, we show that the additional stochasticity of the central critic can increase (but not decrease) these scale factors compared to a decentralized critic;
hence, $\Var(J_c(\theta)) \geq \Var(J_d(\theta))$ element-wise.

In the following subsections, we define and analyze this variance increase by examining its two independent sources:
the ``Multi-Action Variance'' (MAV) induced by the other actors' policies,
and the ``Multi-Observation Variance'' (MOV) induced by uncertainty regarding the other agents' histories from the local perspective of a single agent.
We introduce these concepts along with concrete examples to illustrate how they affect learning.

\subsection{Multi-Action Variance (MAV)}
\label{subsec:mav}
We discuss MAV in the fully observable case and will address the partially observable case in MOV.
Intuitively, from the local perspective of agent $i$, when taking an action
$a^i$ at state $s$, MAV is the variance $\var[G(s,a_i)]$ in return estimates due to the fact that teammates might take different actions according to their own stochastic policies.
With decentralized critics, MAV is averaged into the value
function (Lemma~\ref{lemma:decentral_convergence}) which is an expectation incorporating teammates actions $a_j$; thus, at given timestep $t$, $Q_i(s,a_i)$ has no variance.
On the other hand, a centralized critic $Q(s,a_i,a_j)$ distinguishes between all action combinations $\langle a_i,a_j \rangle$ (Lemma~\ref{lemma:central_convergence}), but $a_j$ is sampled by agent $j$ during execution: $a_j\sim \pi_j(s)$;
therefore, the value of $Q(s,a_i,a_j)$ varies during policy updates depending on $a_j$.
We propose a simple domain to clarify this variance's cause and effect and show that a centralized critic transfers MAV directly to policy updates.

\subsubsection{The Morning Game and the MAV}
The Morning Game, inspired by~\citet{peshkin2000learning}, shown in Table~\ref{breakfast_game_return}, consists of two agents collaborating on making breakfast in which the most desired combination is
\(\langle cereal, milk \rangle\).
Since there is no environmental stochasticity, a centralized critic can robustly learn all the values correctly after only a few samples. In contrast, the decentralized critics need to average over unobserved teammate actions.
Take a closer look at action $cereal$: for a centralized critic, cereal with milk returns 3 and cereal with vodka returns 0; 
meanwhile, a decentralized critic receives stochastic targets (3 or 0) for taking action $cereal$, and only when agent 2 favors $milk$, a return of 3 comes more often, which then would make $Q_1(cereal)$ a higher estimate.
Therefore, the centralized critic has a lower variance (zero in this case), and the decentralized critic has a large variance on the update target.

Often neglected is that using a centralized critic has higher variance when it comes to updating decentralized policies.
Suppose agents employ uniform random policies for both IAC and IACC,
in which case agent 1's local expected return for $cereal$ would be $Q_1(cereal) = \pi_2(milk)\cdot 3 + \pi_2(vodka)\cdot 0 = 1.5$.
Assuming converged value functions, then in IACC, a centralized critic would uniformly give either $Q(cereal,milk)=3$ or $Q(cereal,vodka)=0$ for $\pi_1(cereal)$ updates, and $Q(pickles,milk)=0$ or $Q(pickles,vodka)=1$ for $\pi_1(pickles)$ updates.
With IAC, a decentralized critic always gives $Q(cereal)=1.5$ for $\pi_1(cereal)$, and $Q(pickle)=0.5$ for $\pi_2(pickle)$.
Obviously, under both methods, $\pi_1$ converges towards $cereal$, but the decentralized critic makes the update direction less variable and much more deterministic in favor of $cereal$.

\begin{table}[t]
   \centering
   \renewcommand{\arraystretch}{1.3} %
   \begin{tabular}{l l|l l}
    & & \multicolumn{2}{c}{agent 1} \\
    & & \(pickles\) & \(cereal\)  \\
    \hline
    \multirow{2}{*}{agent 2} &
      \(vodka\) & 1 & 0  \\
    & \(milk\) & 0 & 3    \\
   \end{tabular}
   \vspace{0.1in}
   \caption{Return values for the proposed Morning Game. }
   \label{breakfast_game_return}
   \vspace{-0.6cm}
\end{table}

\subsection{Multi-Observation Variance (MOV)}
\label{subsec:mov}
In the partially observable case, another source of variance in local value $\var[G(h_i,a_i)]$ comes from factored observations.
More concretely, for an agent in a particular local trajectory $h_i$, other agents'
experiences $h_j\in \mathcal{H}_j$ may vary, over which the decentralized agent has to average.
A decentralized critic is designed to average over this observation variance
and provide and single expected value for each local trajectory $Q_i(h_i,a_i)$.
The centralized critic, on the other hand, is able to distinguish each combination of trajectories $\langle h_i,h_j \rangle$, but when used for a decentralized policy at $h_i$, teammate history $h_j$ can be considered to be sampled from $\Pr(h_j \mid h_i)$,
and we expect the mean estimated return during the update process to be $\E_{h_j}Q(h_i,h_j,\bm{a})$.

We use a thought experiment\footnote{
    We also propose a toy domain called Guess Game based on this thought experiment, which we elaborate and show empirical results in Appendix~\ref{guess_game_environment}.
}
as an example.
Consider a one-step task where two agents have binary actions and are individually rewarded $r$ if the action matches the other’s randomly given binary observations and $-r$ for a mismatch; that is, $R(h_i,h_j,a_i)=r$ when $h_j=a_i$ and $-r$ otherwise.
With any policy, assuming converged value functions, a decentralized critic would estimate $Q(h_i,a_i)=0$ with zero variance.
On the other hand, a centralized critic with a global scope is able to recognize whether the current situation $\langle h_i,h_j,a_i \rangle$ would result in a return of 1 or 0, hence estimates $r$ with probability $0.5$ (when $h_j=a_i$ by definition) and $-r$ with probability $0.5$ (when $h_j\neq a_i$),  resulting in a variance of $r^2$.
In this example, we see that a centralized critic produces returns estimates with more significant variance when agents have varying observations.

\section{Discussion}\label{sec:discussion}
In this section, we discuss the anticipated trade-off in practice. We look at the practical aspects of a centralized critic and address the unrealistic true value function assumption.
We note that, although both types of critics have the same expected gradient for policy updates, they have different amount of actual bias in practice.
We also discuss how and why the way of handling variance is different for IAC and IACC. 
We conclude that we do not expect one method to dominate the other in terms of  performance in general.

\subsection{Value Function Learning}\label{sec:value_function_learning}
So far, in terms of theory, we have only considered value functions that are assumed to be correct.
However, in practice, value functions are difficult to learn.
We argue that it is generally more so with decentralized value functions.

In MARL, the on-policy value function is non-stationary since the return distributions heavily depend on the current joint policy.
When policies update and change, the value function is partially obsolete and is biased towards the historical policy-induced return distribution.
Bootstrapping from outdated values creates bias. Compounding bias can cause learning instability depending on the learning rate and how drastically the policy changes.

The non-stationarity applies for both types of critics since they are both on-policy estimates.
However, centralized value function learning is generally better equipped in the face of non-stationarity because it has no variance in the update targets.
Therefore, the bootstrapping may be more stable in the case of a centralized critic. 
As a result, in a cooperative environment with a moderate number of agents (as discussed later), we expect a centralized critic would learn more stably and be less biased, perhaps counteracting the effect of having larger variance in the policy gradient.

\subsection{Handling MAV and MOV}
\label{discussion_policy_learning}

Learning a decentralized policy requires reasoning about local optimality.
That is, given local information $h_i$, agent $i$ needs to explicitly or implicitly consider the distribution of global information: other agents' experiences and actions $\Pr(a_j,h_j \mid h_i)$, through which agent $i$ needs to take the expectation of global action-history values.
Abstractly, the process of learning via sampling over the joint space of $a_j$ and $h_j$ thus generates MAV and MOV for agent $i$'s policy gradient.
Interestingly, this process is inevitable but takes place in different forms:
during IAC and IACC training, this averaging process is done by different entities.
In IAC, those expectations are implicitly taken by the decentralized critic and produce a single expected value for the local history; this is precisely why decentralized critic learning has unstable learning targets as discussed in Section~\ref{sec:value_function_learning}.
On the other hand, in IACC, the expectation takes place directly in the policy learning.
Different samples of global value estimates are used in policy updates for a local trajectory, hence the higher policy gradient variance we discussed in Section~\ref{sec:variance_analysis}.
Thus, for decentralized policy learning purposes, we expect decentralized critics to give estimates with more bias and less variance, and the centralized critic to give estimates with less bias and more variance.
Consequently, the trade-off largely depends on the domain; as we shall see in the next section, certain domains favor a more stable policy while others favor a more accurate critic.

\subsection{Scalability}

Another important consideration is the scale of the task.
A centralized critic's feature representation needs to scale linearly (in the best case) or exponentially (in the worse case) with the system's number of agents. In contrast, a decentralized critic's number of features can remain constant, and in homogeneous-agent systems, decentralized critics can even share parameters.
Also, some environments may not require much reasoning from other agents.
For example, in environments where agents' decisions rarely depend on other agents' trajectories, the gain of learning value functions jointly is likely to be minimal, and we expect decentralized critics to perform better while having better sample efficiency in those domains.
We show this empirically in Section~\ref{subsec:robustness_in_performance}.

The impact of variance will also change as the number of agents increases.
In particular, when learning stochastic policies with a centralized critic in IACC,
the maximum potential variance in the policy gradient also scales with the number of agents (see Theorem~\ref{theorem:variance}).
On the other hand, IAC's decentralized critics potentially have less stable learning targets in critic bootstrapping with increasing numbers of agents, but the policy updates still have low variance.
Therefore, scalability may be an issue for both methods, and the actual performance is likely to depend on the domain, function approximation setups, and other factors.
However, we expect that IAC should be a better starting point due to more stable policy updates and potentially shared parameters.

\begin{figure*}[ht!]
\vspace{-1cm}
  \begin{subfigure}[t]{.3\linewidth}
  \centering
    \includegraphics[height=3.5cm]{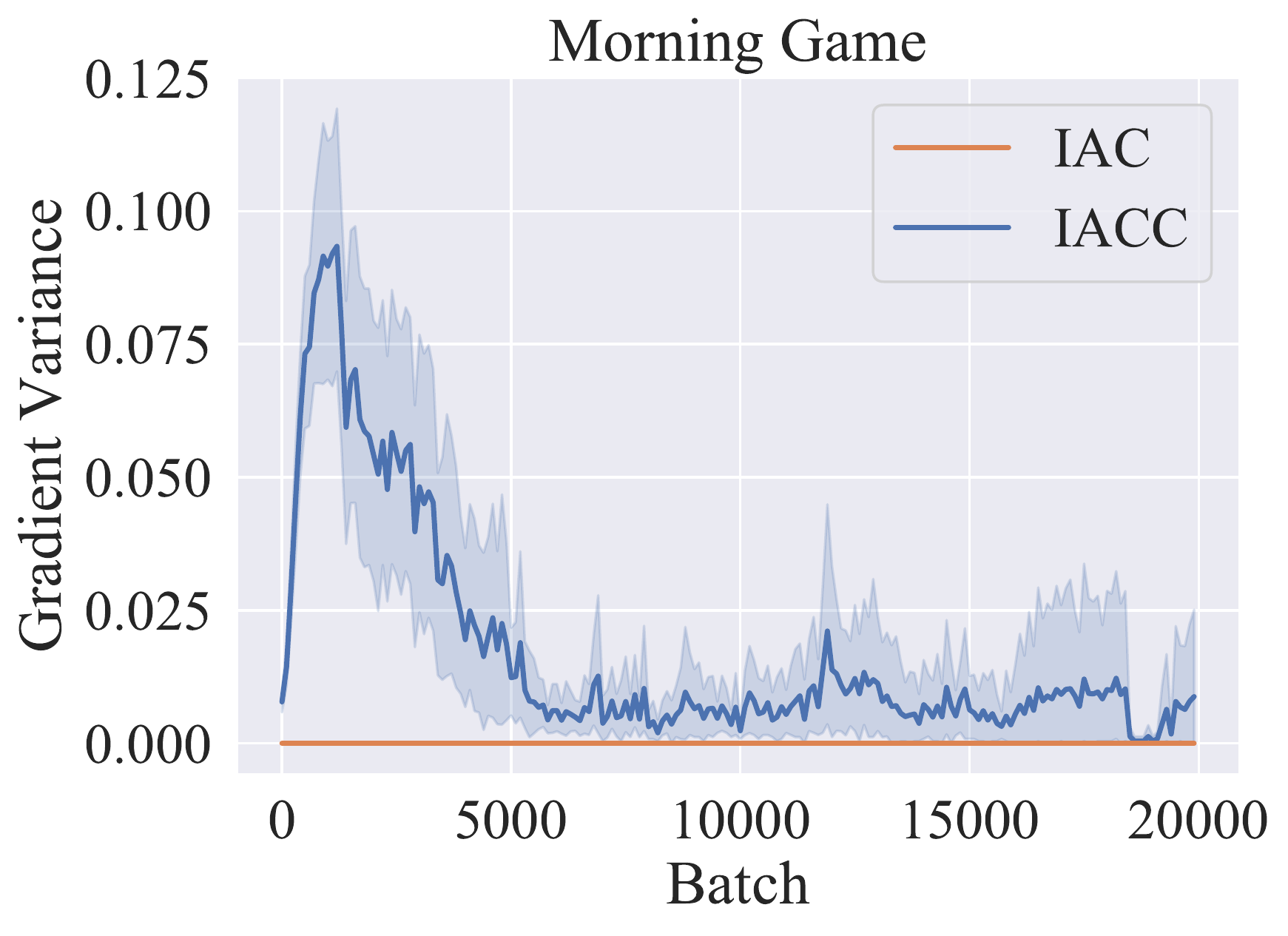}
    \caption{Per-rollout gradient variance for action $pickles$.}\label{subfig:morning_game_var_a1}
  \end{subfigure}\hfill
  \begin{subfigure}[t]{.3\linewidth}
  \centering
    \includegraphics[height=3.5cm]{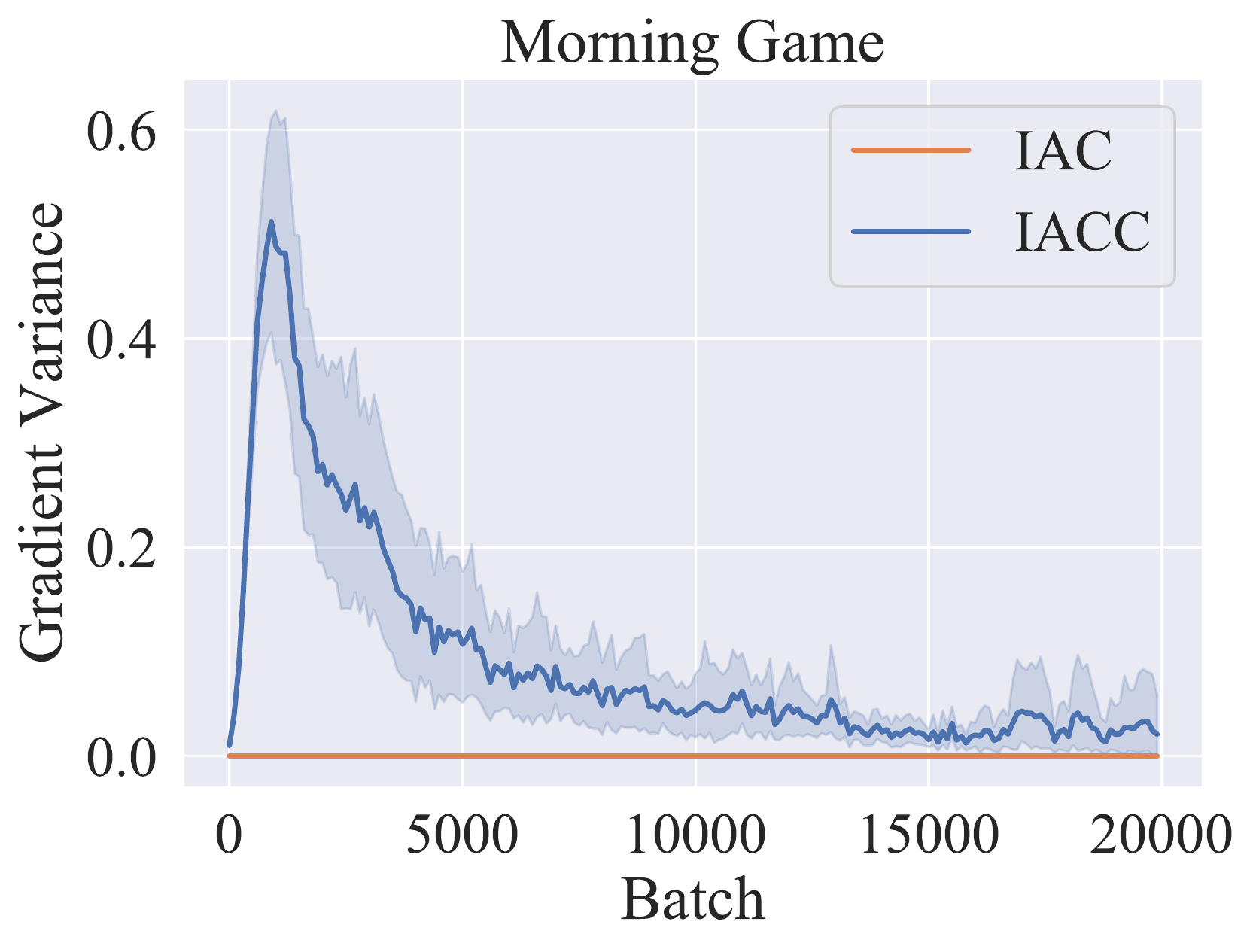}
    \caption{Per-rollout gradient variance for action $cereal$.}\label{subfig:morning_game_var_a2}
  \end{subfigure}\hfill
  \begin{subfigure}[t]{.3\linewidth}
  \centering
    \includegraphics[height=3.5cm]{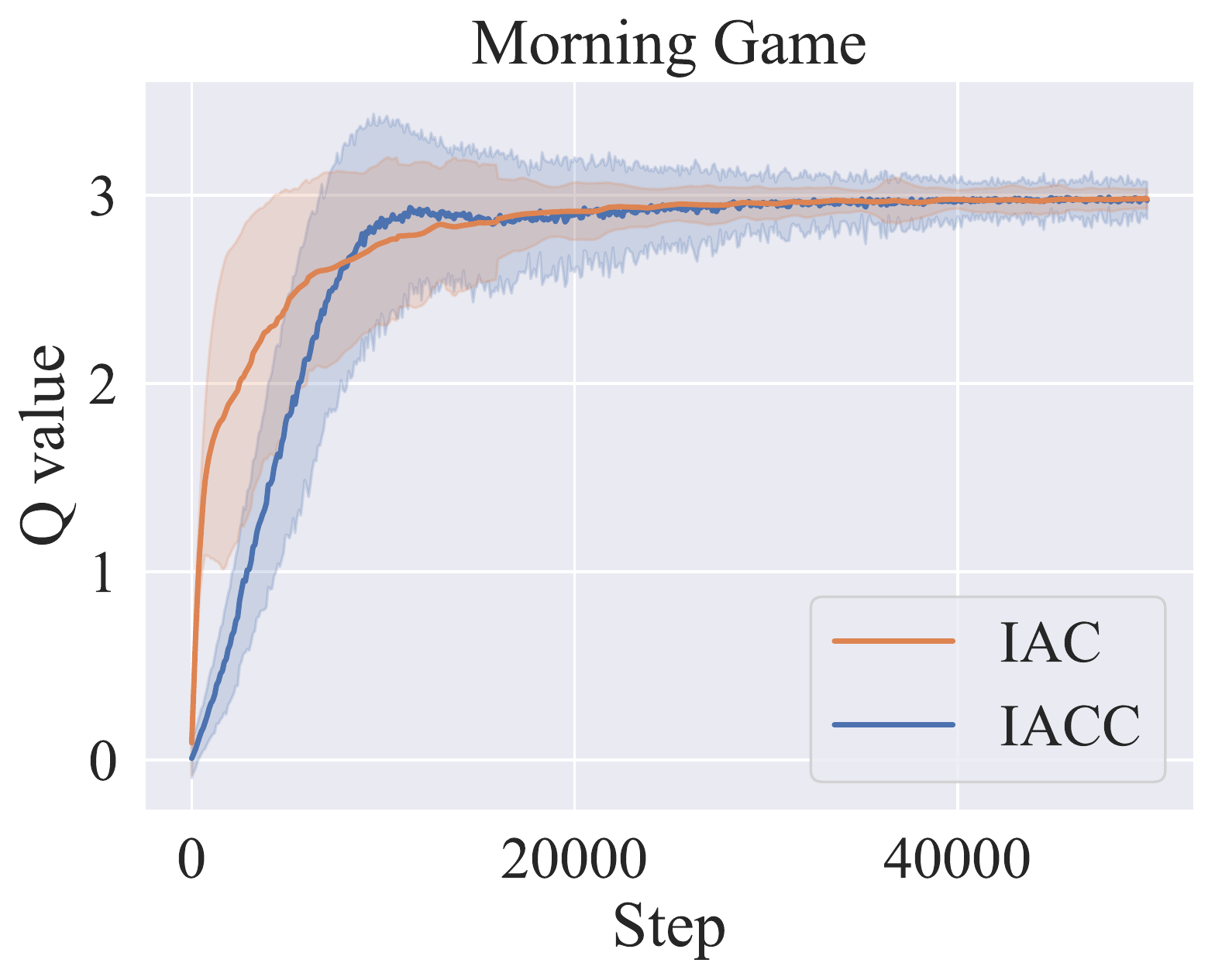}
    \caption{Q values used for updating $\pi(cereal)$; see Figure~\ref{figure:morning_game_value_individual} for clearer illustration.\label{figure:morning_game_value_agg}}\label{subfig:morning_game_var_q}
  \end{subfigure}
  \vspace{-0.1cm}
  \caption{Gradient updates of the Morning Game, shows 200 independent trials for each method.}\label{fig:morning_game_var}
\end{figure*}

\begin{figure*}[ht!]
    \centering
   \captionsetup[subfigure]{labelformat=empty}
    \subcaptionbox{\label{fig:dectiger_results}}
        [0.33\linewidth]{\includegraphics[height=3.5cm]{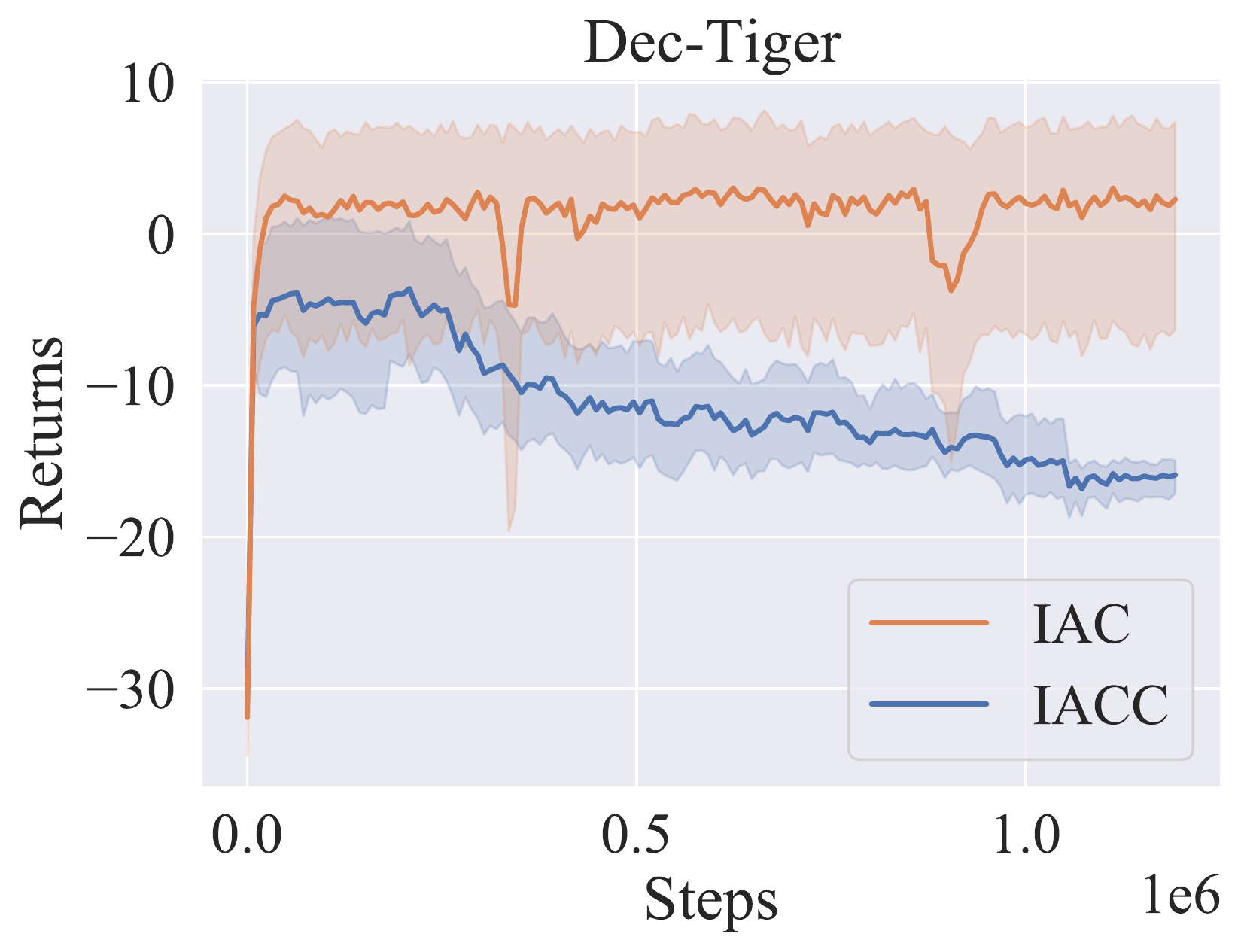}}
    ~
    \subcaptionbox{\label{cleaner}}
        [0.33\linewidth]{\includegraphics[height=3.5cm]{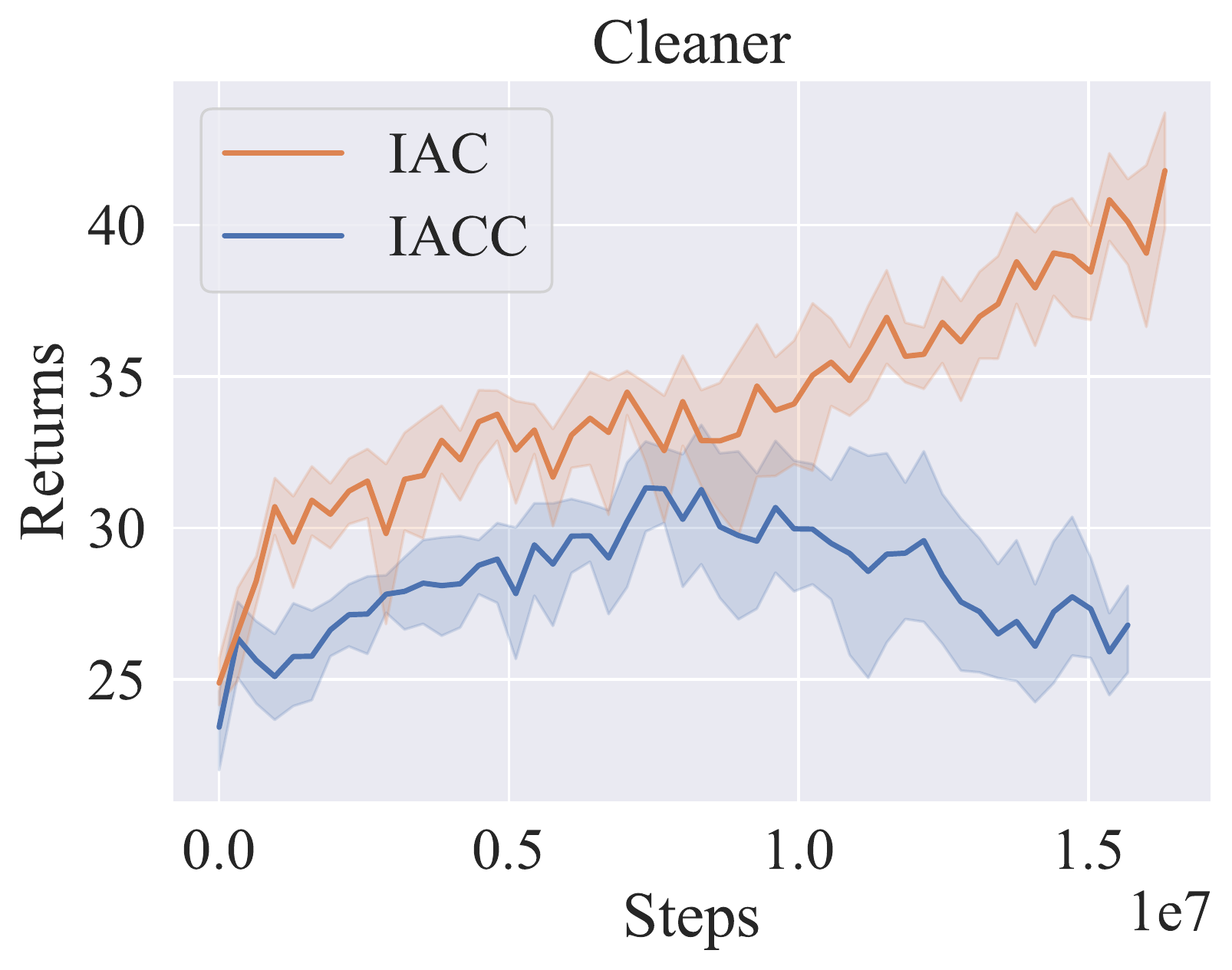}}~
   \subcaptionbox{\label{move_box}}
       [0.33\linewidth]{\includegraphics[height=3.5cm]{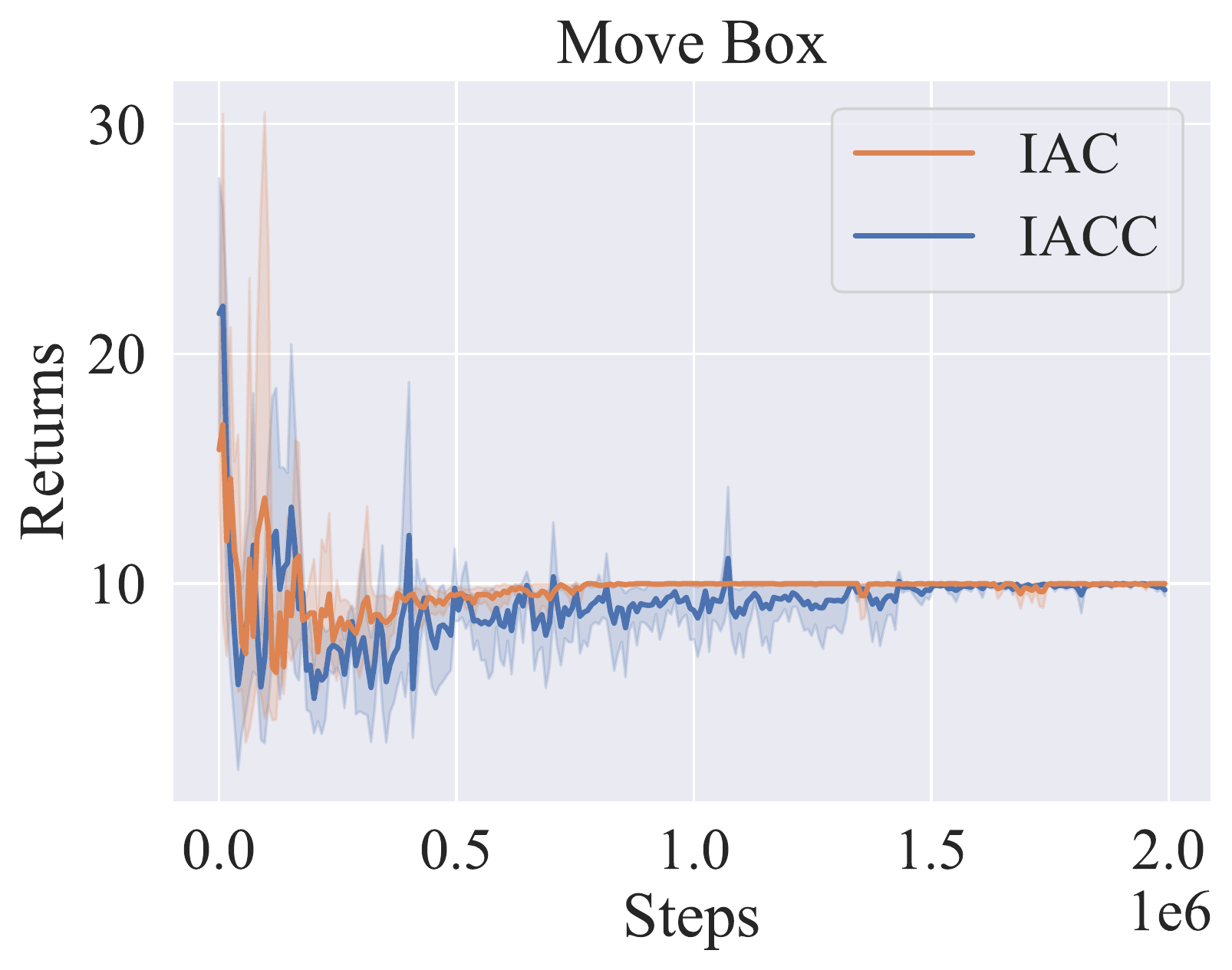}}
    \vspace{-0.6cm}
    \caption{Performance comparison in different domains: (a) Dec-Tiger, (b) Cleaner and (c) Move Box. Dec-Tiger and Cleaner highlight instability resulting from high-variance actor updates employing a centralized critic; Move Box shows the centralized critic is not able to bias the actors towards the joint optimum (at 100).}
    \label{figure:unstable_policies}
    \vspace{-0.2cm}
\end{figure*}

\subsection{The Overall Trade-off}
Combining our discussions in Sections~\ref{sec:value_function_learning} and \ref{discussion_policy_learning}, we conclude that whether to use critic centralization can be essentially considered a bias-variance trade-off decision.
More specifically, it is a trade-off between variance in policy updates and bias in the value function:
a centralized critic should have a lower bias because it will have more stable Q-values that can be updated straightforwardly when policies change, but higher variance because the policy updates need to be averaged over (potentially many) other agents.
In other words, the policies trained by centralized critics avoid more-biased estimates usually produced by decentralized critics, but in return suffer more variance in the training process.
The optimal choice is then largely dependent on the environment settings.
Regardless, the centralized critic likely faces more severe scalability issues in not only critic learning but also in policy gradient variance. 
As a result, we do not expect one method will always dominate the other in terms of performance.

\section{Experiments and Analysis}

In this section, we present experimental results comparing centralized and decentralized critics.
We test on a variety of popular research domains including (but not limited to) classical matrix games, the StarCraft Multi-Agent Challenge~\citep{samvelyan19smac}, the Particle Environments~\citep{mordatch2017emergence}, and the MARL Environments Compilation~\citep{shuo2019maenvs}.
Our hyperparameter tuning uses grid search.
Each figure, if not otherwise specified, shows the aggregation of 20 runs per method.

\subsection{Variance in the Morning Game}
As we can see in Figures~\ref{subfig:morning_game_var_a1} and \ref{subfig:morning_game_var_a2}, the variance per rollout in the policy gradient for both actions is zero for IAC and non-zero for IACC, validating our theoretical variance analysis.
Figure~\ref{subfig:morning_game_var_q} shows how the Q-values evolve for the optimal action $a_2$ in both methods.
First, observe that both types of critics converged to the correct value $3$, which confirms our bias analysis.
Second, the two methods' Q-value variance in fact comes from different sources.
For the decentralized critic, the variance comes from the critics having different biases across trials.
For centralized critics, there is the additional variance that comes from incorporating other agent actions, producing a high value when a teammate chooses $milk$ and a low value when a teammate chooses $vodka$.

\subsection{Unstable Policies}
\label{subsec:unstable_policies}
\subsubsection{Dec-Tiger}
We test on the classic yet difficult Dec-Tiger domain~\citep{nair2003taming},
a multi-agent extension to the Tiger domain~\citep{kaelbling1998planning}.
To end an episode, each agent has a high-reward action (opening a door with treasure inside) and a high-punishment action (opening a door with tiger inside).
The treasure and tiger are randomly initialized in each episode,
hence, a third action (\textit{listen}) gathers noisy information regarding which of the two doors is the rewarding one.
The multi-agent extension of Tiger requires two agents to open the correct door simultaneously in order to gain maximum return.
Conversely, if the bad action is taken simultaneously, the agents take less punishment.
Note that any fast-changing decentralized policies are less likely to coordinate the simultaneous actions with high probability, thus lowering return estimates for the critic and hindering joint policy improvement.
As expected, we see in Figure~\ref{fig:dectiger_results} that IACC (with a centralized critic and higher policy gradient variance) does not perform as well as IAC due.
In the end, the IACC agent learns to avoid high punishment (agents simultaneously open different doors, $-100$) by not gathering any information (\textit{listen}) and opening an agreed-upon door on the first timestep.
IACC gives up completely on high returns (where both agents \textit{listen} for some timesteps and open the correct door at the same time, $+20$) because the unstable policies make coordinating a high return of $+20$ extremely unlikely.

\begin{figure*}
    \vspace{-1cm}
    \centering
   \captionsetup[subfigure]{labelformat=empty}
    \subcaptionbox{}
        [0.33\linewidth]{\includegraphics[height=3.6cm]{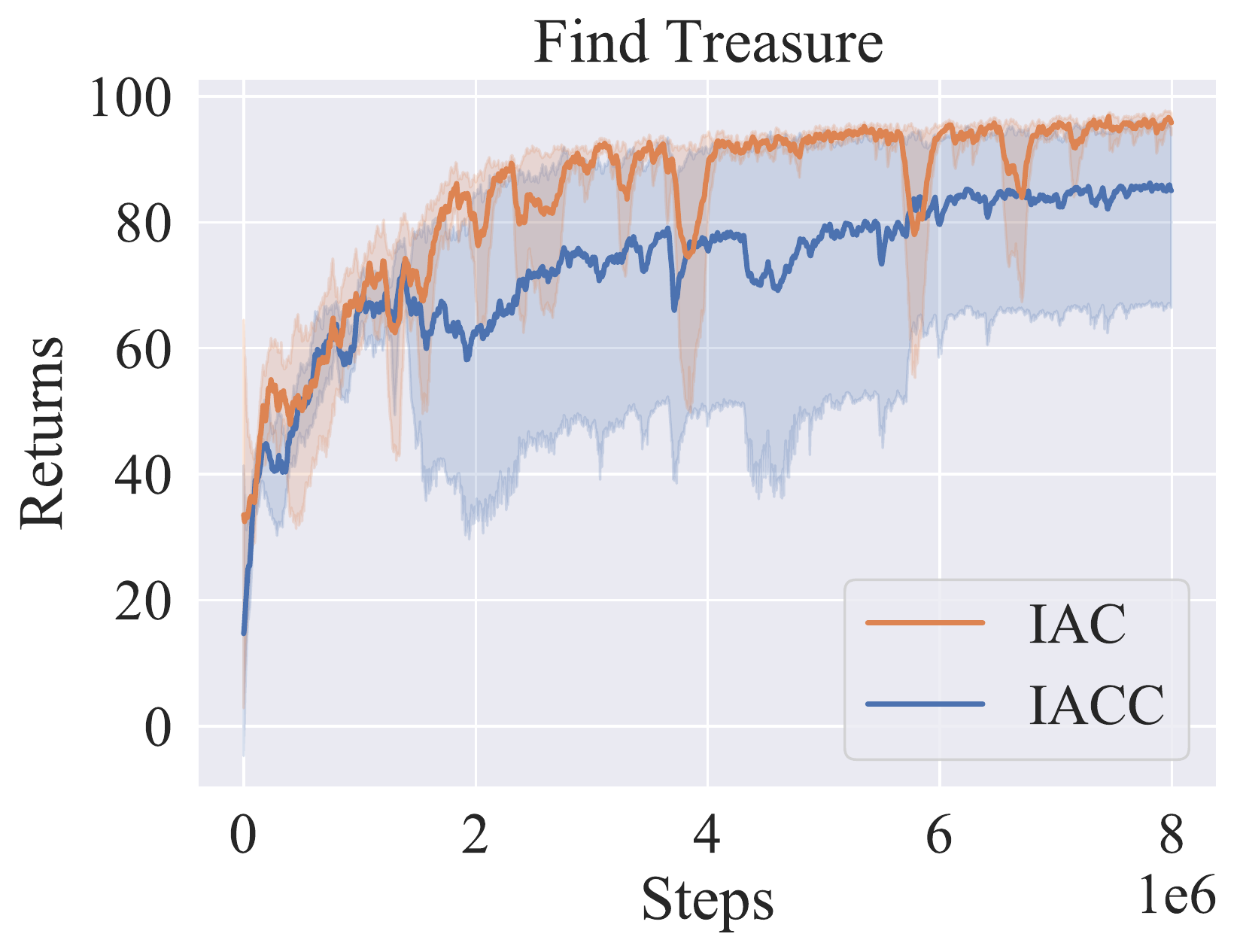}}
    ~
    \subcaptionbox{}
        [0.33\linewidth]{\includegraphics[height=3.6cm]{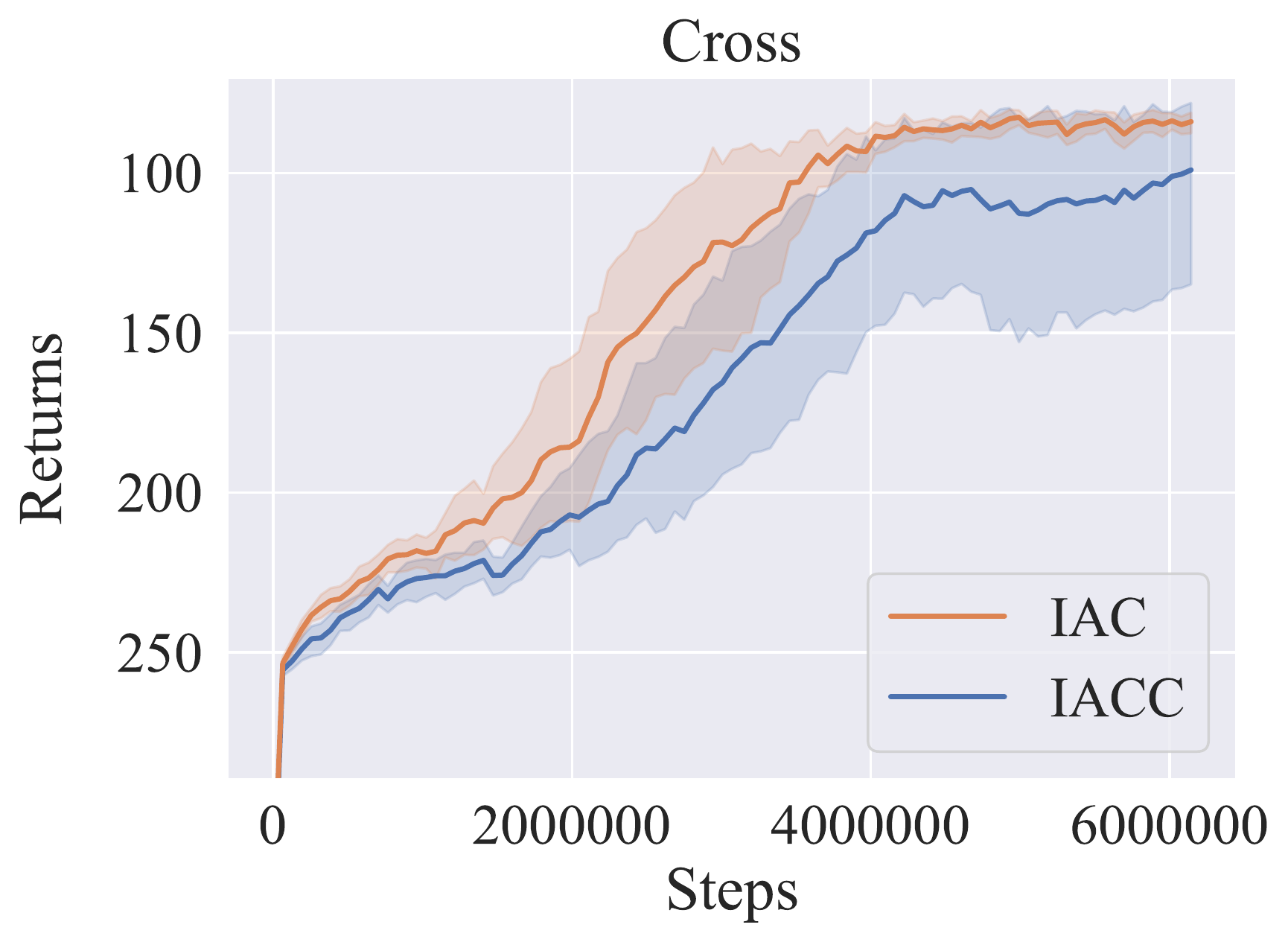}}
    ~
   \subcaptionbox{}
       [0.33\linewidth]{\includegraphics[height=3.6cm]{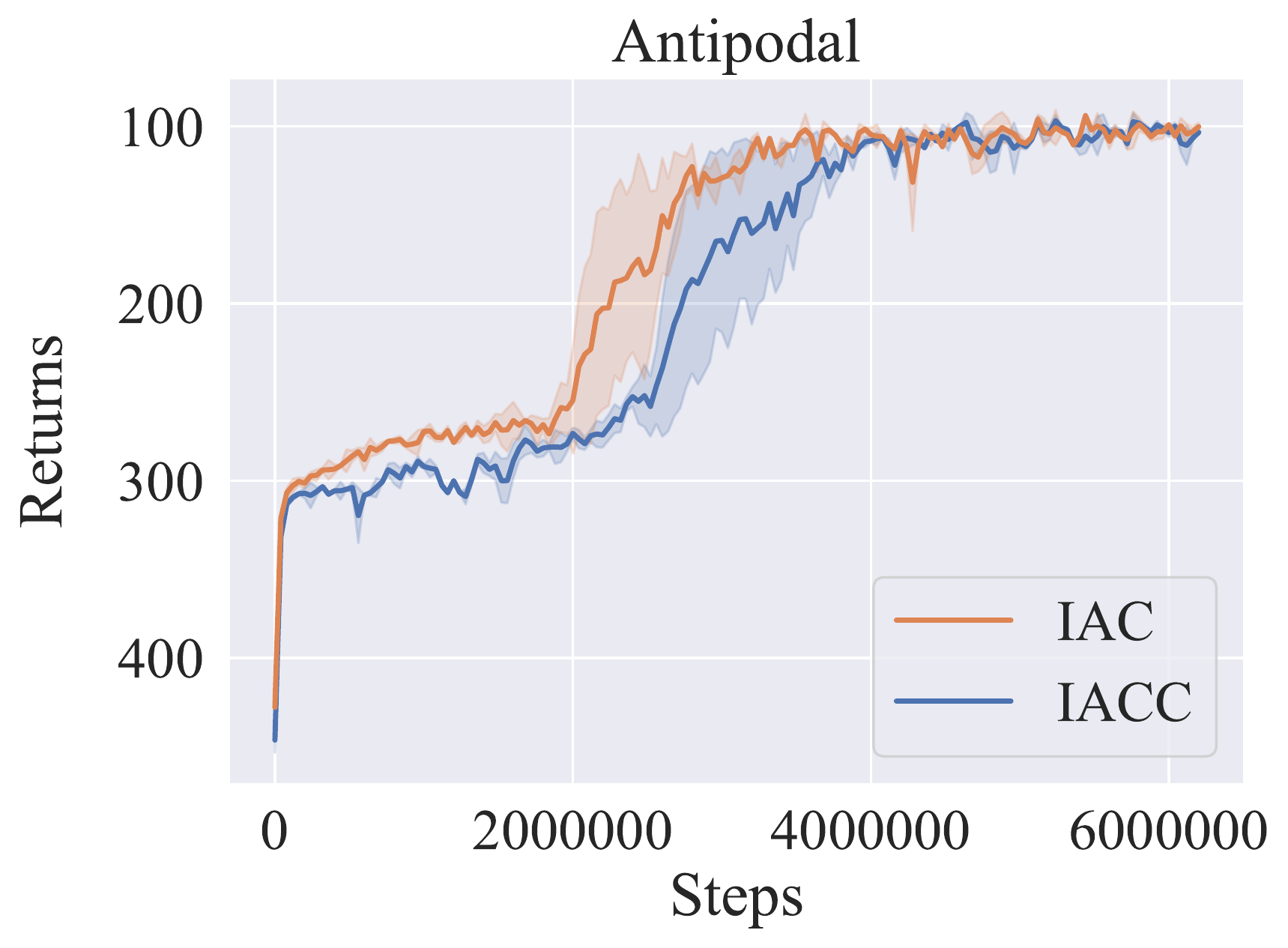}}
    \vspace{-0.7cm}
    \caption{Performance comparison of domains where IAC does better than IACC:, Find Treasure, Cross and Antipodal.}
    \label{figure_larger_var}
\end{figure*}

\begin{figure*}
   \centering
   \captionsetup[subfigure]{labelformat=empty}
   \subcaptionbox{}
       [0.33\linewidth]{\includegraphics[height=3.6cm]{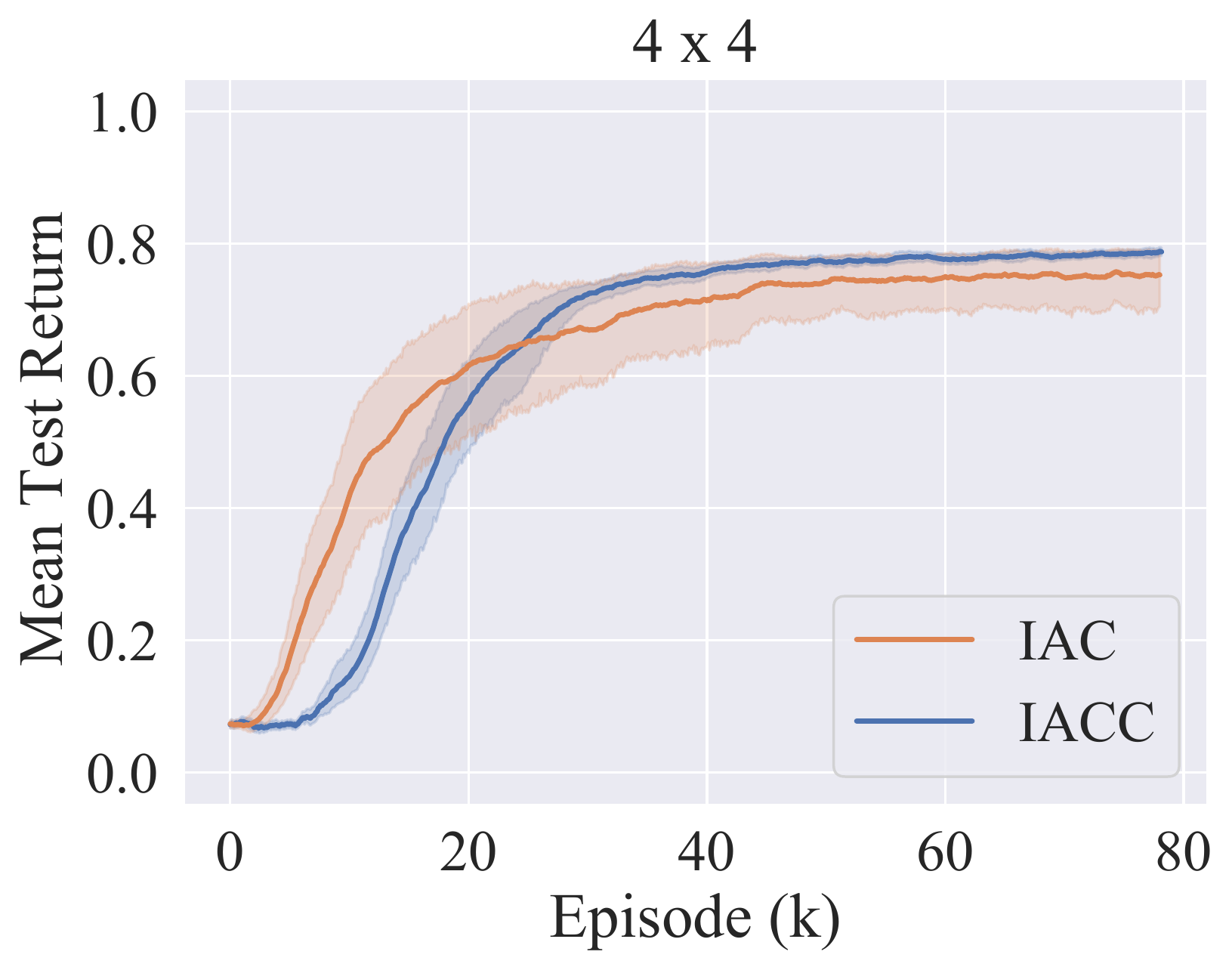}}
   ~
   \centering
   \subcaptionbox{}
       [0.33\linewidth]{\includegraphics[height=3.6cm]{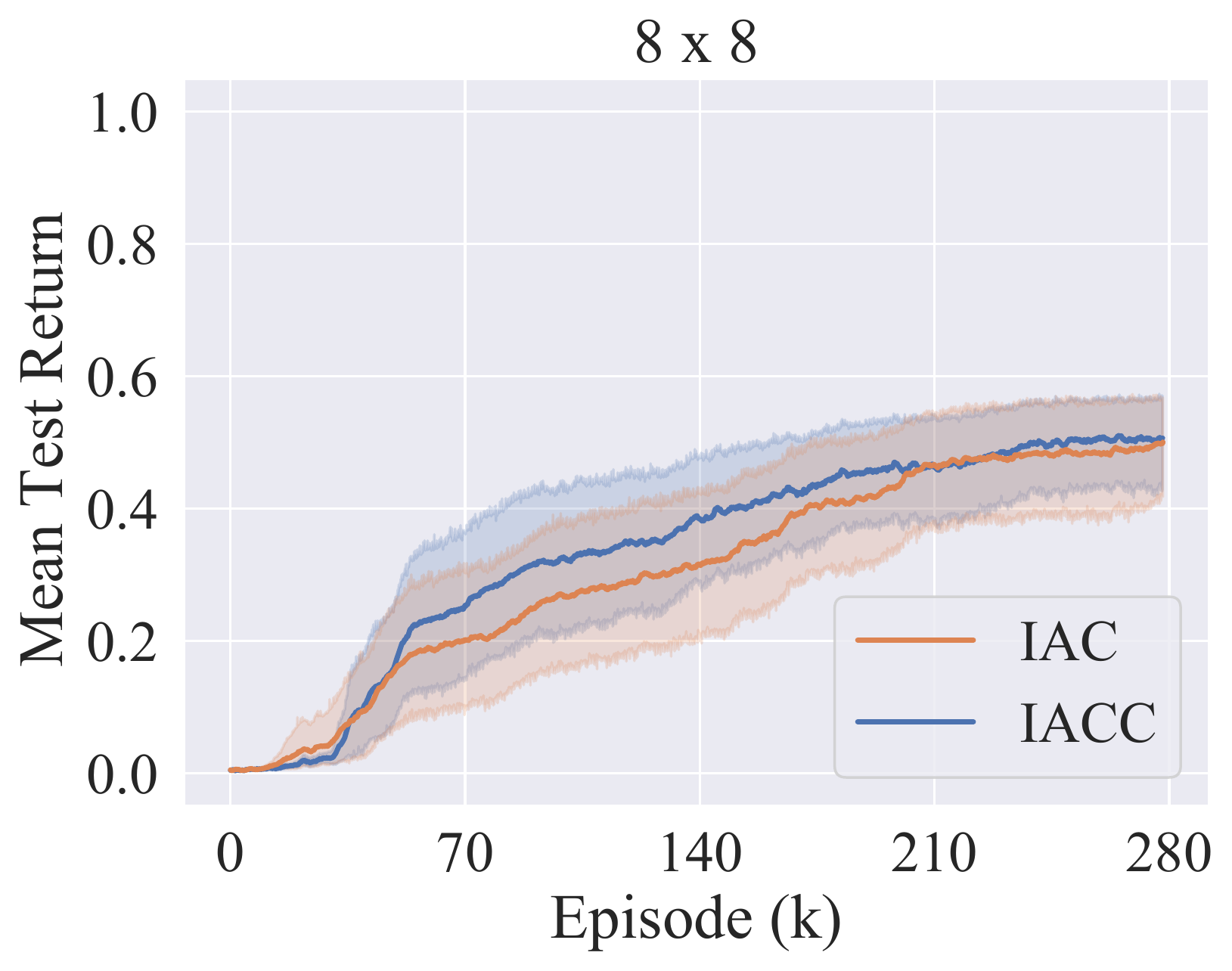}}
   ~
   \centering
   \subcaptionbox{}
       [0.33\linewidth]{\includegraphics[height=3.6cm]{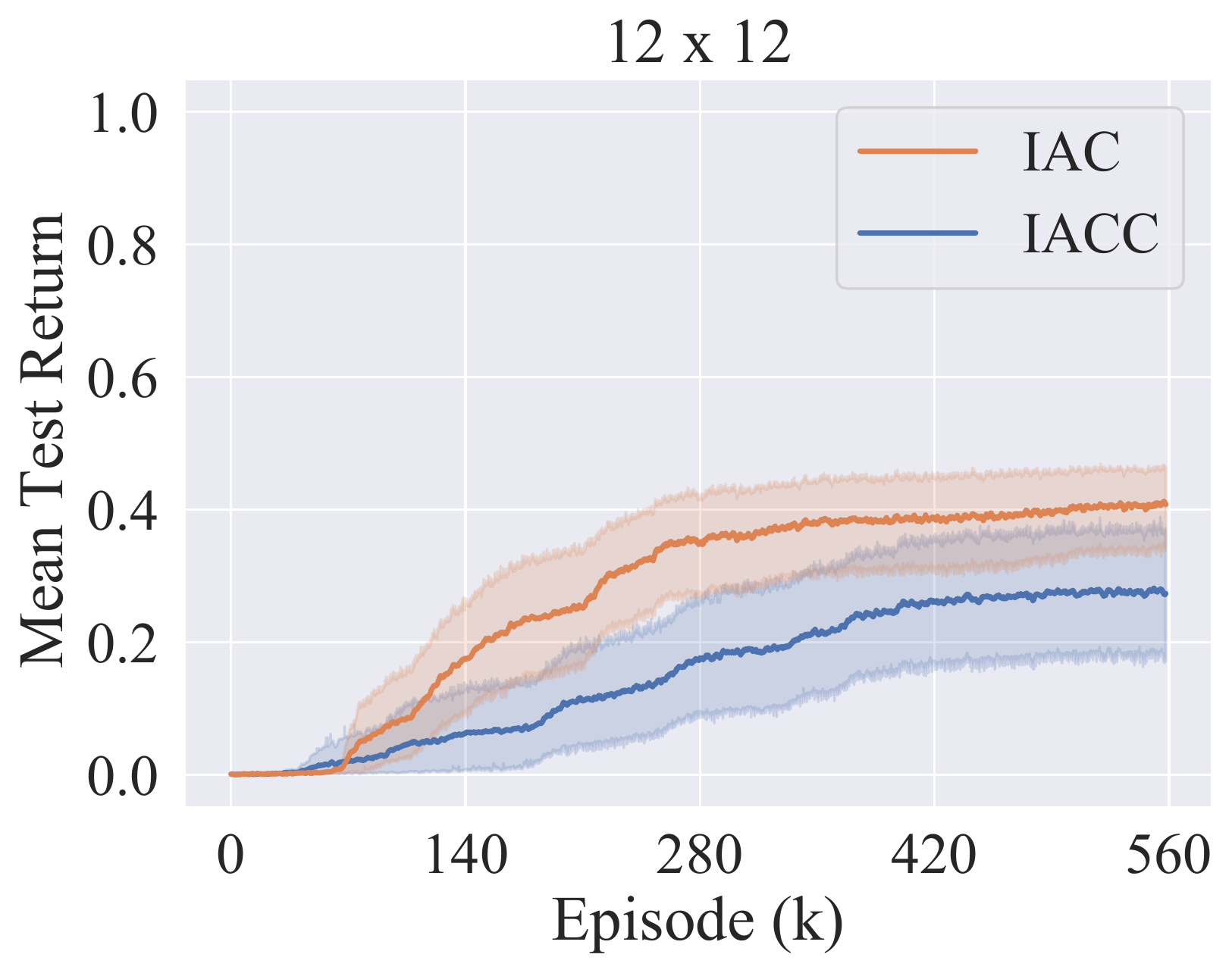}}
    \vspace{-0.6cm}
    \caption{Performance comparison on capture target under various grid world sizes.}
   \label{capture_target_results}
\end{figure*}

\subsubsection{Cleaner}
We observe similar policy degradation in the  Cleaner domain~\citep{shuo2019maenvs}, a grid-world maze in which agents are rewarded for stepping onto novel locations in the maze.
The optimal policy is to have the agents split up and cover as much ground as possible.
The maze has two non-colliding paths (refer to Appendix~\ref{appendix:cleaner} for visualization) so that as soon as the agents split up, they can follow a locally greedy policy to get an optimal return.
However, with a centralized critic (IACC), both agents start to take the longer path with more locations to "clean."
The performance is shown in Figure~\ref{cleaner}.
When policy-shifting agents are not completely locally greedy, the issue is that they cannot "clean" enough ground in their paths.
Subsequently, they discover that having both agents go for the longer path (the lower path) yields a better return, converging to a suboptimal solution.
Again we see that in IACC with centralized critic, due to the high variance we discussed in Section~\ref{discussion_policy_learning}, the safer option is favored, resulting in both agents completely ignoring the other path (performance shown in Figure~\ref{cleaner}). 
Overall, we see that high variance in the policy gradient (in the case of the centralized critic) makes the policy more volatile and can result in poor coordination performance in environments that require coordinated series of actions to discover the optimal solution.

\subsection{Shadowed Equilibrium}

Move Box~\citep{shuo2019maenvs} is another commonly used domain, where grid-world agents are rewarded by pushing a heavy box (requiring both agents) onto any of the two destinations (see Appendix~\ref{appendix:move_box} for details).
The farther destination gives $+100$ reward while the nearer destination gives $+10$.
Naturally, the optimal policy is for both agents to go for $+100$, but if \textit{either} of the agents is unwilling to do so, this optimal option is "shadowed" and \textit{both} agents will have to go for $+10$.
We see in Figure~\ref{move_box} that both methods fall for the \textit{shadowed equilibrium}, favoring the safe but less rewarding option.

Analogous to the Climb Game, even if the centralized critic is able to learn the optimal values due to its unbiased on-policy nature shown in Section~\ref{nobias_subsec},
the on-policy return of the optimal option is extremely low due to an uncooperative teammate policy;
thus, the optimal actions are rarely sampled when updating the policies.
The same applies to both agents, so the system reaches a suboptimal equilibrium, even though IACC is trained in a centralized manner.

\subsection{Robustness in Performance}
\label{subsec:robustness_in_performance}

Through tests on Go Together~\cite{shuo2019maenvs}, Merge~\cite{CM3}, Predator and Prey~\cite{lowe2017multi}, Capture Target~\cite{omidshafiei2017deep,xiao_corl_2019}, Small Box Pushing and SMAC~\cite{samvelyan19smac} tasks (see Appendices~\ref{grid_world_environments}, \ref{cooperative_navigation_environment}, \ref{predator_and_prey_environment}, \ref{capture_target_env}, \ref{small_box_pushing_environment}, \ref{small_box_pushing_3_agents_environment}, and \ref{smac_envionrment}), 
we observe that the two types of critics perform similarly in all these domains,
with IACC (with a centralized critic) being less stable in only a few other domains shown in Figures~\ref{figure:unstable_policies} and \ref{figure_larger_var}.
Since the performance of the two critic types is similar in most results, we expect that it is due to the fact that both are unbiased asymptotically (Lemmas~\ref{lemma:central_convergence}, \ref{lemma:decentral_convergence} and Theorem~\ref{theorem:unbiased_gradient}).
We observe that, although decentralized critics might be more biased when considering finite training, it does not affect real-world performance in a significant fashion in these domains.

In cooperative navigation domains Antipodal, Cross~\citep{mordatch2017emergence, CM3}, and Find Treasure~\citep{shuo2019maenvs},
we observe a more pronounced performance difference among runs (Figure~\ref{figure_larger_var}).
In these cooperative navigation domains (details in Appendices~\ref{appendix:find_treasure} and~\ref{cooperative_navigation_environment}),
there are no suboptimal equilibria that trap the agents, and on most of the timesteps, the optimal action aligns with the locally greedy action.
Those tasks only require agents to coordinate their actions for a few timesteps to avoid collisions.
It appears that those tasks are easy to solve, but the observation space is continuous, thus causing large MOV in the gradient updates for IACC.
Observe that some IACC runs struggle to reach the optimal solution robustly, while IAC robustly converges, conforming to our scalability discussion regarding large MOV.
A centralized critic induces higher variance for policy updates, where the shifting policies can become a drag on the value estimates which, in turn, become a hindrance to improving the policies themselves.

The scalability issue can be better highlighted in environments where we can increase the observation space.
For example, in Capture Target~\cite{lyu2020likelihood}
where agents are rewarded by simultaneously catching a moving target in a grid world (details in Appendix~\ref{capture_target_env}),
by increasing the grid size from $4\times4$ to $12\times12$, we see a notable comparative drop in overall performance for IACC (Figure~\ref{capture_target_results}).
Since an increase in observation space leads to an increase in Multi-Observation Variance (MOV) and nothing else, it indicates that here the policies of IACC do not handle MOV as well as the decentralized critics in IAC.
The result might imply that, for large environments, decentralized critics scale better in the face of MOV due to the fact that they do not involve MOV in policy learning.

\section{Conclusion}

In this paper, we present an examination of critic centralization theoretically and empirically.
The core takeaways are:
1) in theory, centralized and decentralized critics are the same in expectation for the purpose of updating decentralized policies;
2) in theory, centralized critics will lead to higher variance in policy updates;
3) in practice, there is a bias-variance trade-off due to potentially higher bias with limited samples and less-correct value functions with decentralized critics;
4) in practice, a decentralized critic regularly gives more robust performance since stable policy gradients appear to be more crucial than stable value functions in our domains.

Although IACC uses a centralized critic that is trained in a centralized manner, the method does not produce policies that exploit centralized knowledge.
Therefore, future work on IACC may explore feasible ways of biasing the decentralized policies towards a better joint policy by exploiting the centralized information. Reducing variance in policy updates and other methods that make better use of centralized training are promising future directions. 

\section{Acknowledgments}
We thank the reviewers for their helpful feedback.
We also thank Andrea Baisero and Linfeng Zhao for helpful comments and discussions.
This research is supported in part by the U. S. Office of Naval Research under award
number N00014-19-1-2131, Army Research Office award W911NF-20-1-0265 and an Amazon Research Award. 

\clearpage

\bibliographystyle{ACM-Reference-Format}
\bibliography{paper.bib}

\clearpage
\onecolumn

\appendix

\section{Proofs}

\subsection{Proposition~\ref{prop:ss_history}: Existence of Steady-State History Distribution}
\label{appendix:ss_history}

We show that under a (limited-memory) history-based policy, there still exists a stationary history distribution.
Let \(\text{Pr}_0(s)\) denote the probability of \(s\) being the
initial state of an episode, and \(\eta(h,s)\) denote the average number of timesteps spent in state \(s\) with a history of \(h\) in a single episode, where \(\bar{s}\) and
\(\bar{h}\) denote the state and history from the previous timestep and \(h_0\) is
the empty history:
\begin{equation*}
 \begin{aligned}
 \eta(h,s) = \text{Pr}_0(s) \cdot \mathbb{I}(h = h_0) + 
  \sum_{\bar{h},\bar{s}}\eta(\bar{h},\bar{s})
  \sum_a\pi(a\mid \bar{h}) \cdot \Pr(h\mid \bar{h},s) \cdot \Pr(s\mid \bar{s},a) \\
 \quad \text{for all } s, h \in \mathbf{S} \times \mathbf{H}^d.
 \end{aligned}
\end{equation*}
Assuming our environment has a stationary observation model \(\Pr(h\mid \bar{h},s)
= \Pr(o\mid s)\) where the resulting Markov chains are irreducible and aperiodic, a stationary transition model \(\Pr(s\mid \bar{s},a)\), and a
given fixed policy \(\pi(a\mid \bar{h})\), we can solve the system of equations
for \(\eta(h,s)\) and then normalize to get the stationary distribution of state
and history pairs:
\begin{equation*}
    \Pr(h,s) = \frac{\eta(h,s)}{\sum_{s^\prime,h^\prime}\eta(s^\prime,h^\prime)}
    \quad \text{for all } s, h \in \mathbf{S} \times \mathbf{H}^d.
\end{equation*}
However, what interests us is \(\Pr(h\mid s) = \Pr(h,s)/\Pr(s)\),
for that, we can express \(\eta(s)\) in a similar fashion:
\begin{equation*}
 \eta(s) = \text{Pr}_0(s) + \sum_{\bar{h},\bar{s}} \eta(\bar{h},\bar{s})
                    \sum_{a} \pi(a\mid \bar{h})\Pr(s\mid \bar{s},a) 
\end{equation*}
which can be normalized to obtain $\Pr(s)$.
Similarly, one can obtain a stationary distribution $\Pr(h)$ for fixed-length histories.

\subsection{Lemma~\ref{lemma:central_convergence}: Convergence of Centralized Critic}
\label{appendix:central_convergence}

\begin{proof}
    We begin by observing the $1$-step update rule for the central critic:
    \begin{equation}
        \label{eq:central_backup}
        Q(\joint{h},\joint{a}) \gets Q(\joint{h},\joint{a})
        + \alpha \left[ R(s,\joint{a},s') + \gamma Q(\joint{h}',\joint{a}') \right]
    \end{equation}
    This update occurs with conditional probability
    $\Pr(s',\joint{h}',\joint{a}' \mid \joint{h},\joint{a})
    = \Pr(s \mid \joint{h})
    \cdot \Pr(o \mid s,\joint{a})
    \cdot \Pr(s' \mid s,\joint{a})
    \cdot \joint{\pi}(\joint{a}' \mid \joint{h}')$
    given that $(\joint{h},\joint{a})$ occurred,
    where $\Pr(s \mid \joint{h})$ exists by Proposition~\ref{prop:ss_history}.
    Assuming that this $(\joint{h},\joint{a})$-combination is visited infinitely often during an infinite amount of training,
    and that $\alpha$ is annealed according to the stochastic approximation criteria from \cite{robbins1951stochastic},
    then $Q(\joint{h},\joint{a})$ will asymptotically converge to the expectation of the bracketed term in (\ref{eq:central_backup}).
    We can compute the expected value by summing over all histories and actions weighted by their joint probabilities of occurrence:
    \begin{align}
        \nonumber
        Q(\joint{h},\joint{a})
        &\gets \sum_{s,o,s',\joint{a}'} \Pr(s \mid \joint{h}) \cdot \Pr(o \mid s,\joint{a}) \cdot \Pr(s' \mid s,\joint{a}) \cdot \joint{\pi}(\joint{a}' \mid \joint{h}')
        \left[ R(s,\joint{a},s') + \gamma Q(\joint{h}',\joint{a}') \right] \\
        \label{eq:central_bellman}
        &= \sum_{s,s'} \Pr(s \mid \joint{h}) \cdot \Pr(s' \mid s,\joint{a})
        \left[ R(s,\joint{a},s') + \gamma \sum_{o,\joint{a}'} \Pr(o \mid s,\joint{a}) \cdot \joint{\pi}(\joint{a}' \mid \joint{h}') \cdot Q(\joint{h}',\joint{a}') \right]
    \end{align}
    This represents the Bellman equation for the centralized critic.
    Let us denote this update rule by the operator $B_c$ such that $Q \gets B_c Q$ is equivalent to (\ref{eq:central_bellman}).
    We can show that $B_c$ is a contraction mapping.
    Letting $\norm{Q} \coloneqq \max\limits_{\joint{h},\joint{a}} \abs{Q(\joint{h},\joint{a})}$,
    \begin{align*}
        \norm{B_c Q_1 - B_c Q_2}
        &= \max\limits_{\joint{h},\joint{a}} \abs{
        \sum_{s,s'} \Pr(s \mid \joint{h}) \cdot \Pr(s' \mid s,\joint{a})
        \left[ \gamma \sum_{o,\joint{a}'} \Pr(o \mid s,\joint{a}) \cdot \joint{\pi}(\joint{a}' \mid \joint{h}') \left[Q_1(\joint{h}',\joint{a}') - Q_2(\joint{h}',\joint{a}')\right] \right]
        } \\
        &\leq \gamma \max\limits_{\joint{h},\joint{a}} \sum_{s,o,s',\joint{a}'} \Pr(s \mid \joint{h}) \cdot \Pr(s' \mid s,\joint{a}) \cdot \Pr(o \mid s,\joint{a}) \cdot \joint{\pi}(\joint{a}' \mid \joint{h}')
        \abs{Q_1(\joint{h}',\joint{a}') - Q_2(\joint{h}',\joint{a}')} \\
        &\leq \gamma \max\limits_{\joint{h}',\joint{a}'} \abs{Q_1(\joint{h}',\joint{a}') - Q_2(\joint{h}',\joint{a}')} \\
        &= \gamma \norm{Q_1 - Q_2}
    \end{align*}
    The first inequality is due to Jensen's inequality, while the second follows because a convex combination cannot be greater than the maximum.
    Hence, $B_c$ is a contraction mapping whenever $\gamma < 1$, which immediately implies the existence of a unique fixed point.
    (If $B_c$ had two fixed points, then
    $\norm{B_c Q^*_1 - B_c Q^*_2} = \norm{Q^*_1 - Q^*_2}$,
    which would contradict our above result for $\gamma < 1$.)
    We now must identify the fixed point to complete our proof.
    Note that the on-policy expected return $Q^\joint{\pi}(\joint{h},\joint{a})$ is invariant under the expectation
    $\expect{\joint{\pi}}{Q^\joint{\pi}(\joint{h},\joint{a})} = Q^\joint{\pi}(\joint{h},\joint{a})$,
    making it a likely candidate for the fixed point.
    We can verify by applying $B_c$ to it:
    \begin{align*}
        B_c Q^\joint{\pi}
        &= \sum_{s,s'} \Pr(s \mid \joint{h}) \cdot \Pr(s' \mid s,\joint{a})
        \left[ R(s,\joint{a},s') + \gamma \sum_{o,\joint{a}'} \Pr(o \mid s,\joint{a}) \cdot \joint{\pi}(\joint{a}' \mid \joint{h}') \cdot Q^\joint{\pi}(\joint{h}',\joint{a}') \right] \\
        &= \sum_{s,s'} \Pr(s \mid \joint{h}) \cdot \Pr(s' \mid s,\joint{a})
        \left[ R(s,\joint{a},s') + \gamma \sum_{o} \Pr(o \mid s,\joint{a}) \cdot V^\joint{\pi}(\joint{h}') \right] \\
        &= \sum_{s,o,s'} \Pr(s \mid \joint{h}) \cdot \Pr(s' \mid s,\joint{a}) \cdot \Pr(o \mid s,\joint{a})
        \left[ R(s,\joint{a},s') + \gamma V^\joint{\pi}(\joint{h}') \right] \\
        &= Q^\joint{\pi}
    \end{align*}
    The final step follows from the recursive definition of the Bellman equation.
    We therefore have $B_c Q^\joint{\pi} = Q^\joint{\pi}$, which makes $Q^\joint{\pi}$ the unique fixed point of $B_c$ and completes the proof.
\end{proof}

\subsection{Lemma~\ref{lemma:decentral_convergence}: Convergence of Decentralized Critic}
\label{appendix:decentral_convergence}

\begin{proof}
    Without loss of generality, we will consider the $2$-agent case.
    The result can easily be generalized to the $n$-agent case by treating all of the agent policies except $\pi_i$ as a joint policy $\joint{\pi}_j$.
    We begin by observing the $1$-step update rule of the $i$-th decentralized critic:
    \begin{equation}
        \label{eq:decentral_backup}
        Q(h_i,a_i) \gets Q(h_i,a_i)
        + \alpha \left[ R(s,\joint{a},s') + \gamma Q(h_i',a_i') \right]
    \end{equation}
    Letting $\pi \coloneqq \pi_i$, this update occurs with conditional probability
    $\Pr(\joint{a},s',h_i',a_i' \mid h_i,a_i)
    = \Pr(h_j \mid h_i)
    \cdot \pi_j(a_j \mid h_j)
    \cdot \Pr(s \mid \joint{h})
    \cdot \Pr(o \mid s,a_i)
    \cdot \Pr(s' \mid s,\joint{a})
    \cdot \pi(a_i' \mid h_i')$
    given that $(h_i,a_i)$ occurred,
    where $\Pr(s \mid \joint{h})$ exists by Proposition~\ref{prop:ss_history}.
    Assuming that this $(h_i,a_i)$-combination is visited infinitely often during an infinite amount of training,
    and that $\alpha$ is annealed according to the stochastic approximation criteria from \cite{robbins1951stochastic},
    then $Q(h_i,a_i)$ will asymptotically converge to the expectation of the bracketed term in (\ref{eq:decentral_backup}).
    We can compute the expected value by summing over all histories and actions weighted by their joint probabilities of occurrence:
    \begin{align}
        \nonumber
        Q(h_i,a_i)
        &\gets \sum_{h_j,a_j,s,o,s',a_i'} \Pr(h_j \mid h_i) \cdot \pi_j(a_j \mid h_j) \cdot \Pr(s \mid \joint{h}) \cdot \Pr(o \mid s,a_i) \cdot \Pr(s' \mid s,\joint{a}) \cdot \pi(a_i' \mid h_i')
        \left[ R(s,\joint{a},s') + \gamma Q(h_i',a_i') \right] \\
        \label{eq:decentral_bellman}
        &= \sum_{h_j,a_j,s,s'} \Pr(h_j \mid h_i) \cdot \pi_j(a_j \mid h_j) \cdot \Pr(s \mid \joint{h}) \cdot \Pr(s' \mid s,\joint{a})
        \left[ R(s,\joint{a},s') + \gamma \sum_{o,a_i'} \Pr(o \mid s,a_i) \cdot \pi(a_i' \mid h_i') \cdot Q(h_i',a_i') \right]
    \end{align}
    This represents the Bellman equation for a decentralized critic.
    Let us denote this update rule by the operator $B_d$ such that $Q \gets B_d Q$ is equivalent to (\ref{eq:decentral_bellman}).
    We can show that $B_d$ is a contraction mapping.
    Letting $\norm{Q} \coloneqq \max\limits_{h_i,a_i} \abs{Q(h_i,a_i)}$,
    \begin{align*}
        \norm{B_d Q_1 - B_d Q_2}
        &= \max\limits_{h_i,a_i} \abs{
        \sum_{h_j,a_j,s,s'} \Pr(h_j \mid h_i) \cdot \pi_j(a_j \mid h_j) \cdot \Pr(s \mid \joint{h}) \cdot \Pr(s' \mid s,\joint{a})
        \left[ \gamma \sum_{o,a_i'} \Pr(o \mid s,a_i) \cdot \pi(a'_i \mid h_i) \left[ Q_1(h'_i,a'_i) - Q_2(h'_i,a'_i) \right] \right]
        } \\
        &\leq \gamma \max\limits_{h_i,a_i} \sum_{h_j,a_j,s,o,s',a_i'}
        \Pr(h_j \mid h_i) \cdot \pi_j(a_j \mid h_j) \cdot \Pr(s \mid \joint{h}) \cdot \Pr(s' \mid s,\joint{a}) \cdot \Pr(o \mid s,a_i) \cdot \pi(a'_i \mid h'_i)
        \abs{Q_1(h'_i,a'_i) - Q_2(h'_i,a'_i)} \\
        &\leq \gamma \max\limits_{h'_i,a'_i} \abs{Q_1(h'_i,a'_i) - Q_2(h'_i,a'_i)} \\
        &= \gamma \norm{Q_1 - Q_2}
    \end{align*}
    The first inequality is due to Jensen's inequality, while the second follows because a convex combination cannot be greater than the maximum.
    Hence, $B_d$ is a contraction mapping whenever $\gamma < 1$, which immediately implies the existence of a unique fixed point;
    we must identify the fixed point to complete our proof.
    We will test the marginal expectation of the central critic $\expect{h_j,a_j}{Q^\joint{\pi}(h_i,h_j,a_i,a_j)}$.
    We can verify that this is the fixed point by applying $B_d$ to it:
    \begin{align*}
        B_d Q^\pi
        &= \sum_{h_j,a_j,s,s'} \Pr(h_j \mid h_i) \cdot \pi_j(a_j \mid h_j) \cdot \Pr(s \mid \joint{h}) \cdot \Pr(s' \mid s,\joint{a})
        \left[ R(s,\joint{a},s') + \gamma \sum_{o,a_i'} \Pr(o \mid s,a_i) \cdot \pi(a'_i \mid h_i) \cdot \expect{h'_j,a'_j}{Q^\joint{\pi}(h'_i,h'_j,a'_i,a'_j)} \right] \\
        &= \sum_{h_j,a_j,s,s'} \Pr(h_j \mid h_i) \cdot \pi_j(a_j \mid h_j) \cdot \Pr(s \mid \joint{h}) \cdot \Pr(s' \mid s,\joint{a})
        \left[ R(s,\joint{a},s') + \gamma \sum_{o} \Pr(o \mid s,a_i) \cdot \expect{h'_j}{V^\joint{\pi}(h'_i,h'_j)} \right] \\
        &= \sum_{h_j,a_j,s,o,s'} \Pr(h_j \mid h_i) \cdot \pi_j(a_j \mid h_j) \cdot \Pr(s \mid \joint{h}) \cdot \Pr(s' \mid s,\joint{a}) \cdot \Pr(o \mid s,a_i)
        \left[ R(s,\joint{a},s') + \gamma \expect{h'_j}{V^\joint{\pi}(h'_i,h'_j)} \right] \\
        &= Q^\pi
    \end{align*}
    The final step follows from the recursive definition of the Bellman equation.
    We therefore have
    $\smash{ B_d \left( \expect{h_j,a_j}{Q^\joint{\pi}(h_i,h_j,a_i,a_j)} \right) } =\linebreak \expect{h_j,a_j}{Q^\joint{\pi}(h_i,h_j,a_i,a_j)}$,
    which makes
    $\expect{h_j,a_j}{Q^\joint{\pi}(h_i,h_j,a_i,a_j)}$
    the unique fixed point of $B_d$ and completes the proof.
\end{proof}

\subsection{Theorem~\ref{theorem:unbiased_gradient}: Unbiased Policy Gradient}
\label{appendix:unbiased_gradient}

\begin{proof}
    From Lemmas~\ref{lemma:central_convergence} and~\ref{lemma:decentral_convergence}, we know that
    $Q(h_i,h_j,a_i,a_j) \to Q^\joint{\pi} (h_i,h_j,a_i,a)$
    and
    $Q(h_i,a_i) \to \expect{h_j,a_j}{Q^\joint{\pi} (h_i,h_j,a_i,a_j)}$
    for the centralized and decentralized critics, respectively, after an infinite amount of training.
    Substituting these into the definitions in~(\ref{central_critic_policy_update}) and~(\ref{decentral_critic_policy_update}), respectively, we can begin to relate the expected policy gradients after the critics have converged.
    Without loss of generality, we will compute these objectives for the $i$-th agent with policy $\pi_i$.
    Starting with the decentralized critic,
    \begin{equation}
        \label{eq:central_objective_converged}
        \expect{h_i,a_i}{J_d(\theta)} = \expect{h_i,a_i}{ \expect{h_j,a_j}{Q^\joint{\pi} (h_i,h_j,a_i,a_j)} \cdot \nabla_\theta \log \pi_i(a_i \mid h_i) }
    \end{equation}
    Next, we can do the same for the centralized critic and split the expectation:
    \begin{align}
        \label{eq:decentral_objective_converged}
        \expect{h_i,h_j,a_i,a_j}{J_c(\theta)}
        &= \expect{h_i,h_j,a_i,a_j}{Q^\joint{\pi} (h_i,h_j,a_i,a_j) \cdot \nabla_\theta \log \pi_i(a_i \mid h_i)} \\
        \nonumber
        &= \expect{h_i,a_i}{\expect{h_j,a_j}{ Q^\joint{\pi} (h_i,h_j,a_i,a_j) \cdot \nabla_\theta \log \pi_i(a_i \mid h_i) } } \\
        \nonumber
        &= \expect{h_i,a_i}{ \expect{h_j,a_j}{ Q^\joint{\pi} (h_i,h_j,a_i,a_j) } \cdot \nabla_\theta \log \pi_i(a_i \mid h_i) } \\
        \nonumber
        &= \expect{h_i,a_i}{J_d(\theta)}
    \end{align}
    We therefore have
    $\expect{h_i,h_j,a_i,a_j}{J_c(\theta)} = \expect{h_i,a_i}{J_d(\theta)}$,
    so both policy gradients are equal in expectation and can be interchanged without introducing bias.
\end{proof}

\subsection{Theorem~\ref{theorem:variance}: Policy Gradient Variance Inequality}
\label{appendix:variance}

\begin{proof}
    In Theorem~\ref{theorem:unbiased_gradient}, we derived policy gradients for the centralized and decentralized critics in (\ref{eq:central_objective_converged}) and (\ref{eq:decentral_objective_converged}), respectively.
    We also showed that they are equal in expectation;
    hence, let $\mu = \expect{h_i,h_j,a_i,a_j}{J_c(\theta)} = \expect{h_i,a_i}{J_d(\theta)}$.
    Additionally, let
    ${A = (\nabla_\theta \log \pi_i(a_i \mid h_i)) (\nabla_\theta \log \pi_i(a_i \mid h_i))\tran}$.
    We can analyze the relative magnitudes of the policy gradient variances by subtracting them and comparing to zero:
    \begin{align*}
        \Var(J_c(\theta)) - \Var(J_d(\theta))
        &= \left( \expect{h_i,h_j,a_i,a_j}{J_c(\theta) J_c(\theta)\tran} - \mu\mu\tran \right) - \left( \expect{h_i,a_i}{J_d(\theta) J_d(\theta)\tran} - \mu\mu\tran \right) \\
        &= \expect{h_i,h_j,a_i,a_j}{J_c(\theta) J_c(\theta)\tran} - \expect{h_i,a_i}{J_d(\theta) J_d(\theta)\tran} \\
        &= \expect{h_i,h_j,a_i,a_j}{Q^\joint{\pi} (h_i,h_j,a_i,a_j)^2 A}
        - \expect{h_i,a_i}{\expect{h_j,a_j}{Q^\joint{\pi} (h_i,h_j,a_i,a_j)}^2 A} \\
        &= \expect{h_i,a_i}{\underbrace{\left( \expect{h_j,a_j}{Q^\joint{\pi} (h_i,h_j,a_i,a_j)^2} - \expect{h_j,a_j}{Q^\joint{\pi} (h_i,h_j,a_i,a_j)}^2 \right)}_{c} A} \\
        &\geq 0 \quad \text{(element-wise)}
    \end{align*}
    The final inequality follows because $c \geq 0$ by Jensen's inequality (the quadratic is a convex function), and all of the elements of $A$ are nonnegative because it is an outer product of a vector with itself.
    We must therefore have that each element of the centralized critic's policy covariance matrix is at least as large as its corresponding element in the decentralized critic's policy covariance matrix.
    This completes the proof.
\end{proof}

\clearpage

\section{Morning Game}

\begin{figure}[ht!]
  \centering
  \includegraphics[width=0.5\textwidth]{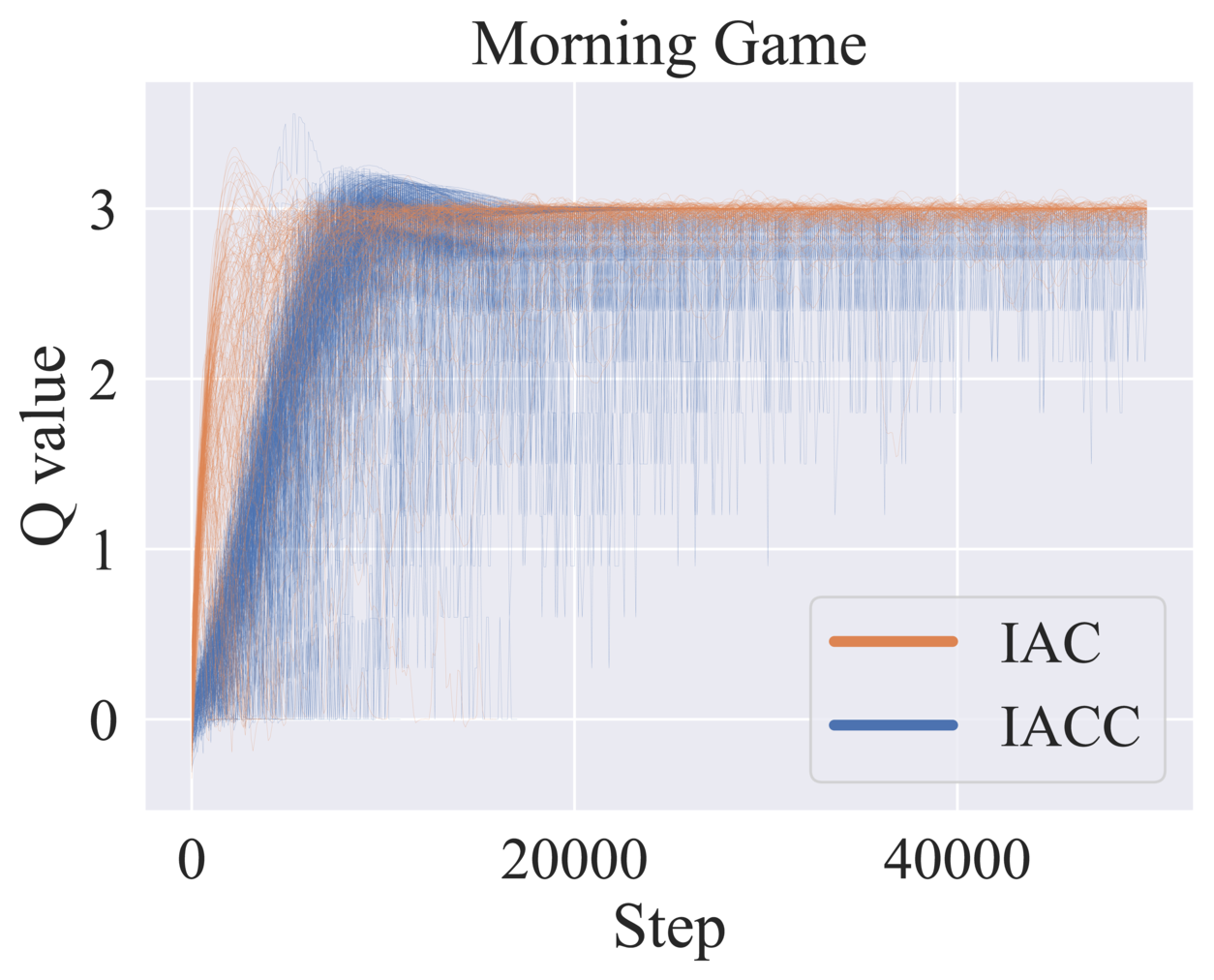}
  \caption{Morning Game Q-values used for updates of $\pi(a_2)$.
  Same as Figure~\ref{figure:morning_game_value_agg} but plotted with individual runs.}
  \label{figure:morning_game_value_individual}
\end{figure}

\section{Guess Game}\label{guess_game_environment}

We consider a simple one-step environment where two agents are trying to guess each other’s binary observations: $\Omega^i = \{o^i_1,\ o^i_2\}$, with action space $ \mathcal{A}^i = \{a^i_1,\ a^i_2,\ a^i_3\}$. Every timestep, the state $\langle s^1 \ s^2 \rangle$ is uniformly sampled ($\mathcal{T} = \langle U(\{s^1_1,s^1_2\}), U(\{s^2_1 , s^2_2\}) \rangle$) and is observed directly by the respective agents (\(o^i = s^i\)).
Agents are rewarded $(+10)$ if both their actions match the other agents’ observations (for $a^i_n, s^j_m$, $n=m$), 0 for one mismatch, and penalized ($-10$) for two mismatches.
On the other hand, $\langle a_3,a_3 \rangle$ gives a deterministic, low reward ($+5$) no matter the state.

\begin{figure}[ht!]
  \centering
  \includegraphics[width=0.3\textwidth]{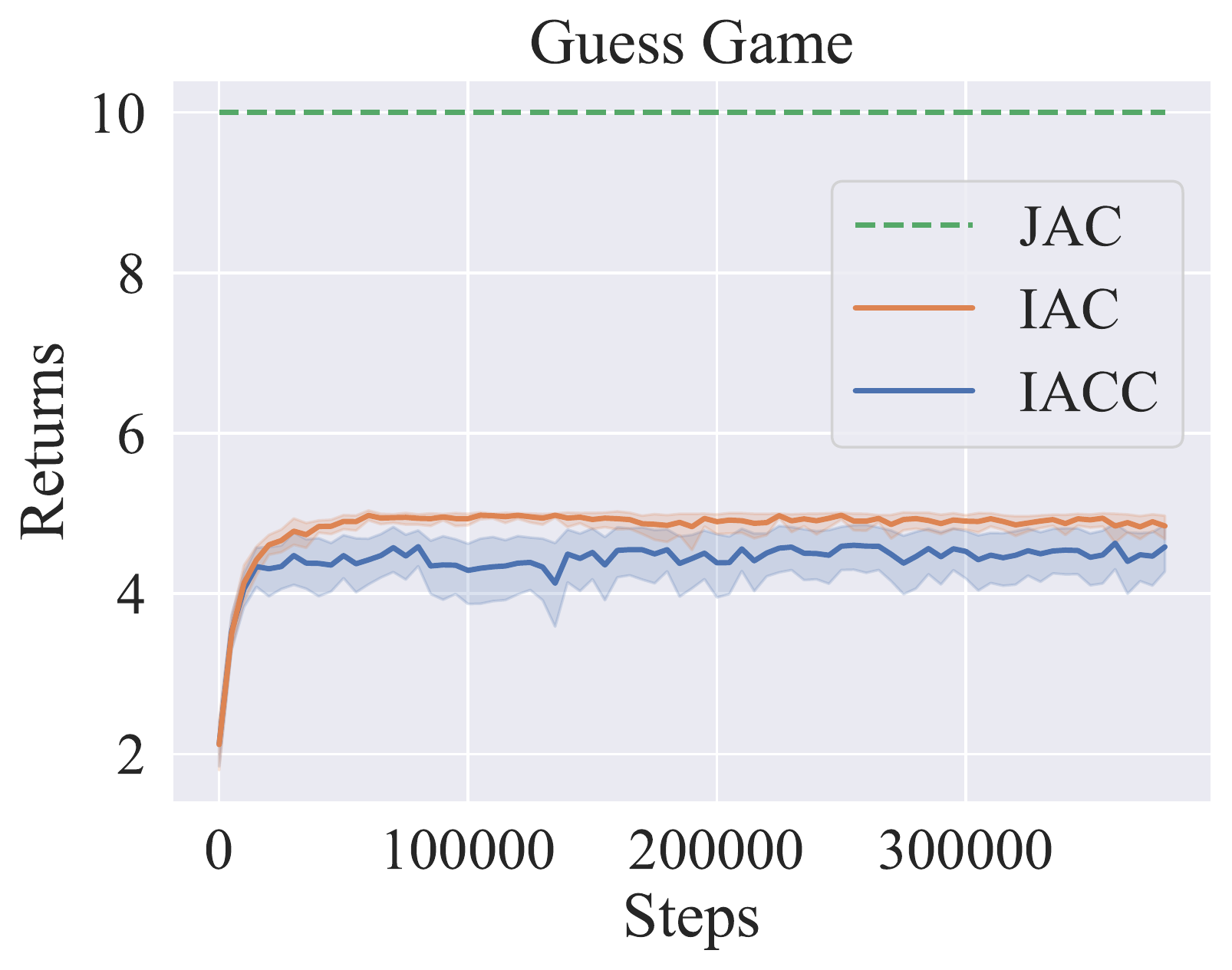}
  \caption{Performance comparison in Guess Game, showing centralized critic cannot bias the actors towards the global optimum in the simplest situation.}
  \label{matrix_return_with_jac}
\end{figure}

\begin{figure}[ht!]
  \centering
  \centering
  \subcaptionbox{Same data as shown in Figure~\ref{matrix_return_with_jac}}
      [0.31\linewidth]{\includegraphics[height=3.5cm]{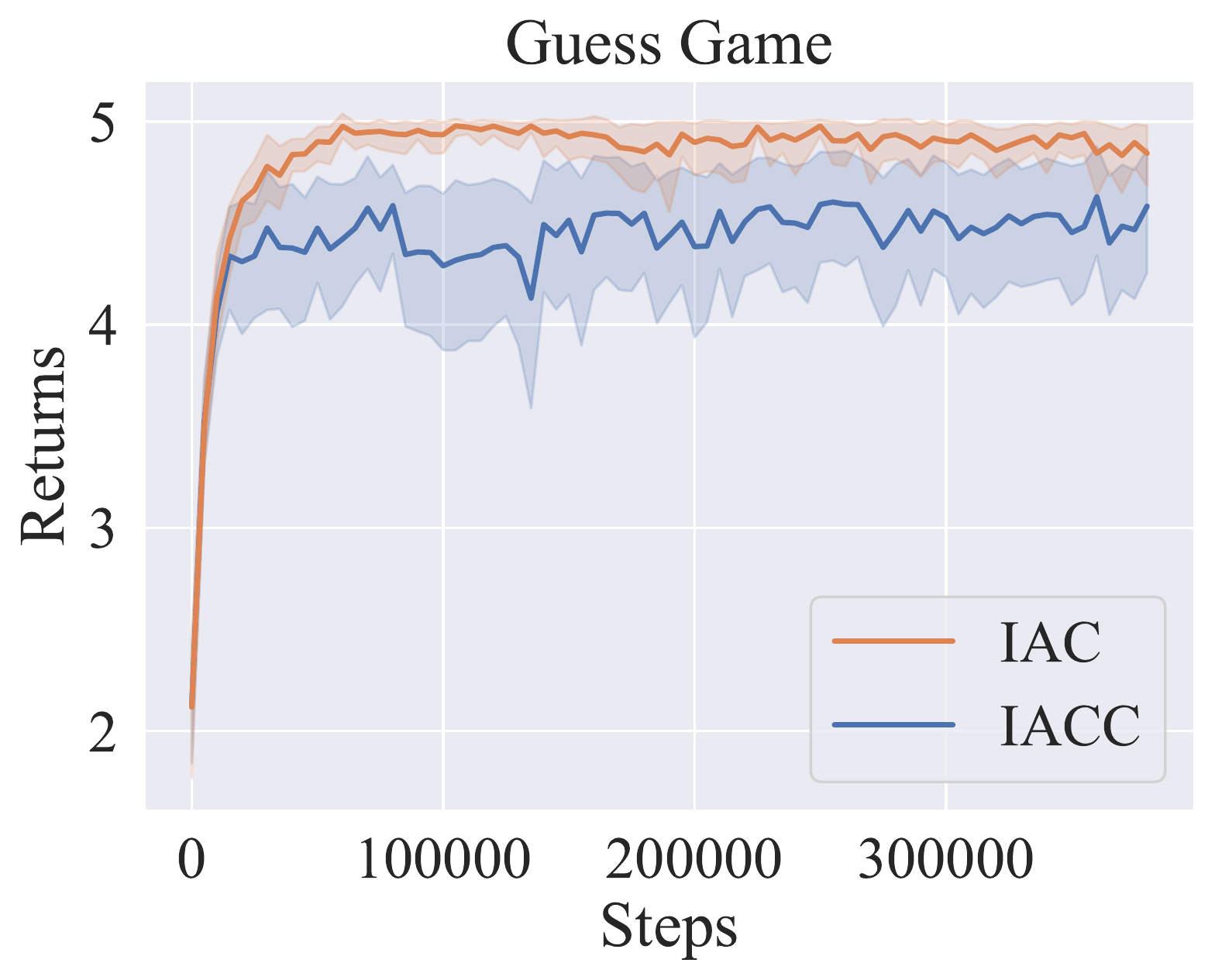}}
  ~
  \centering
  \subcaptionbox{$+100$ and $-100$ rewards\label{guess_game_100}}
      [0.31\linewidth]{\includegraphics[height=3.5cm]{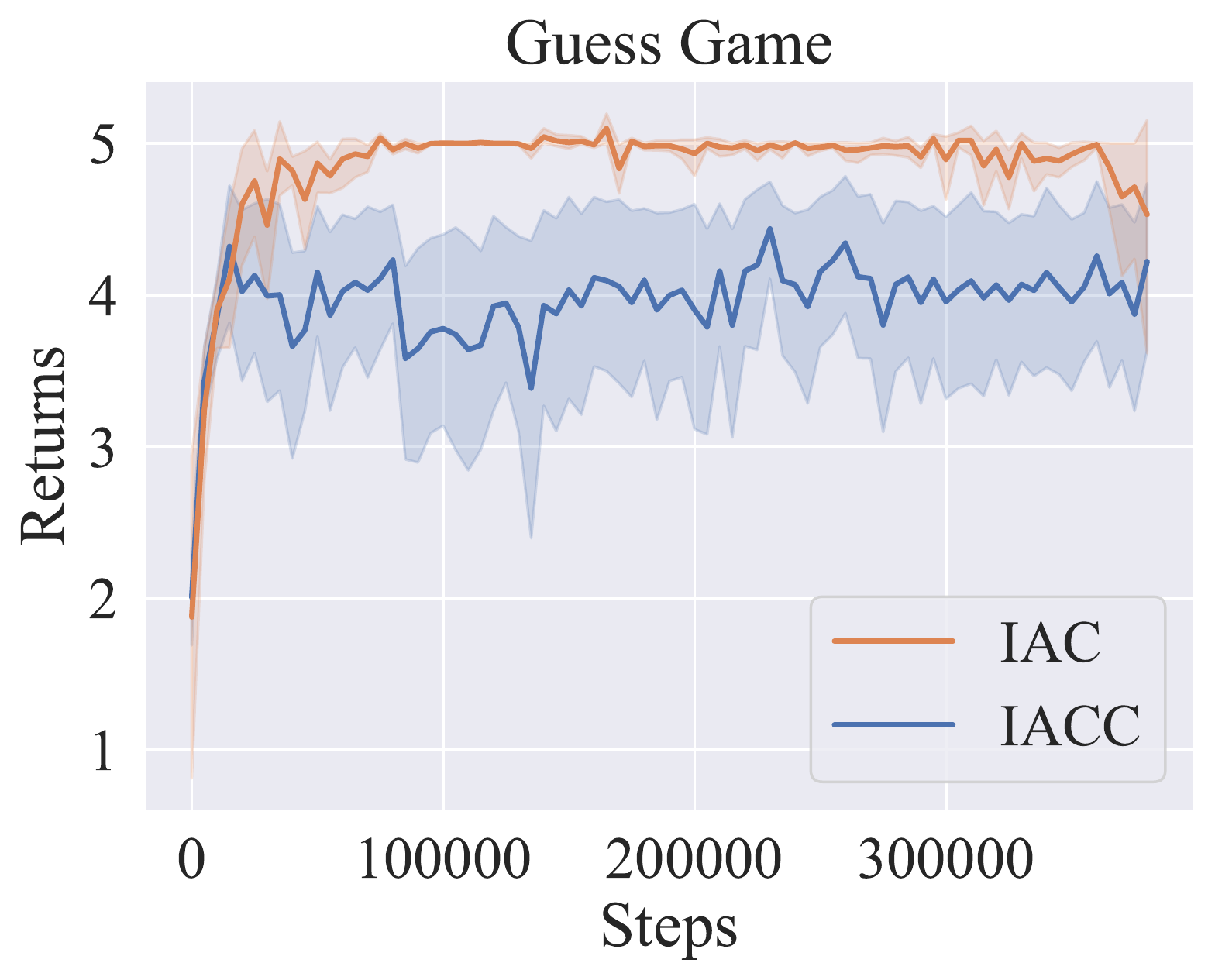}}
  ~
  \centering
  \subcaptionbox{Batch size increased from 64 to 512\label{guess_game_bs512}}
      [0.31\linewidth]{\includegraphics[height=3.5cm]{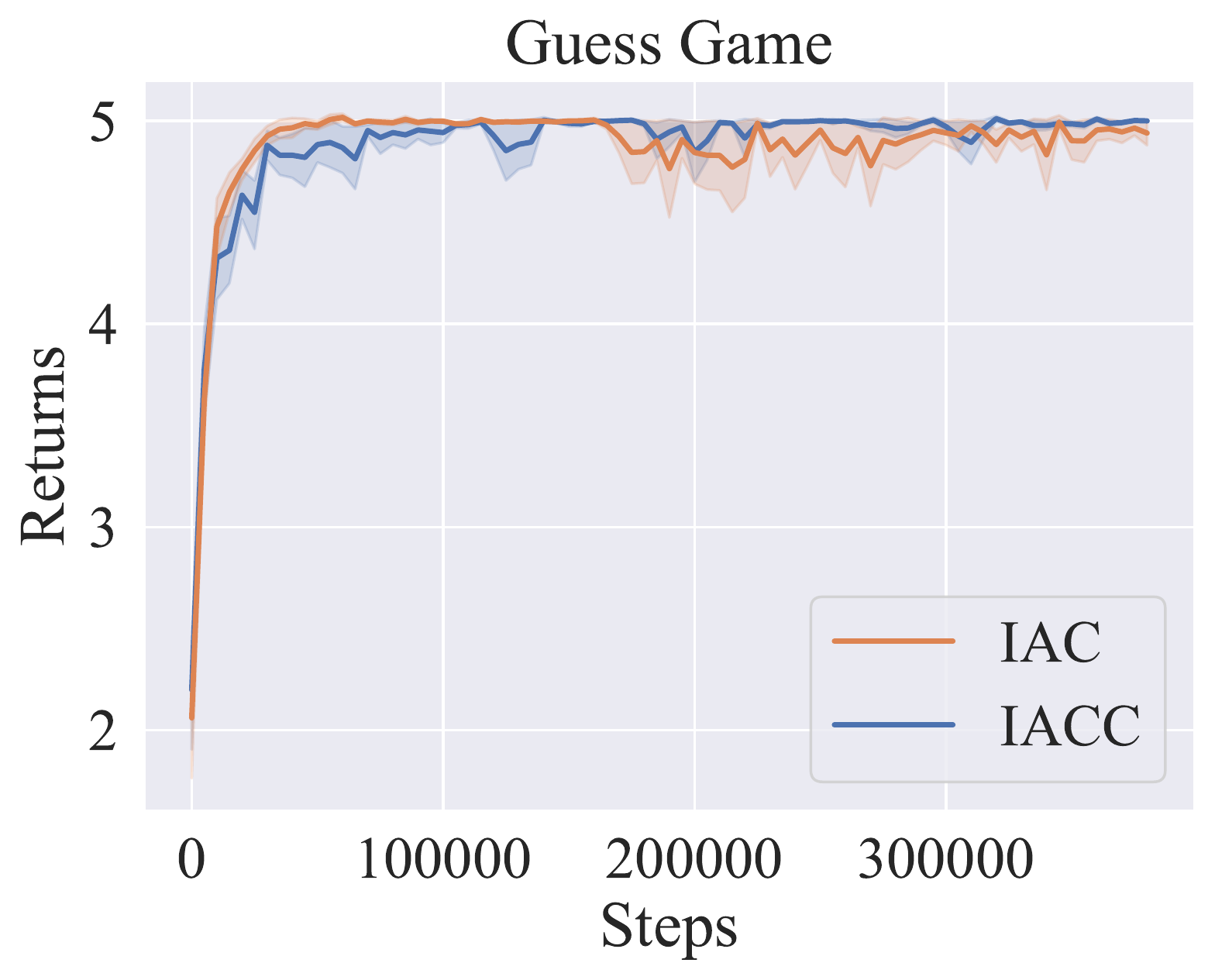}}
  \caption{Guess Game performance comparison under different settings.}
  \label{guess_game}
\end{figure}

We observe that the agents with decentralized critics can robustly converge to the optimal action $a_3$ with a return of 5, whereas the agents with a centralized critic do so with less stability.
Notice that empirical result shown in Figure~\ref{matrix_return_with_jac} the policy of JAC is not executable by either IAC or IACC due to their decentralized execution nature and an optimal decentralized set of policies only achieves a return of 5.
We also observe that although IACC is not biased, the lower performance is caused by larger variance in joint observations.
More specifically, the decentralized critic simply learned that \(a_1\) and \(a_2\) would average to an approximate return of 0 under both observations, while the explicit centralized critic estimates the two matching states (e.g $\{\langle o_1,o_2\rangle ,\langle a_1,a_2\rangle \}$) gives $+10$, one mismatch (e.g. $\{\langle o_1,o_1\rangle ,\langle a_1,a_2\rangle \}$) gives $0$ and two mismatches (e.g. $\{\langle o_2,o_2\rangle ,\langle a_1,a_1\rangle \}$) gives $-10$.
Because the observations are uniformly distributed, in the actual roll-outs, the probability of getting $+10$ is equal to that of $-10$, it does not effect the expected return (thus the expected gradient), but it does incur much higher variance in this case.
The variance is exacerbated if we use $\langle+100,-100\rangle$ instead of $\langle +10,-10\rangle$ (Figure~\ref{guess_game_100}); and can be mitigated using larger batch sizes (Figure~\ref{guess_game_bs512}).

\section{Grid World Environments}\label{grid_world_environments}

Grid World Environments from~\citep{shuo2019maenvs}, composed of variety of grid world tasks.
Readers are referred to the source code for more details.\footnote{
    https://github.com/Bigpig4396/Multi-Agent-Reinforcement-Learning-Environment
}

\subsection{Go Together}\label{appendix:go_together}
Two agents goes to a same goal for a reward of $10$, penalized for being too close or too far away.
The domain is fully observable, agents receive both agent's coordinates as observations.

\subsection{Find Treasure}\label{appendix:find_treasure}
Two rooms, two agents initialized in one room and try to navigate to the other room for a treasure ($+100$).
A door can be opened by having one agent stepping on a "switch" while the other agent goes through the door. 

\subsection{Cleaner}\label{appendix:cleaner}
\begin{figure}[ht!]
  \centering
  \captionsetup[subfigure]{labelformat=empty}
  \centering
  \subcaptionbox{}
      [0.31\linewidth]{\includegraphics[height=3.5cm]{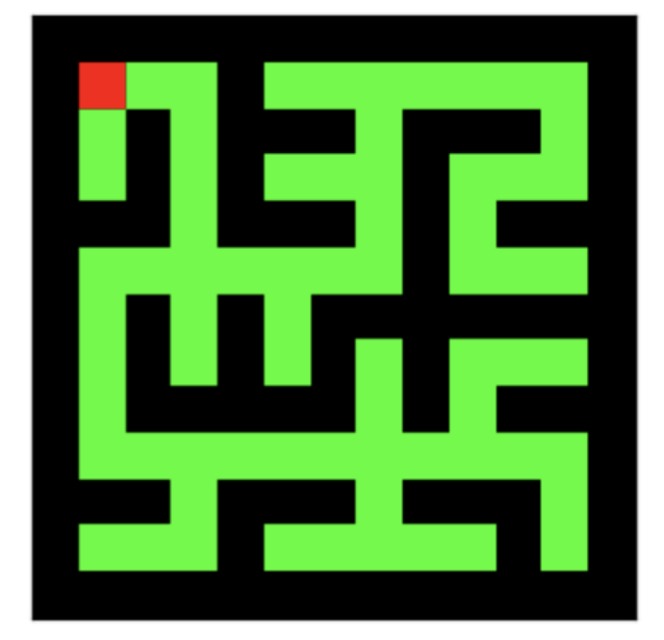}}
  ~
  \centering
  \subcaptionbox{}
      [0.31\linewidth]{\includegraphics[height=3.5cm]{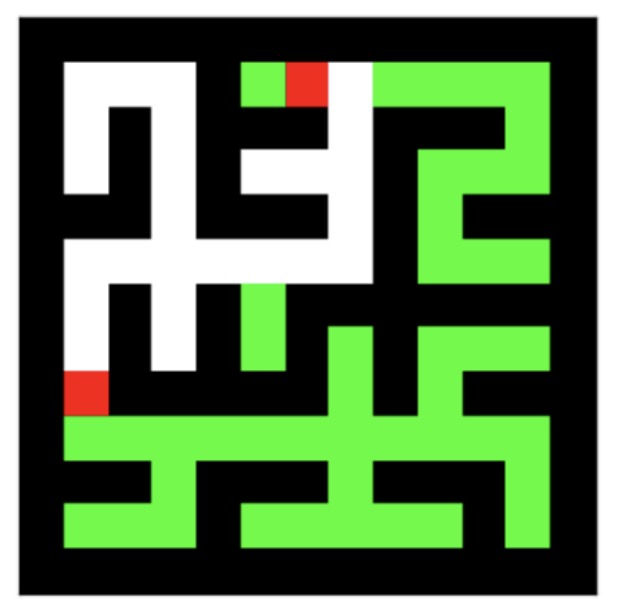}}
  ~
  \centering
  \subcaptionbox{}
      [0.31\linewidth]{\includegraphics[height=3.5cm]{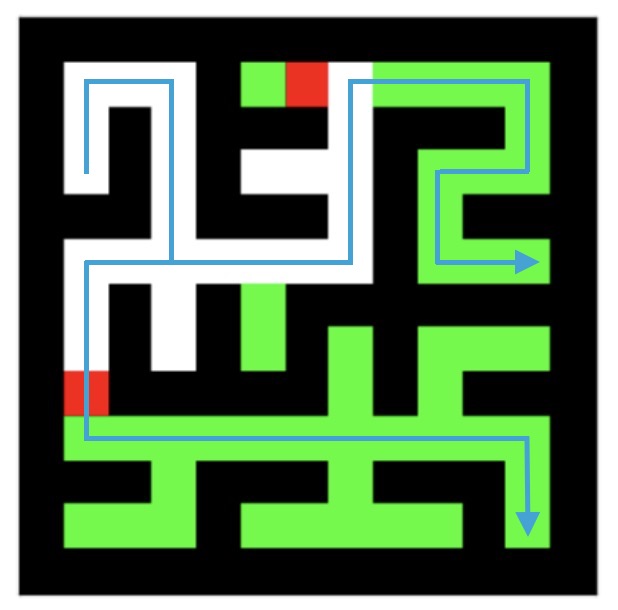}}

  \caption{Grid word visualization for Cleaner domain. Agent shown in red, uncleaned locations shown in green. Blue line shows the general two paths taken by an optimal policy.}
  \label{figure:cleaner_results}
\end{figure}

Two agents initialized in a maze, in which every untouched ground gives a reward of $+1$;
an optimal policy would seek to navigate to as many cells as possible while avoiding teammate's path.
Results are shown in Figure~\ref{figure:cleaner_results}.

\subsection{Move Box}\label{appendix:move_box}
Two agents collectively carry a heavy box (requires two agents to carry) to a destination, Two destinations are present; the nearer destination gives $+10$ and the farther destination gives $+100$.
\begin{figure}[ht!]
  \centering
  \captionsetup[subfigure]{labelformat=empty}
  \centering
  \subcaptionbox{}
      [0.31\linewidth]{\includegraphics[height=3.5cm]{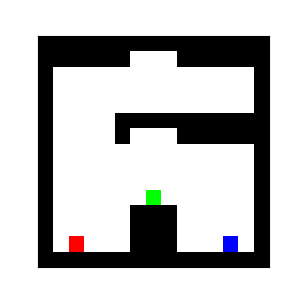}}
  ~
  \centering
  \subcaptionbox{}
      [0.31\linewidth]{\includegraphics[height=3.5cm]{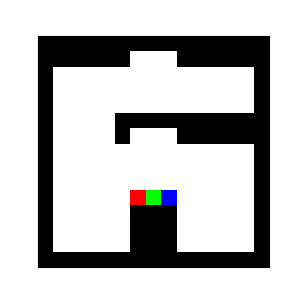}}
  ~
  \centering
  \subcaptionbox{}
      [0.31\linewidth]{\includegraphics[height=3.5cm]{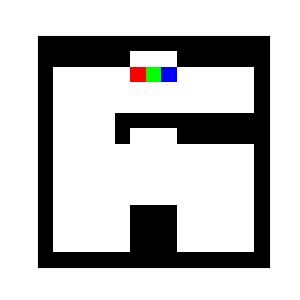}}
  \caption{Grid word visualization for Move Box domain. Agent shown in red and blue, box shown in green. }
\end{figure}

\begin{figure}[ht!]
  \centering
    \captionsetup[subfigure]{labelformat=empty}
  \centering
  \subcaptionbox{Go Together domain}
      [0.31\linewidth]{\includegraphics[height=3.5cm]{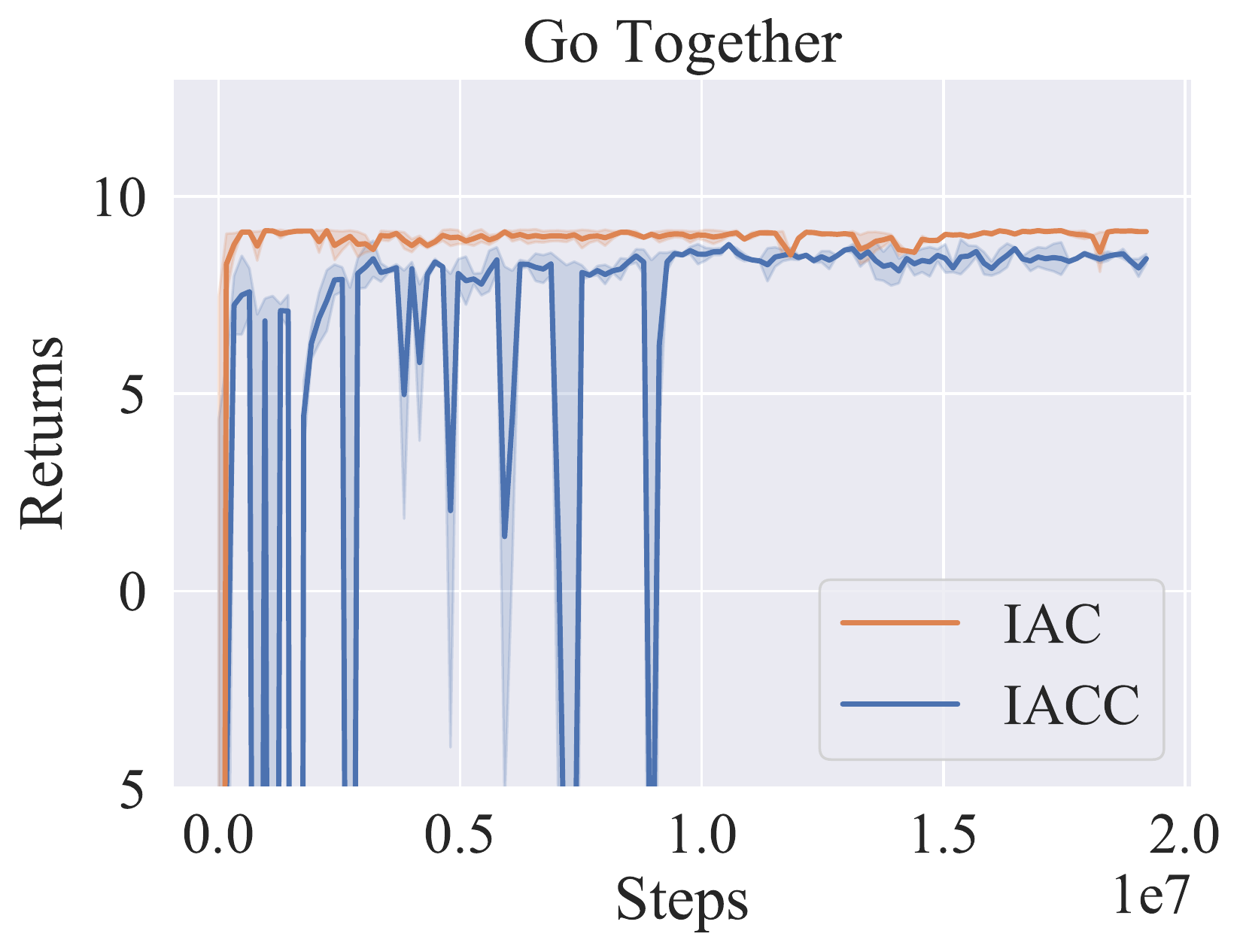}}
  ~
  \centering
  \subcaptionbox{Find Treasure domain}
      [0.31\linewidth]{\includegraphics[height=3.5cm]{figure/find_treasure.pdf}}
  ~
  \centering
  \subcaptionbox{Cleaner domain}
      [0.31\linewidth]{\includegraphics[height=3.5cm]{figure/cleaner.pdf}}
       
  \vspace{.2in}
  \centering
  \subcaptionbox{Move Box domain}
      [0.31\linewidth]{\includegraphics[height=3.5cm]{figure/move_box.pdf}}
  \caption{Performance comparison in various grid-world domains}
\end{figure}

\section{Cooperative Navigation}\label{cooperative_navigation_environment}
\begin{figure}[ht!]
  \centering
  \captionsetup[subfigure]{labelformat=empty}
  \centering
  \subcaptionbox{}
      [0.31\linewidth]{\includegraphics[height=3.5cm]{figure/cross.pdf}}
  ~
  \centering
  \subcaptionbox{}
      [0.31\linewidth]{\includegraphics[height=3.5cm]{figure/antipodal.pdf}}
  ~
  \centering
  \subcaptionbox{}
      [0.31\linewidth]{\includegraphics[height=3.5cm]{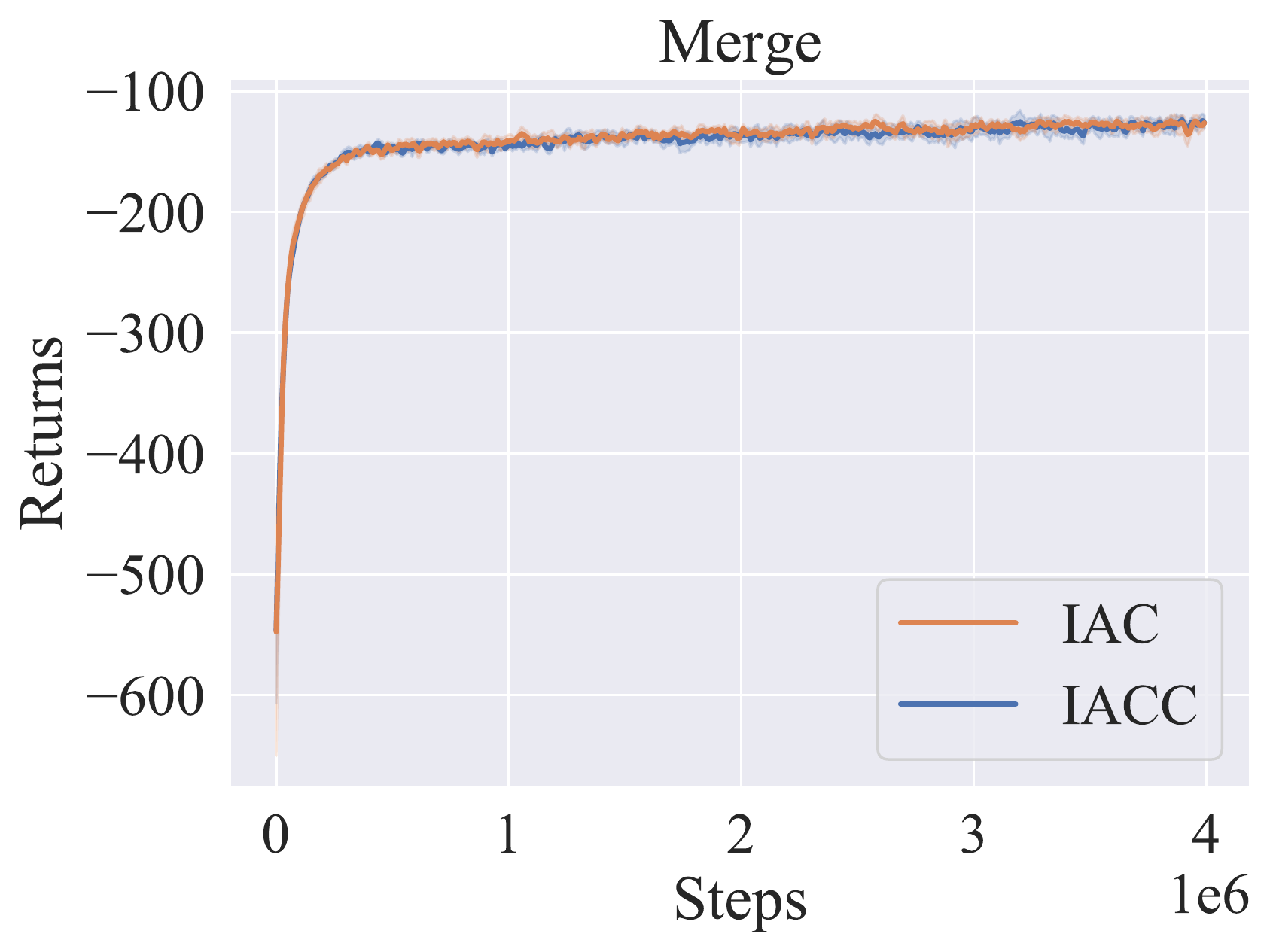}}

  \caption{Performance comparison in cooperative navigation domains. In these domains, agents navigate to designated target locations for reward, and are penalized for collisions.}
  \label{cooperative_navigation}
\end{figure}

Agents cooperatively navigate to target locations, agents are penalized for collisions.
The optimal routes (policies) are highly dependent on the other agents' routes (policies).
Results are shown in Figure~\ref{cooperative_navigation}

\vspace{2cm}

\clearpage
\section{Partially Observable Predator and Prey}\label{predator_and_prey_environment}

\begin{figure}[ht!]
  \centering
  \captionsetup[subfigure]{labelformat=empty}
  \centering
  \subcaptionbox{obs\_range 0.4}
      [0.31\linewidth]{\includegraphics[height=4cm]{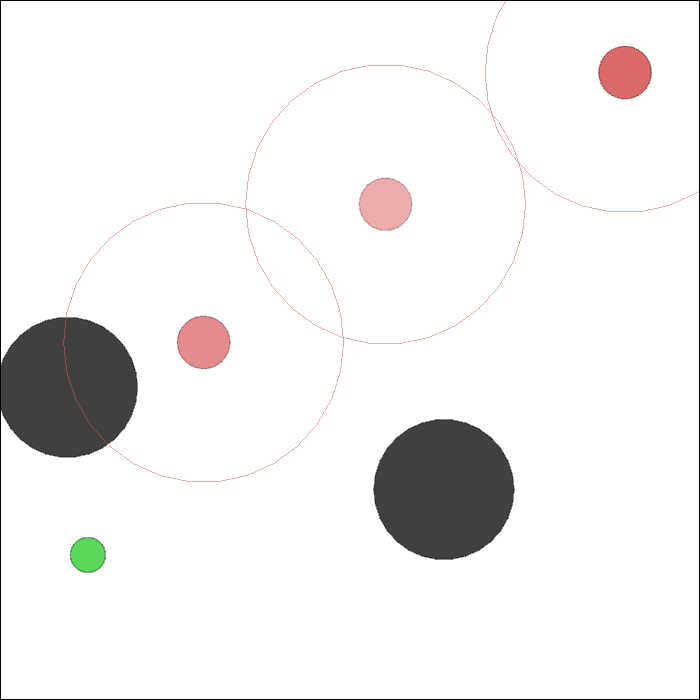}}
  ~
  \centering
  \subcaptionbox{obs\_range 0.6}
      [0.31\linewidth]{\includegraphics[height=4cm]{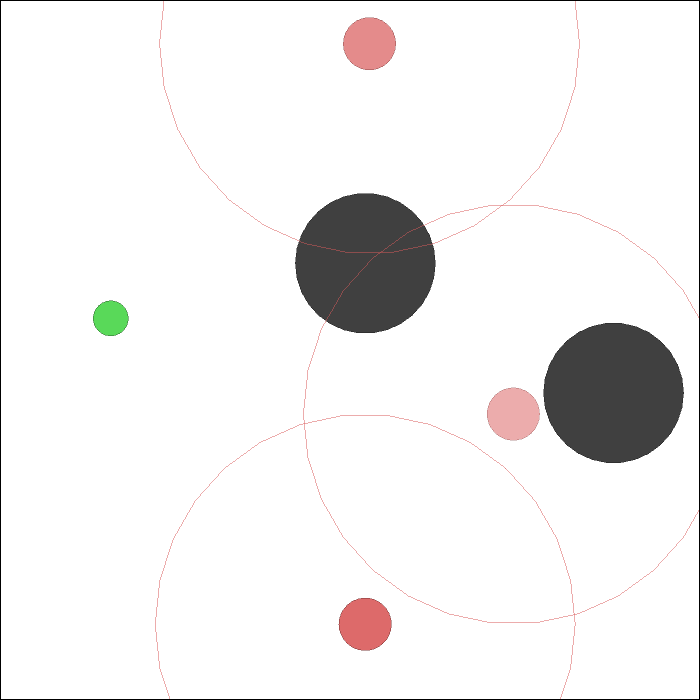}}
  ~
  \centering
  \subcaptionbox{obs\_range 0.8}
      [0.31\linewidth]{\includegraphics[height=4cm]{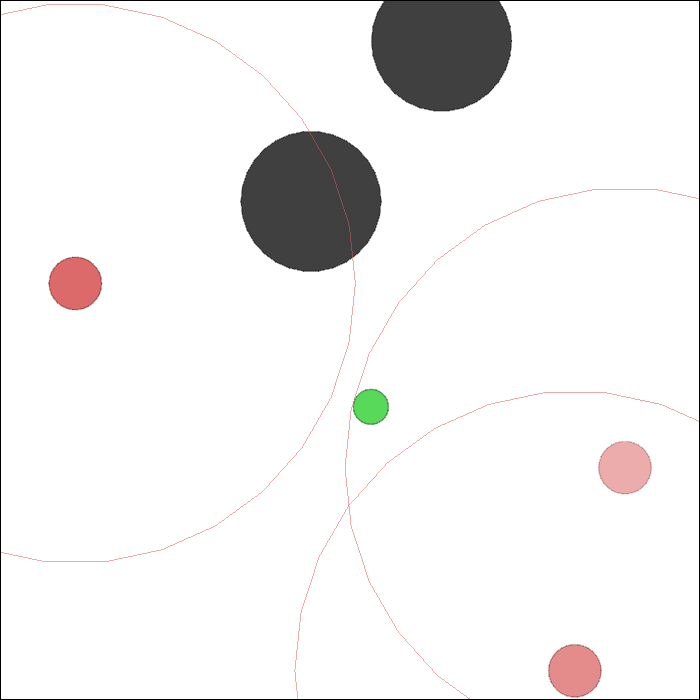}}

  \caption{Domain configurations under various observation ranges.}
  \label{pap_config}
\end{figure}

We conducted experiments in a variant version of the Predator and Prey domain (originally introduced in MADDPG~\cite{lowe2017multi} and widely used by later works), where three slower predators (red circle) must collaboratively capture a faster prey (green circle) in a random initialized 2-D plane with two landmarks.
The prey's movement is bounded within the $2\times2$ space shown in Fig.~\ref{pap_config}, and a hand-crafted policy is run by the prey to avoid the predators by moving towards a sampled position furthest to all predators.
Each predator's movement space, however is unbounded and its policy is trained via RL algorithms.
In order to achieve partial observability, we introduce an observation range that restricts each predator from detecting the relative positions and velocities of teammates and the prey and the relative position of each landmark out of a certain radius.
Each predator has 8 evenly discretized moving directions centered on itself.
A team reward of +10 is assigned when any one of the predators touches the prey.
The terminal condition of an episode is the maximum time-step, 25, is reached.  

\begin{figure}[ht!]
  \centering
  \captionsetup[subfigure]{labelformat=empty}
  \subcaptionbox{}
      [0.31\linewidth]{\includegraphics[height=3.5cm]{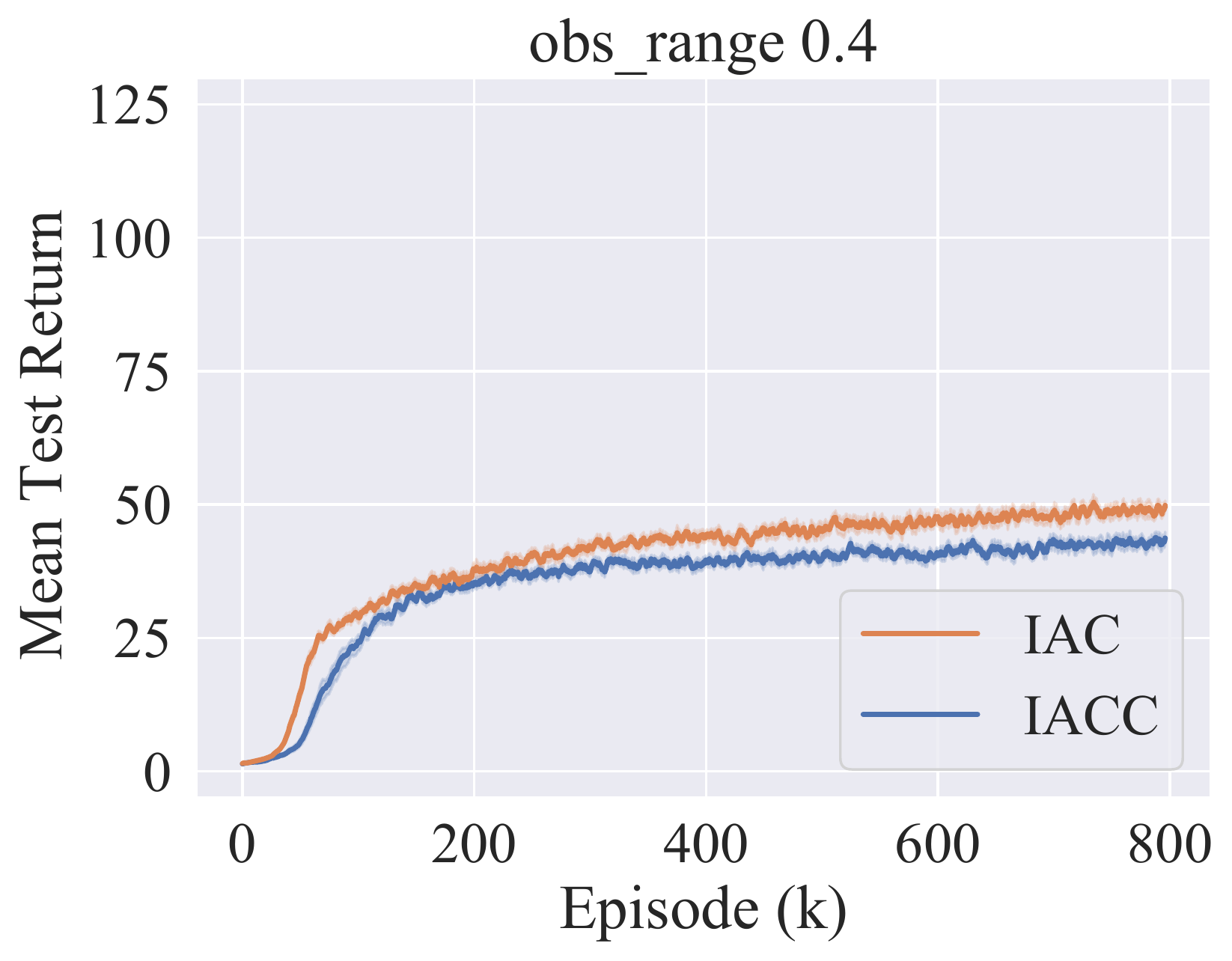}}
  ~
  \centering
  \subcaptionbox{}
      [0.31\linewidth]{\includegraphics[height=3.5cm]{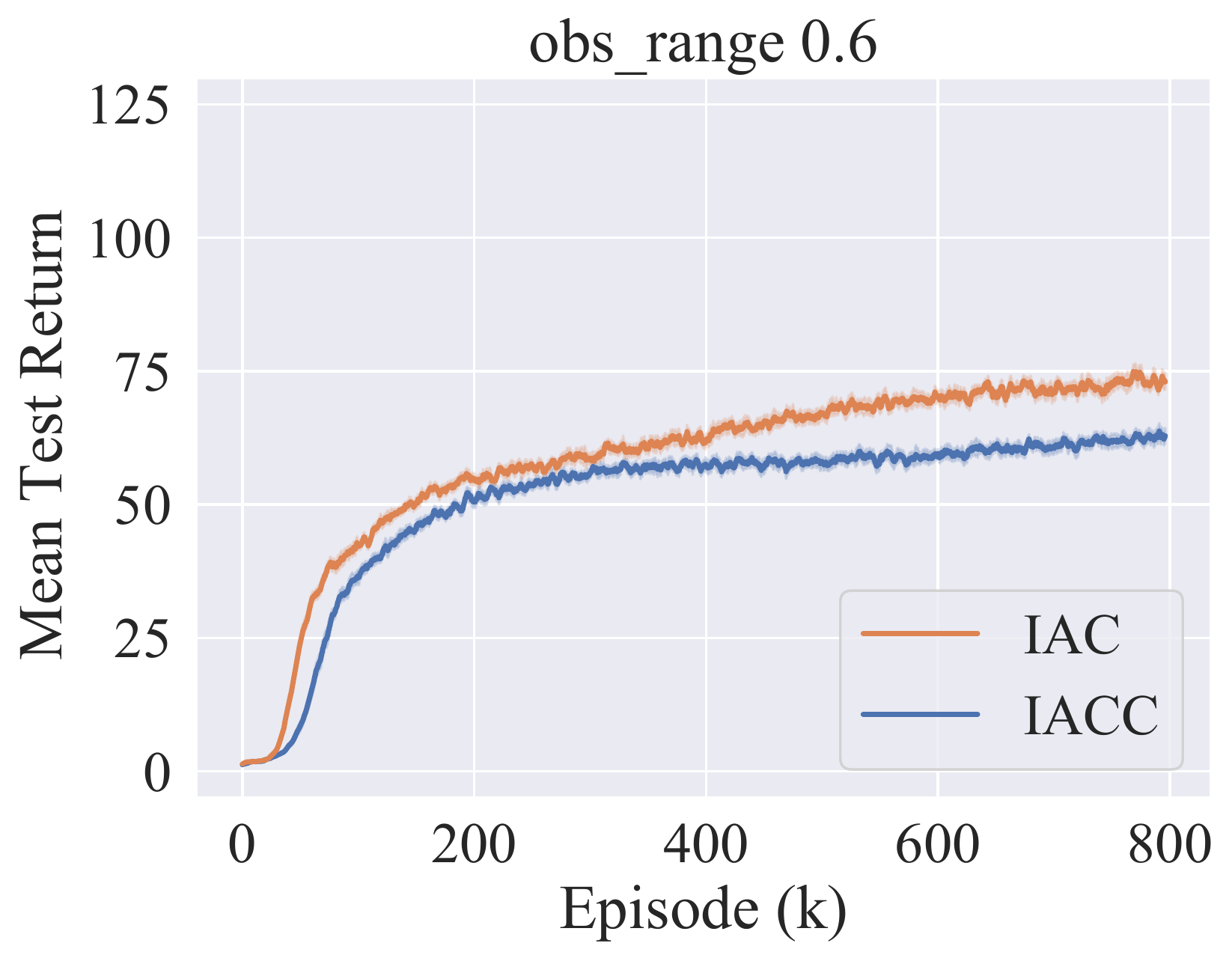}}
  ~
  \centering
  \subcaptionbox{}
      [0.31\linewidth]{\includegraphics[height=3.5cm]{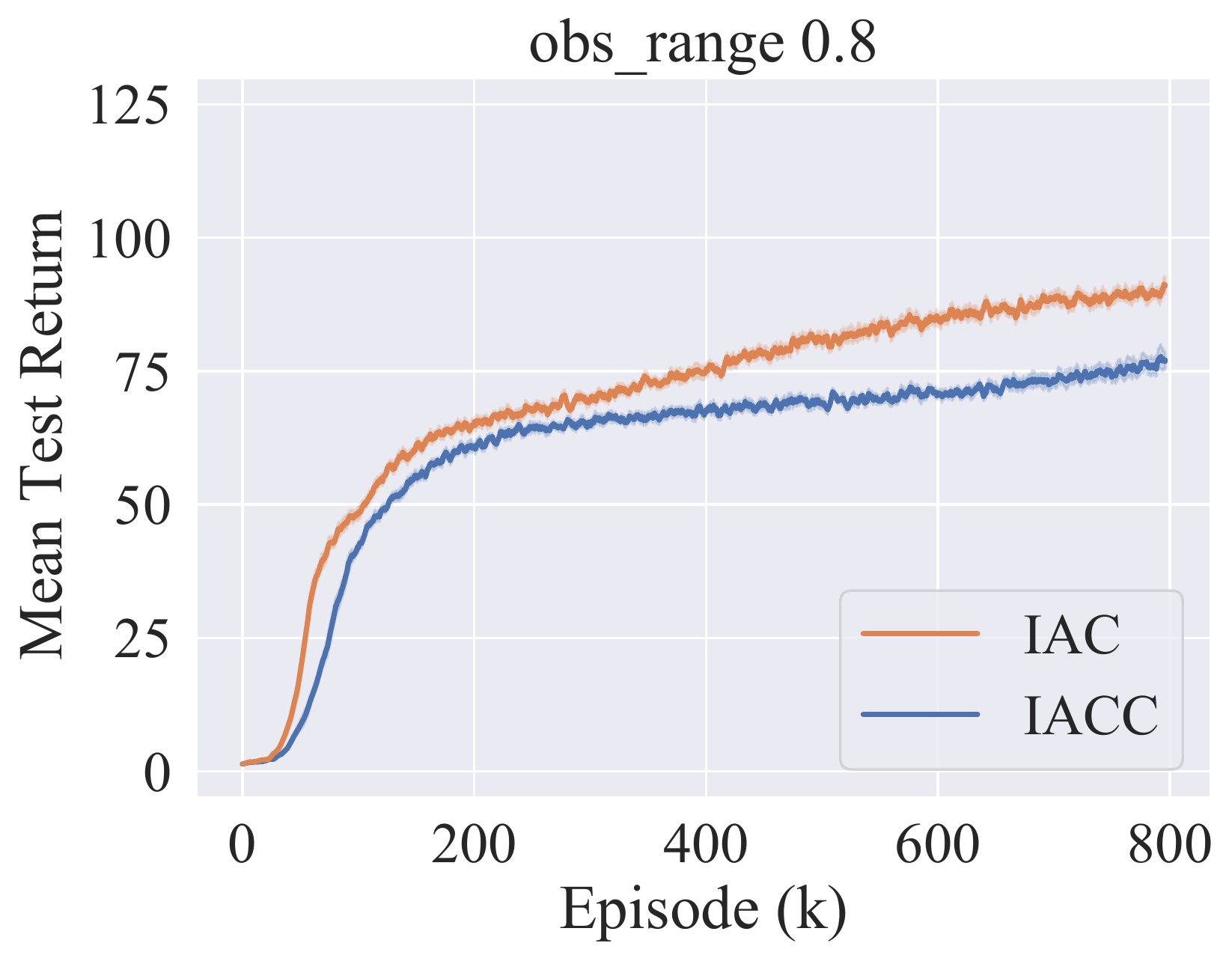}}
   
  \centering
  \subcaptionbox{}
      [0.31\linewidth]{\includegraphics[height=3.5cm]{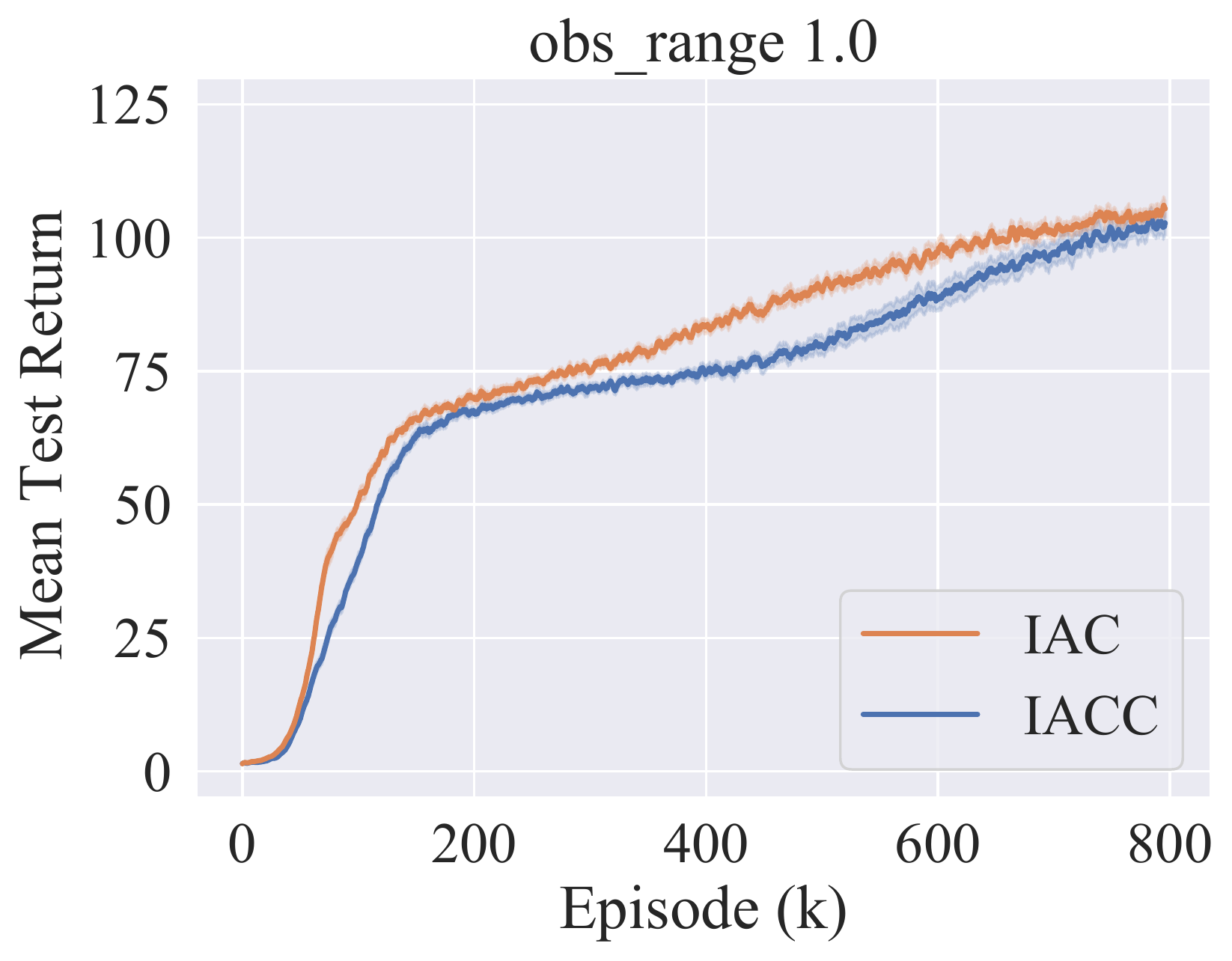}}
  ~
  \centering
  \subcaptionbox{}
      [0.31\linewidth]{\includegraphics[height=3.5cm]{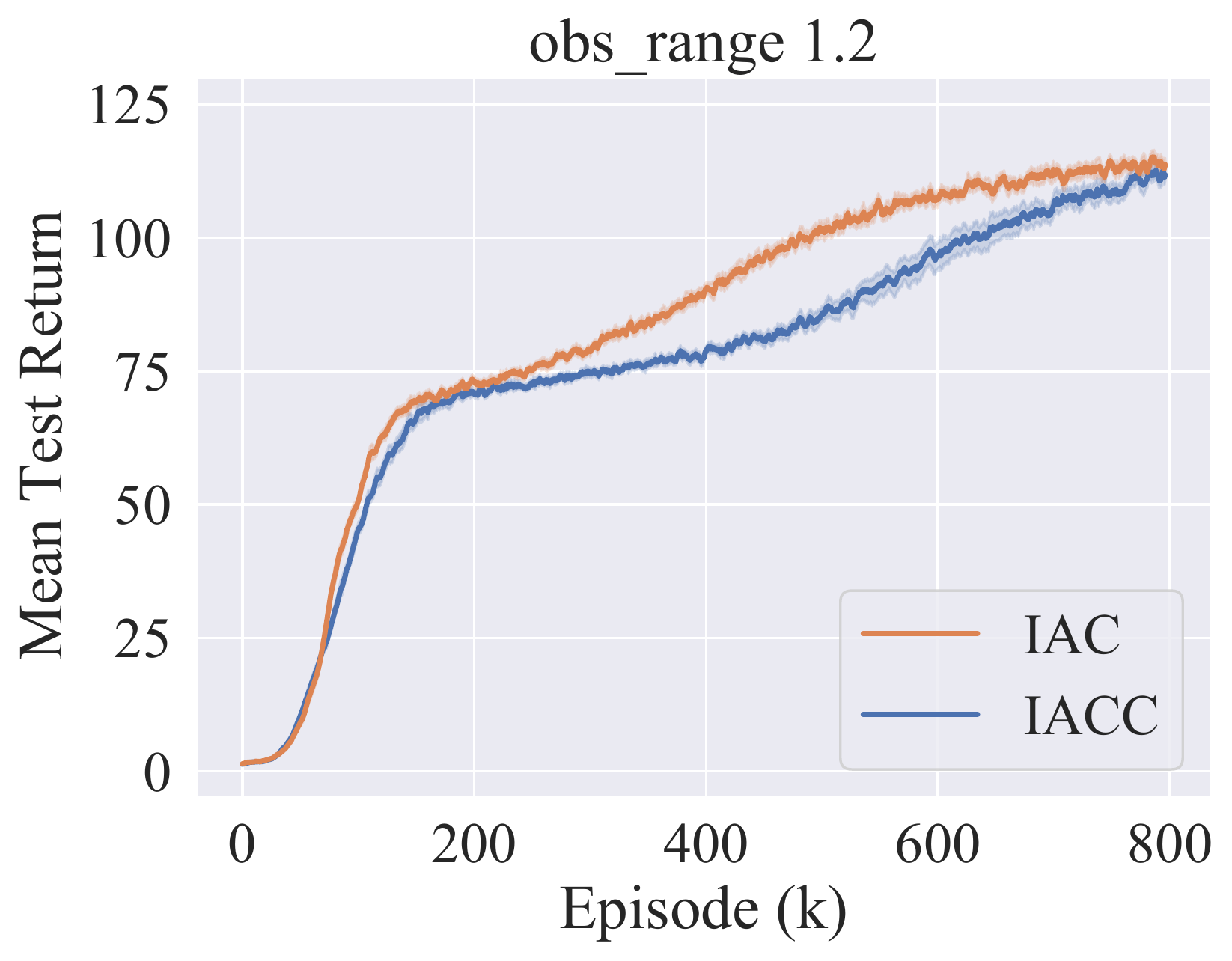}}
  \caption{Performance comparison on Predator-and-Prey under various observation ranges.}
\end{figure}

\clearpage

\section{Capture Target}\label{capture_target_env}

Capture Target is another widely used domain \cite{amato2009incremental,xiao_corl_2019,omidshafiei2017deep,lyu2020likelihood}, in which two agents (green and blue circles) move in a $m \times m$ toroidal grid world to capture a moving target (red cross) simultaneously.
The positions of the agents and the target are randomly initialized at the beginning of each episode.
At each time step, the target keeps moving right without any transition noise, while the agents have five moving options (\emph{up, down, left, right} and \emph{stay}) with 0.1 probability of accidentally arriving any one of the adjacent cells.
Each agent is allowed to always observe its own location but the target's is blurred with a probability 0.3.
Agents can receive a terminal reward $+1.0$ only when they locate at a same cell with the target.
The horizon of this problem is 60 timesteps.

\begin{figure}[ht!]
  \centering
  \captionsetup[subfigure]{labelformat=empty}
  \centering
  \subcaptionbox{8 x 8}
      [0.31\linewidth]{\includegraphics[height=2.5cm]{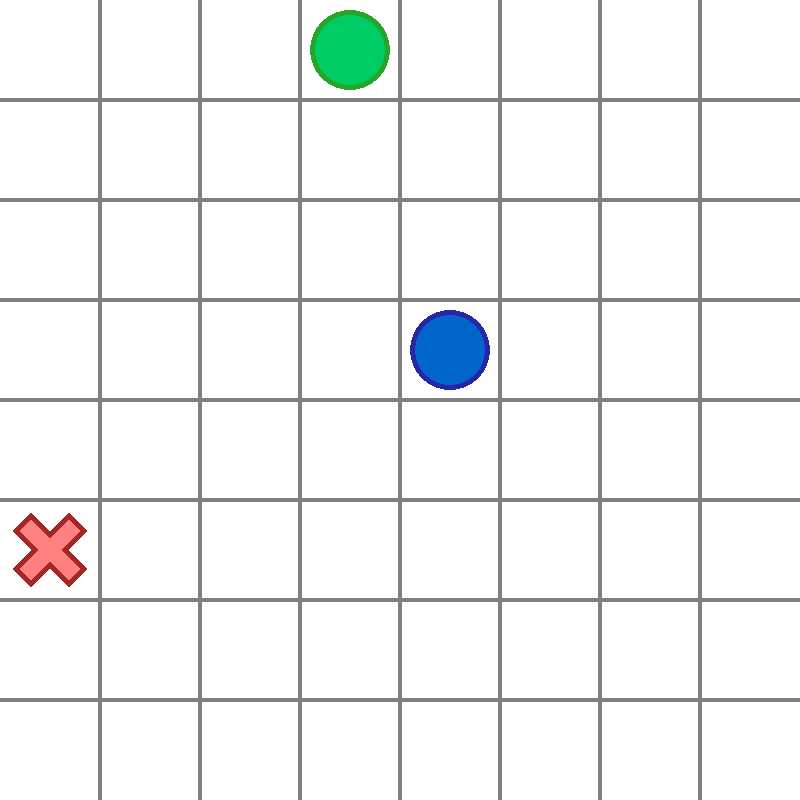}}
  ~
  \centering
  \subcaptionbox{10 x 10}
      [0.31\linewidth]{\includegraphics[height=3.25cm]{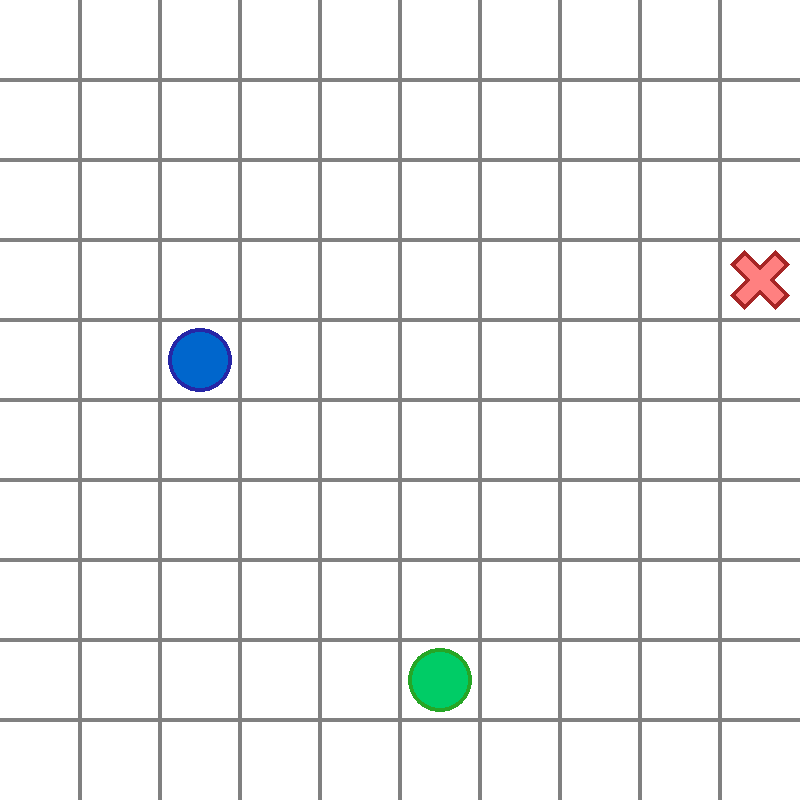}}
  ~
  \centering
  \subcaptionbox{12 x 12}
      [0.31\linewidth]{\includegraphics[height=4cm]{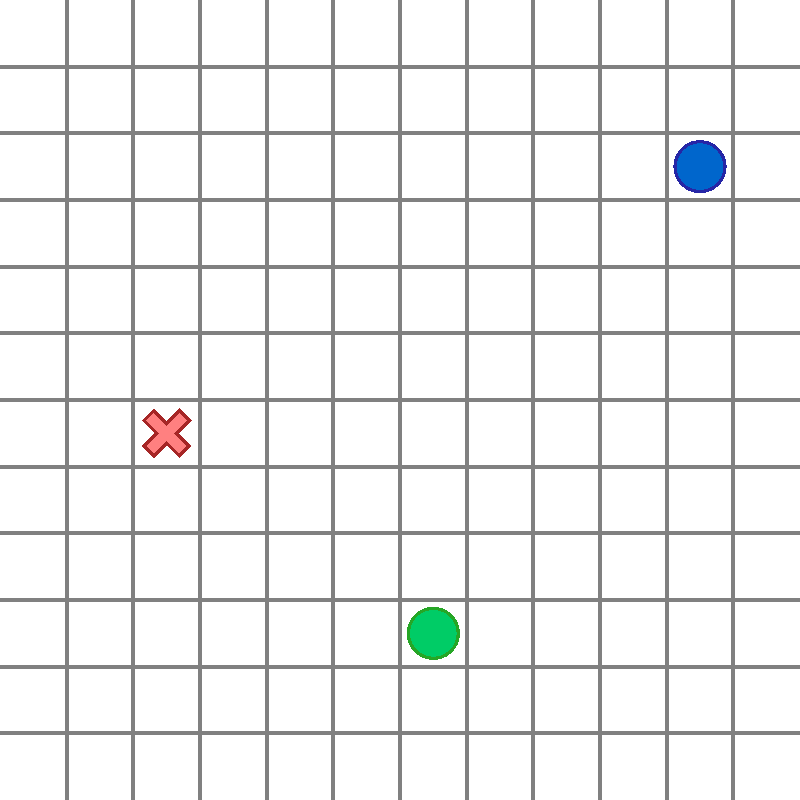}}
  \caption{Domain configurations under various grid world sizes.}
  \label{ct_config}
\end{figure}

\begin{figure}[ht!]
  \centering
  \captionsetup[subfigure]{labelformat=empty}
  \subcaptionbox{}
      [0.31\linewidth]{\includegraphics[height=3.5cm]{figure/ct4x4.pdf}}
  ~
  \centering
  \subcaptionbox{}
      [0.31\linewidth]{\includegraphics[height=3.5cm]{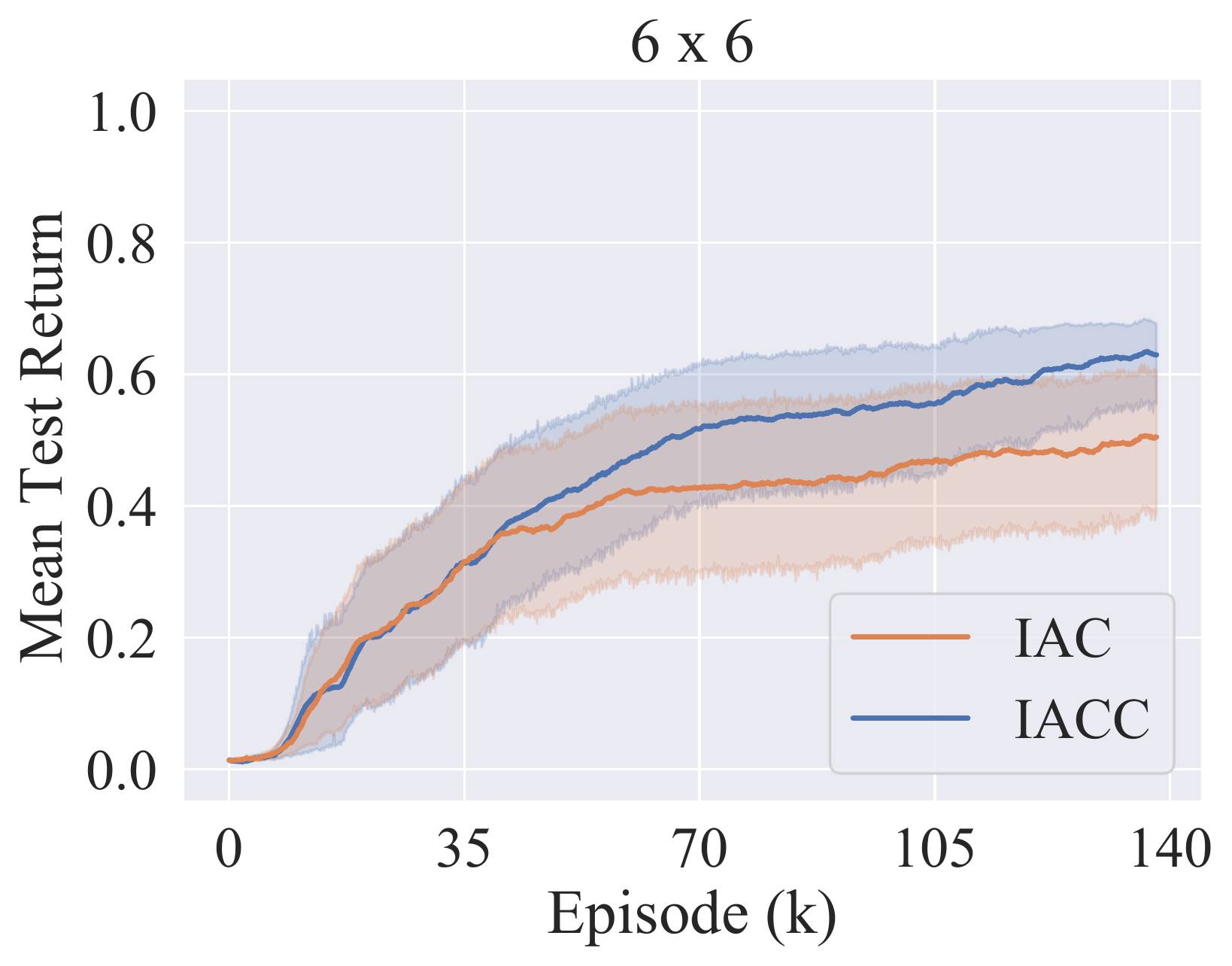}}
  ~
  \centering
  \subcaptionbox{}
      [0.31\linewidth]{\includegraphics[height=3.5cm]{figure/ct8x8.pdf}}
   
  \centering
  \subcaptionbox{}
      [0.31\linewidth]{\includegraphics[height=3.5cm]{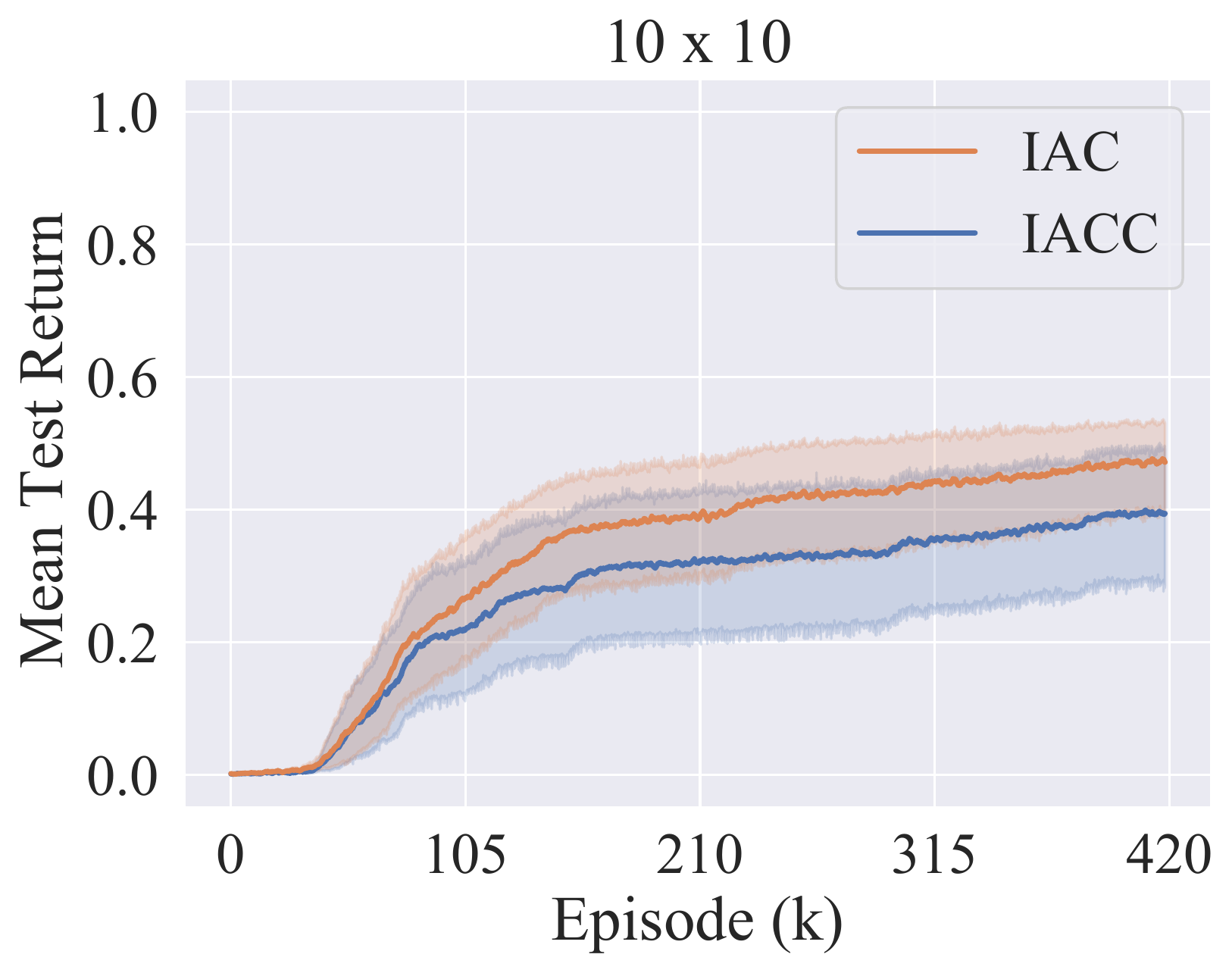}}
  ~
  \centering
  \subcaptionbox{}
      [0.31\linewidth]{\includegraphics[height=3.5cm]{figure/ct12x12.pdf}}

  \caption{Comparison on Capture Target under various grid world sizes.}
  \label{CT}
\end{figure}

\clearpage
\section{Small Box Pushing}\label{small_box_pushing_environment}

The objective of two agents, in this domain, is to push two small boxes to the goal (yellow) area at the top of the grid world with a discount factor $\gamma=0.98$.
When any one of the small boxes is pushed to the goal area, the team receives a $+100$ reward and the episode terminates.
Each agent has four applicable actions:
\emph{move forward, turn left, turn right} and \emph{stay}.
Each small box moves forward one grid cell when any agent faces it and operates \emph{move forward} action.
The observation captured by each agent consists of only the status of the one front cell which can be empty, teammate, small box or boundary.
The ideal behavior is that two agents push two small boxes to the goal area concurrently to obtain the maximal discounted return.

\begin{figure}[ht!]
  \centering
  \captionsetup[subfigure]{labelformat=empty}
  \centering
  \subcaptionbox{8 x 8}
      [0.31\linewidth]{\includegraphics[height=2.5cm]{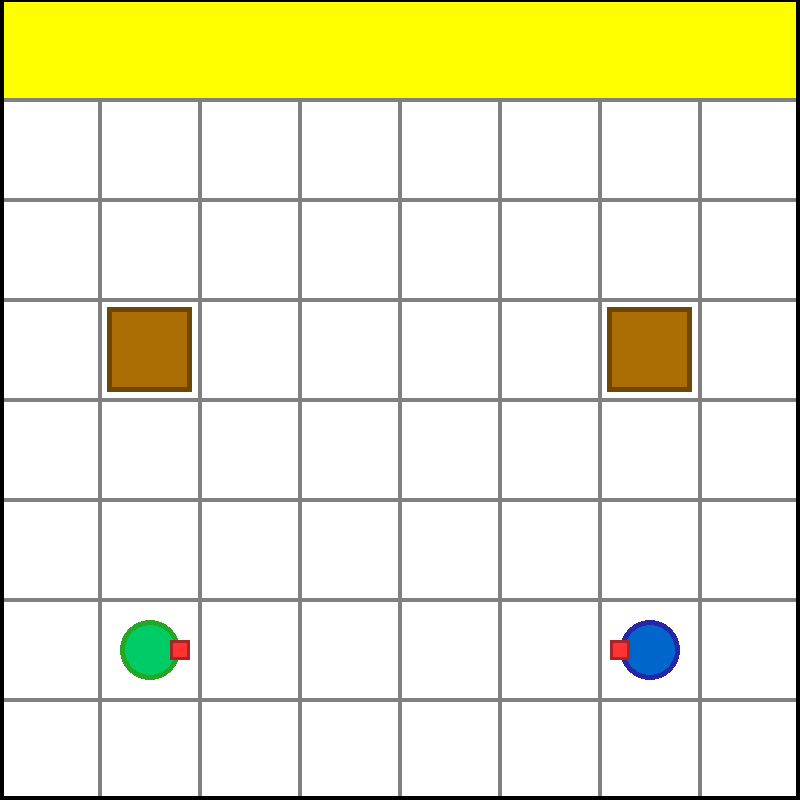}}
  ~
  \centering
  \subcaptionbox{10 x 10}
      [0.31\linewidth]{\includegraphics[height=3.25cm]{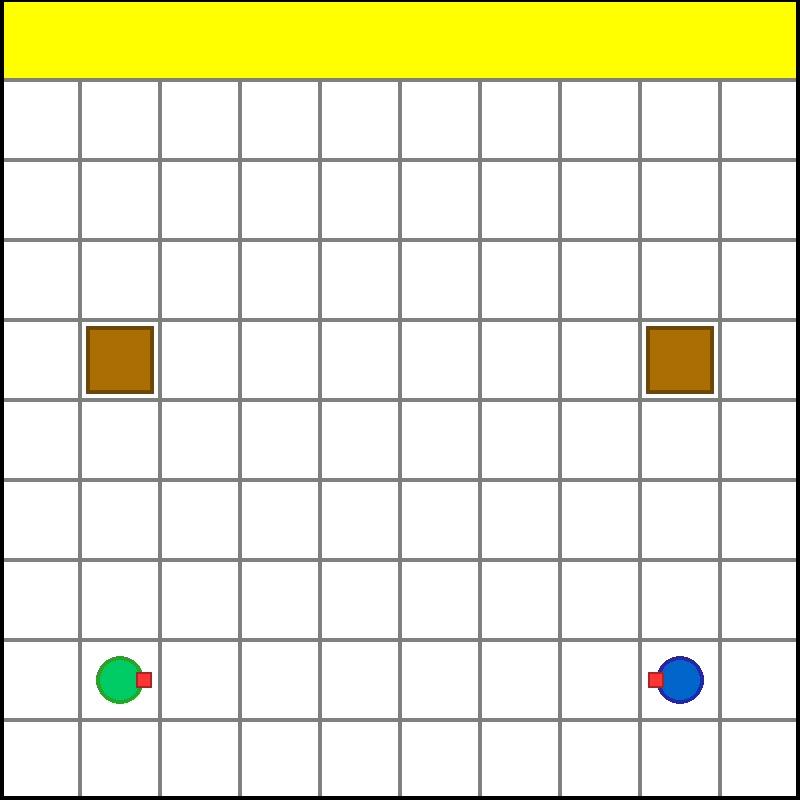}}
  ~
  \centering
  \subcaptionbox{12 x 12}
      [0.31\linewidth]{\includegraphics[height=4cm]{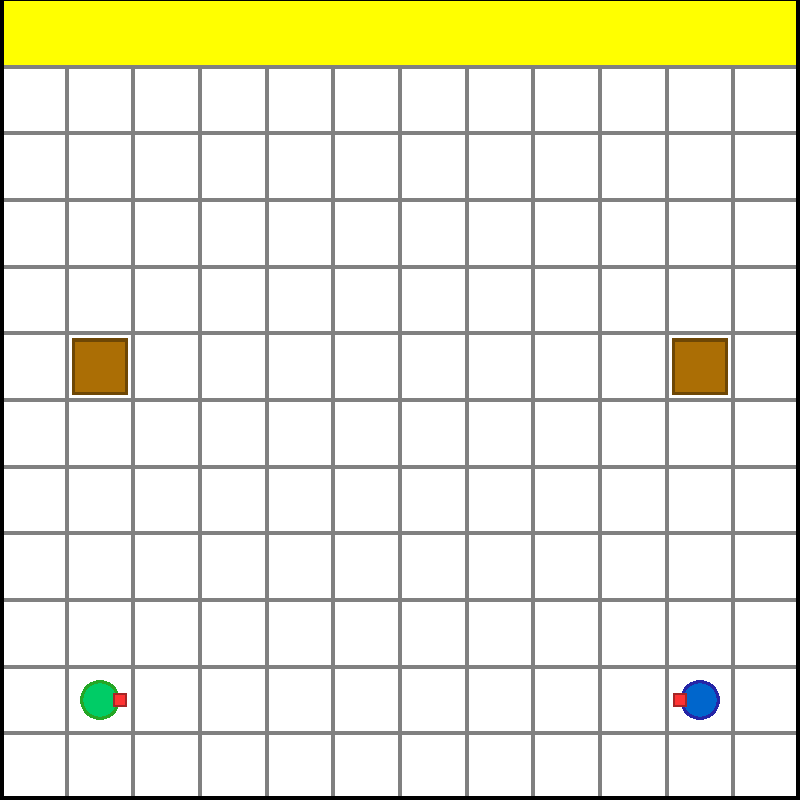}}

  \caption{Domain configurations under various grid world sizes.}
  \label{bp_config}
\end{figure}

\begin{figure}[ht!]
  \centering
  \captionsetup[subfigure]{labelformat=empty}
  \subcaptionbox{}
      [0.31\linewidth]{\includegraphics[height=3.5cm]{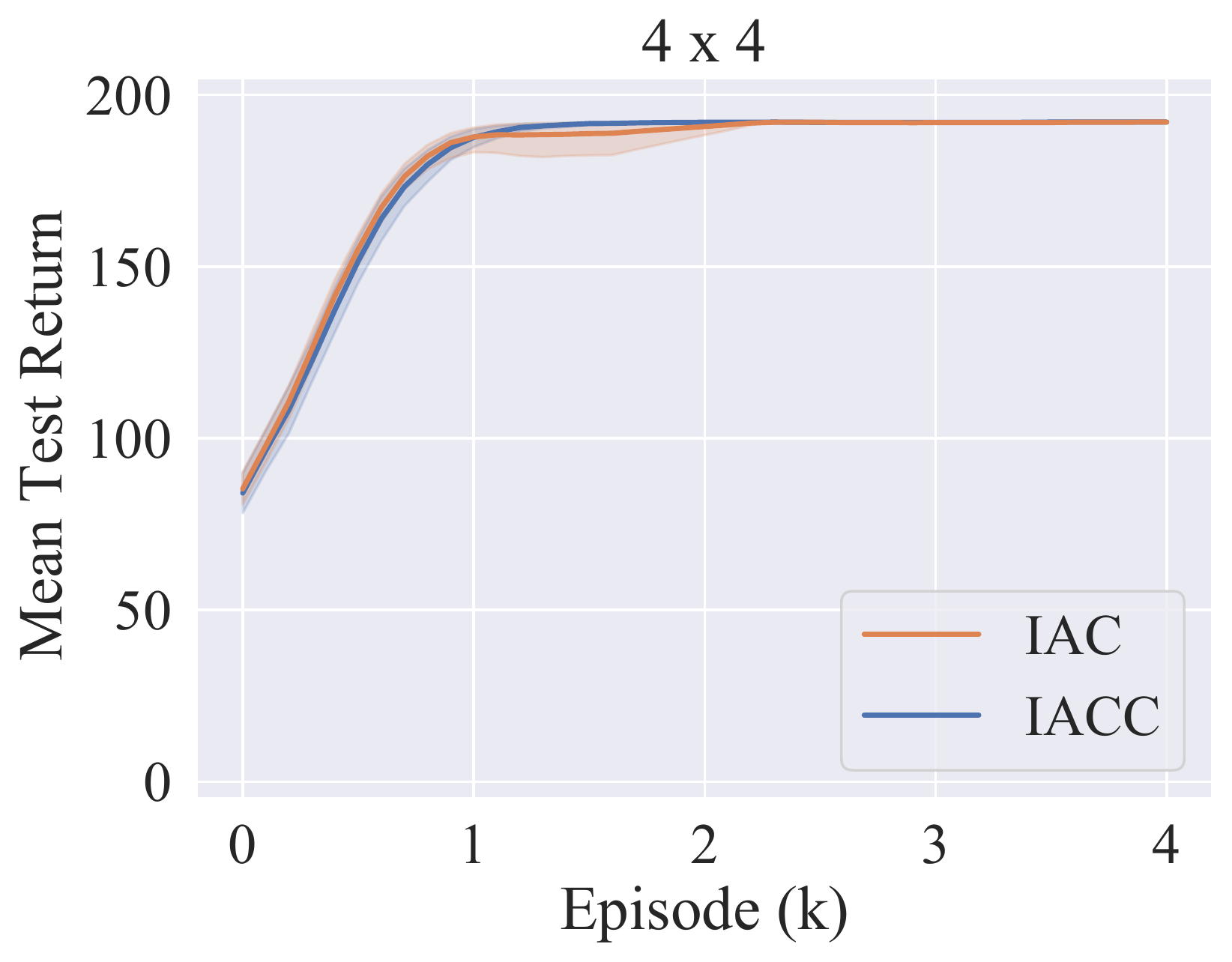}}
  ~
  \centering
  \subcaptionbox{}
      [0.31\linewidth]{\includegraphics[height=3.5cm]{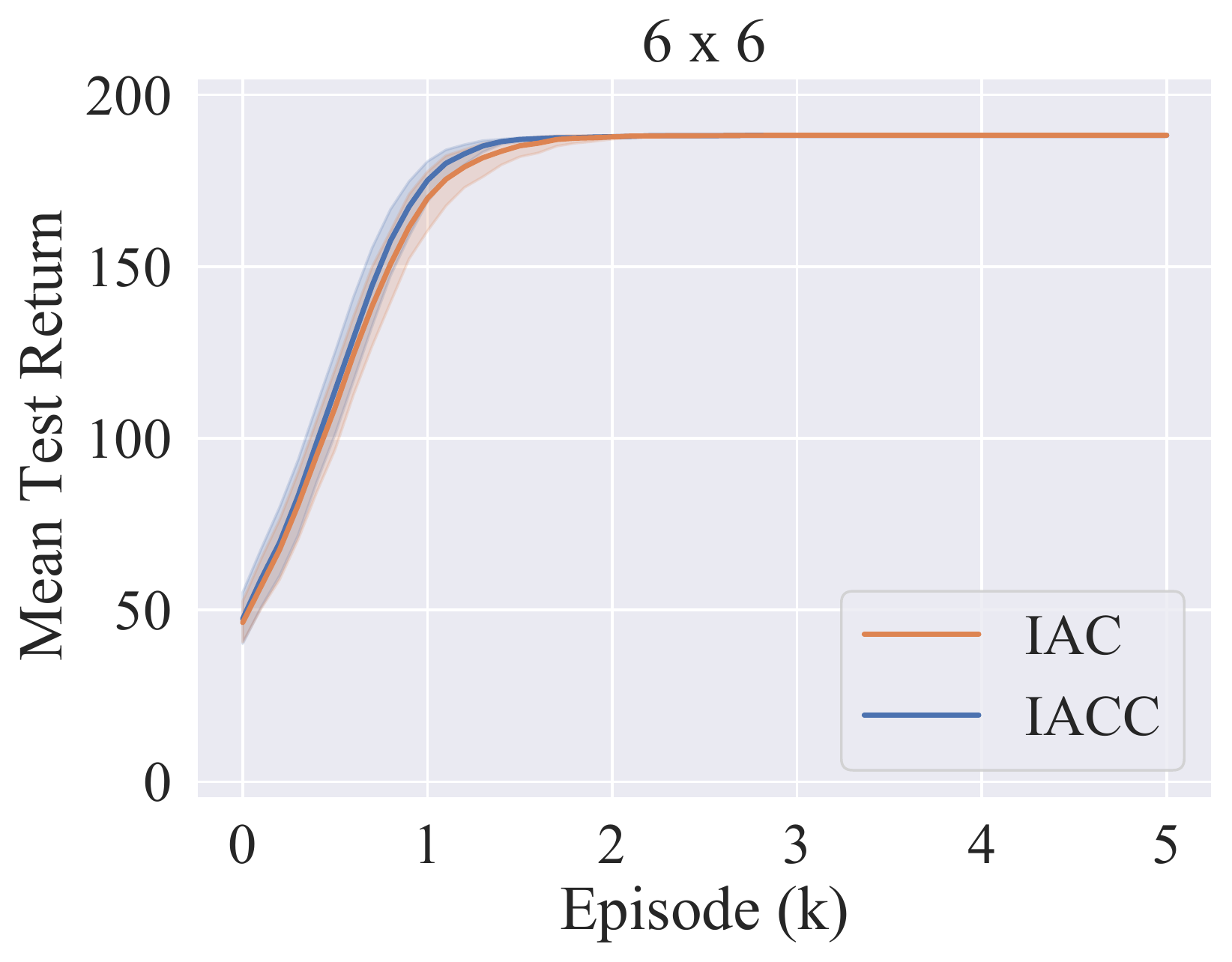}}
  ~
  \centering
  \subcaptionbox{}
      [0.31\linewidth]{\includegraphics[height=3.5cm]{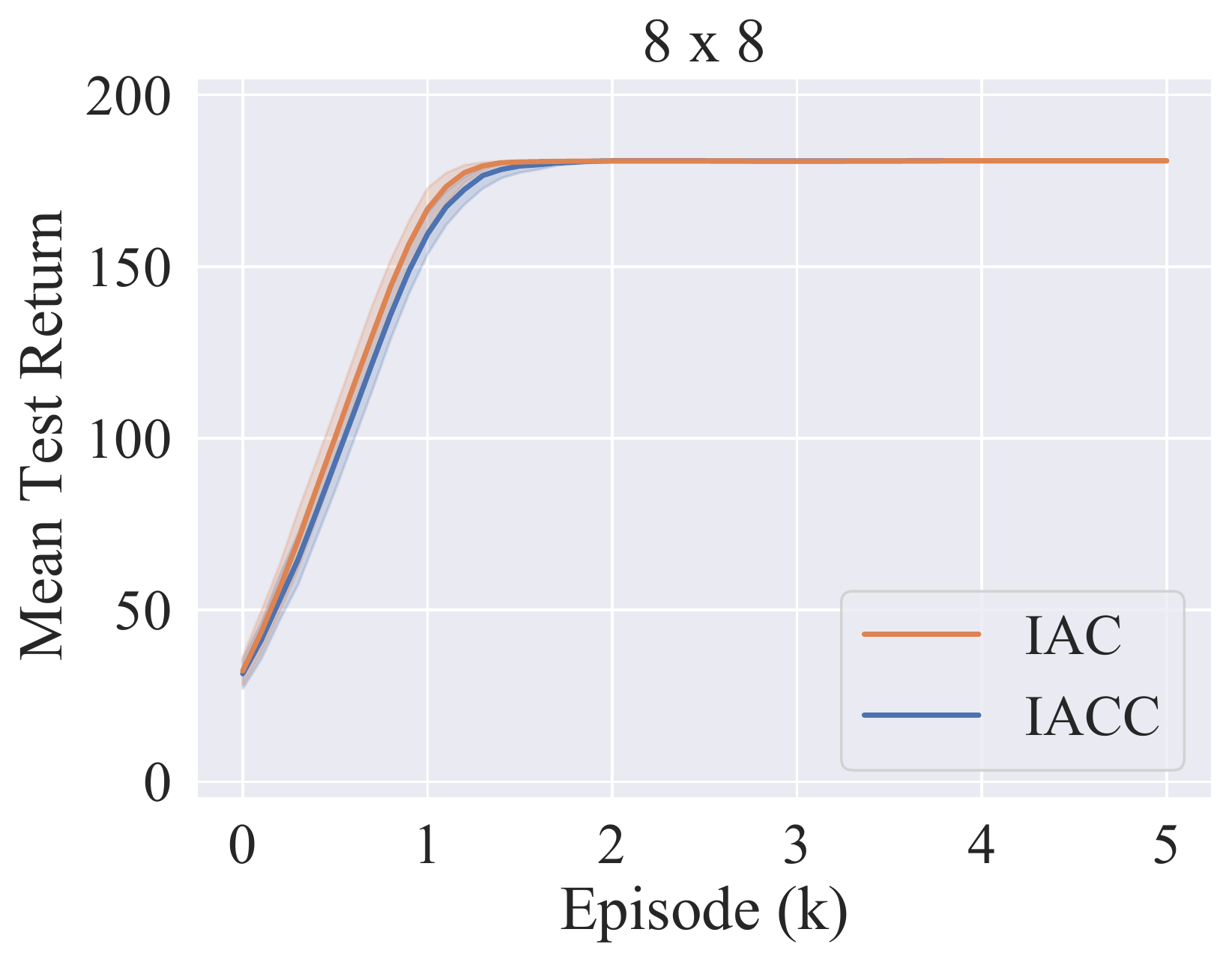}}
   
  \centering
  \subcaptionbox{}
      [0.31\linewidth]{\includegraphics[height=3.5cm]{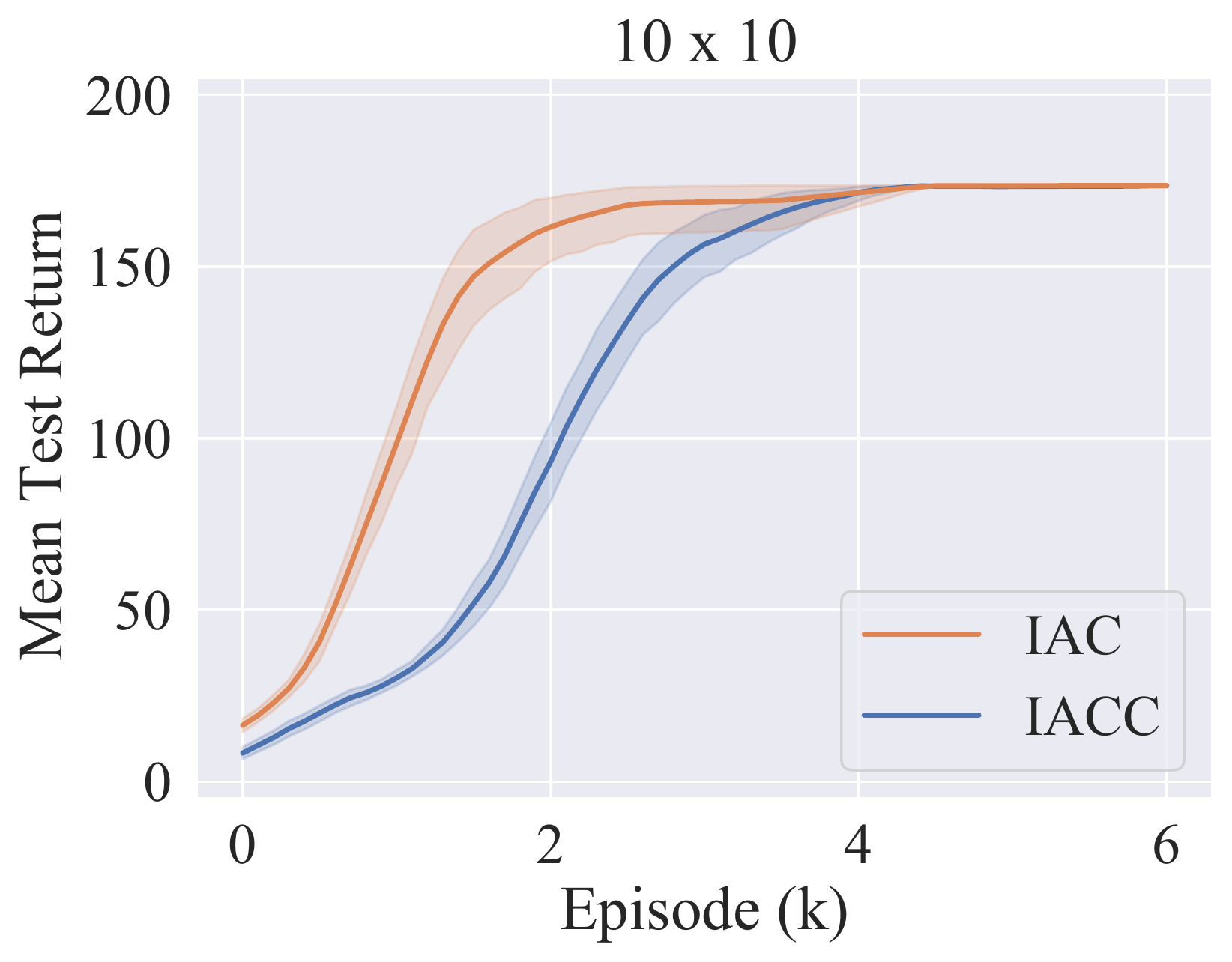}}
  ~
  \centering
  \subcaptionbox{}
      [0.31\linewidth]{\includegraphics[height=3.5cm]{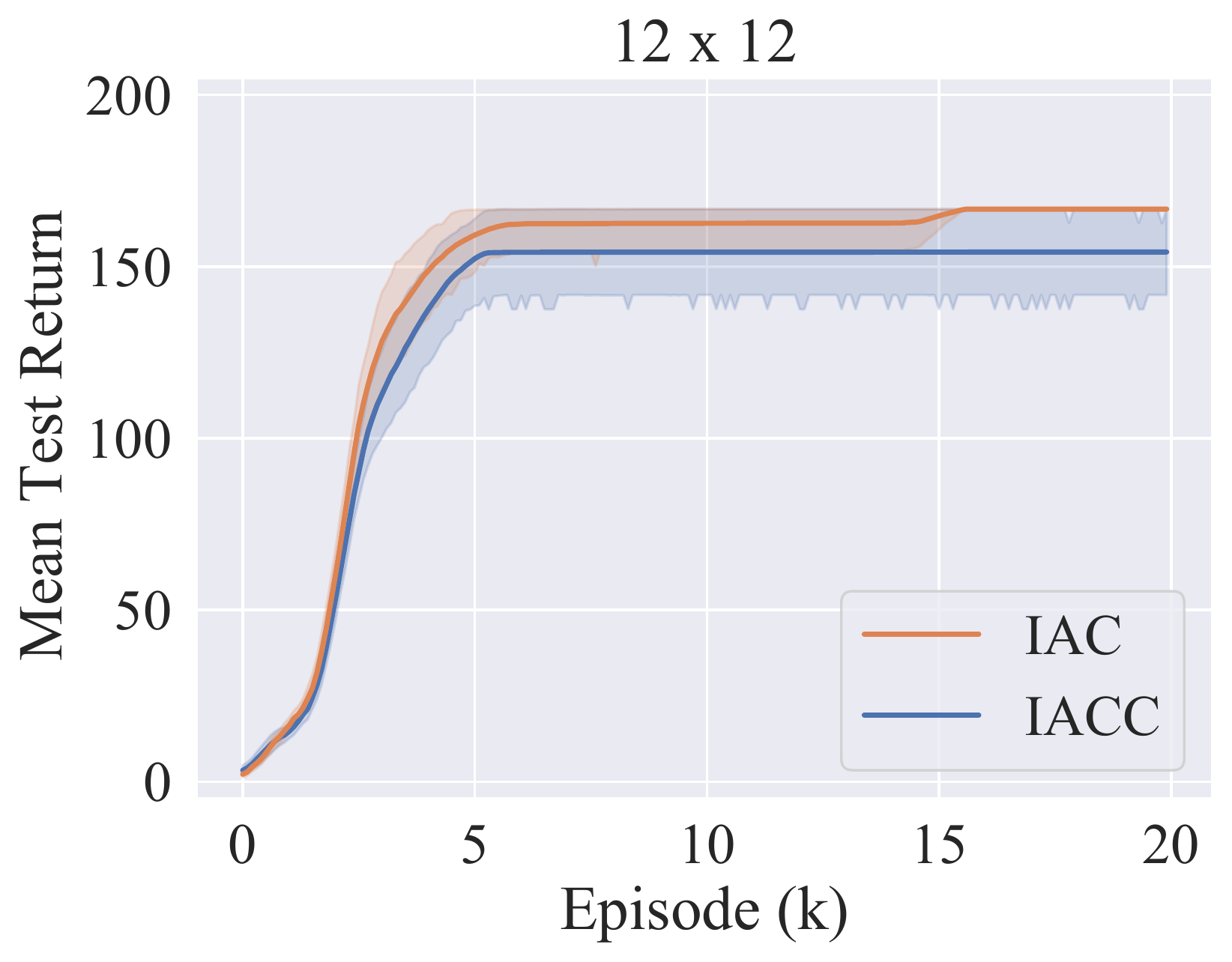}}

  \caption{Performance comparison on Small Box Pushing under various grid world sizes.}
  \label{SBP2}
\end{figure}

\clearpage
\section{Small Box Pushing (3 Agents)}\label{small_box_pushing_3_agents_environment}

The domain settings are same as the above one, except one more agent involved to make the task a little more challenging. 

\begin{figure}[ht!]
  \centering
  \captionsetup[subfigure]{labelformat=empty}
  \centering
  \subcaptionbox{8 x 8}
      [0.31\linewidth]{\includegraphics[height=2.5cm]{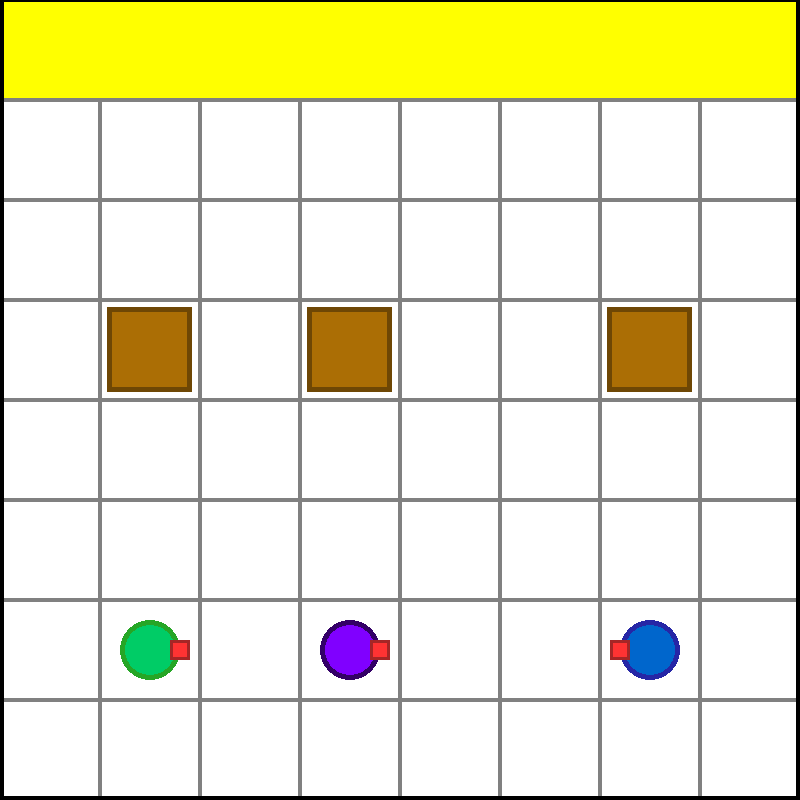}}
  ~
  \centering
  \subcaptionbox{10 x 10}
      [0.31\linewidth]{\includegraphics[height=3.25cm]{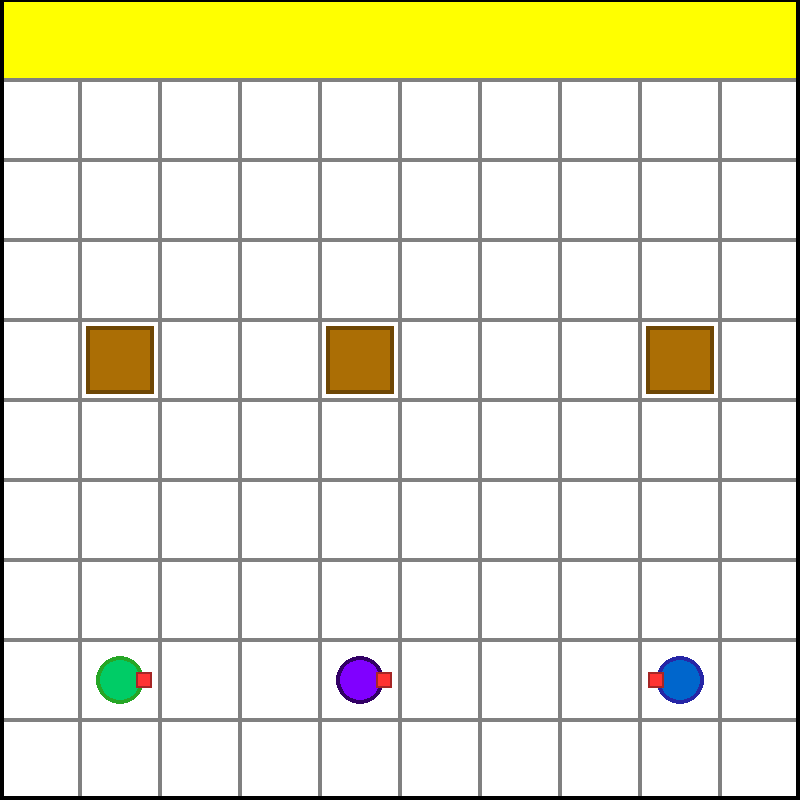}}
  ~
  \centering
  \subcaptionbox{12 x 12}
      [0.31\linewidth]{\includegraphics[height=4cm]{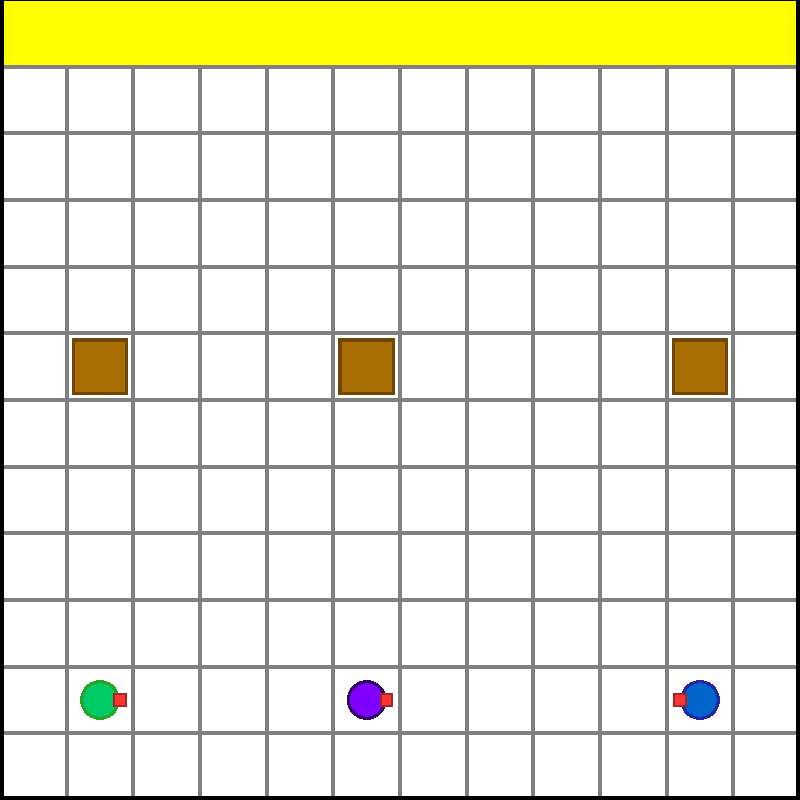}}

  \caption{Domain configurations under various grid world sizes.}
  \label{3bp_config}
\end{figure}

\begin{figure}[ht!]
  \centering
  \captionsetup[subfigure]{labelformat=empty}
  \subcaptionbox{}
      [0.31\linewidth]{\includegraphics[height=3.5cm]{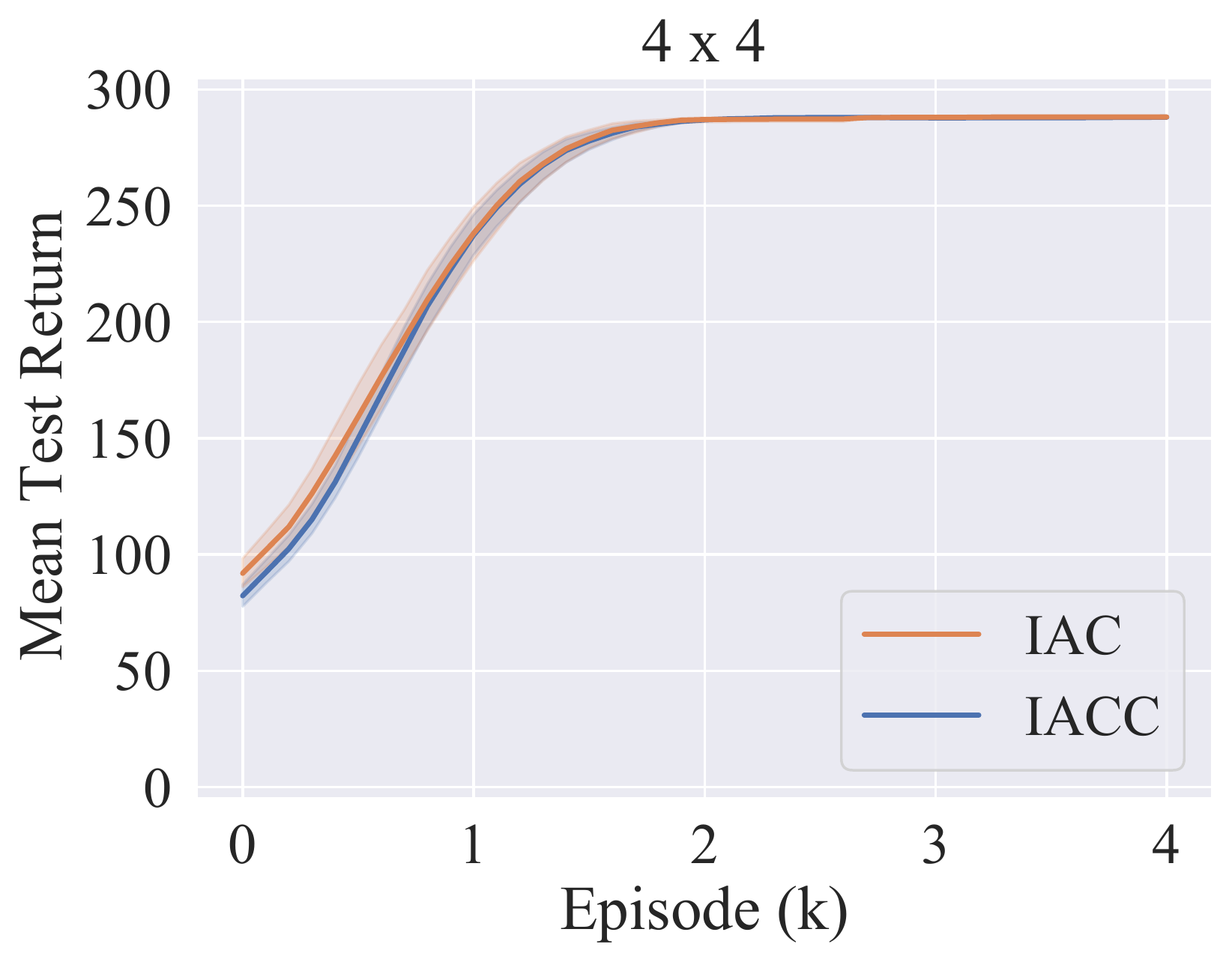}}
  ~
  \centering
  \subcaptionbox{}
      [0.31\linewidth]{\includegraphics[height=3.5cm]{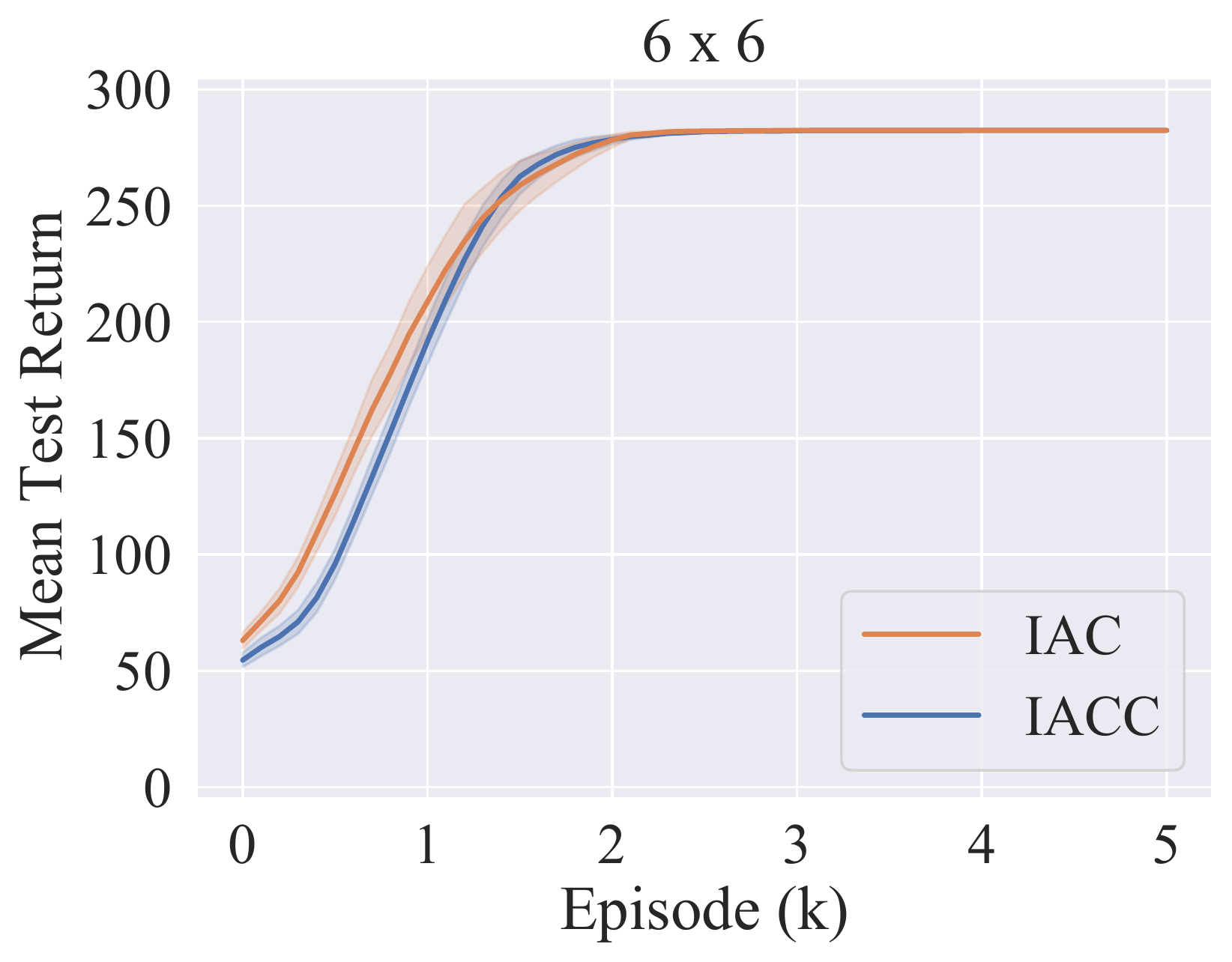}}
  ~
  \centering
  \subcaptionbox{}
      [0.31\linewidth]{\includegraphics[height=3.5cm]{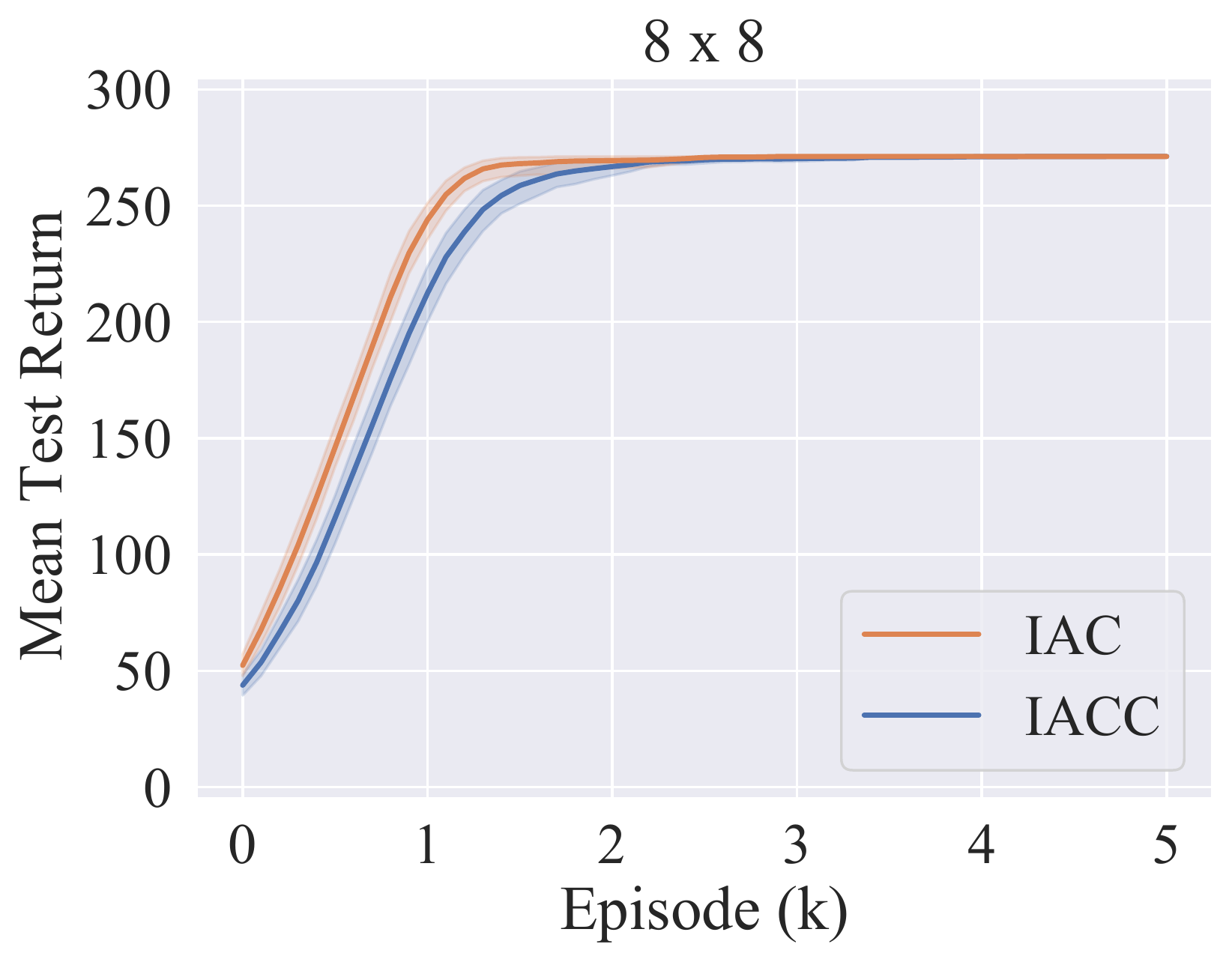}}
   
  \centering
  \subcaptionbox{}
      [0.31\linewidth]{\includegraphics[height=3.5cm]{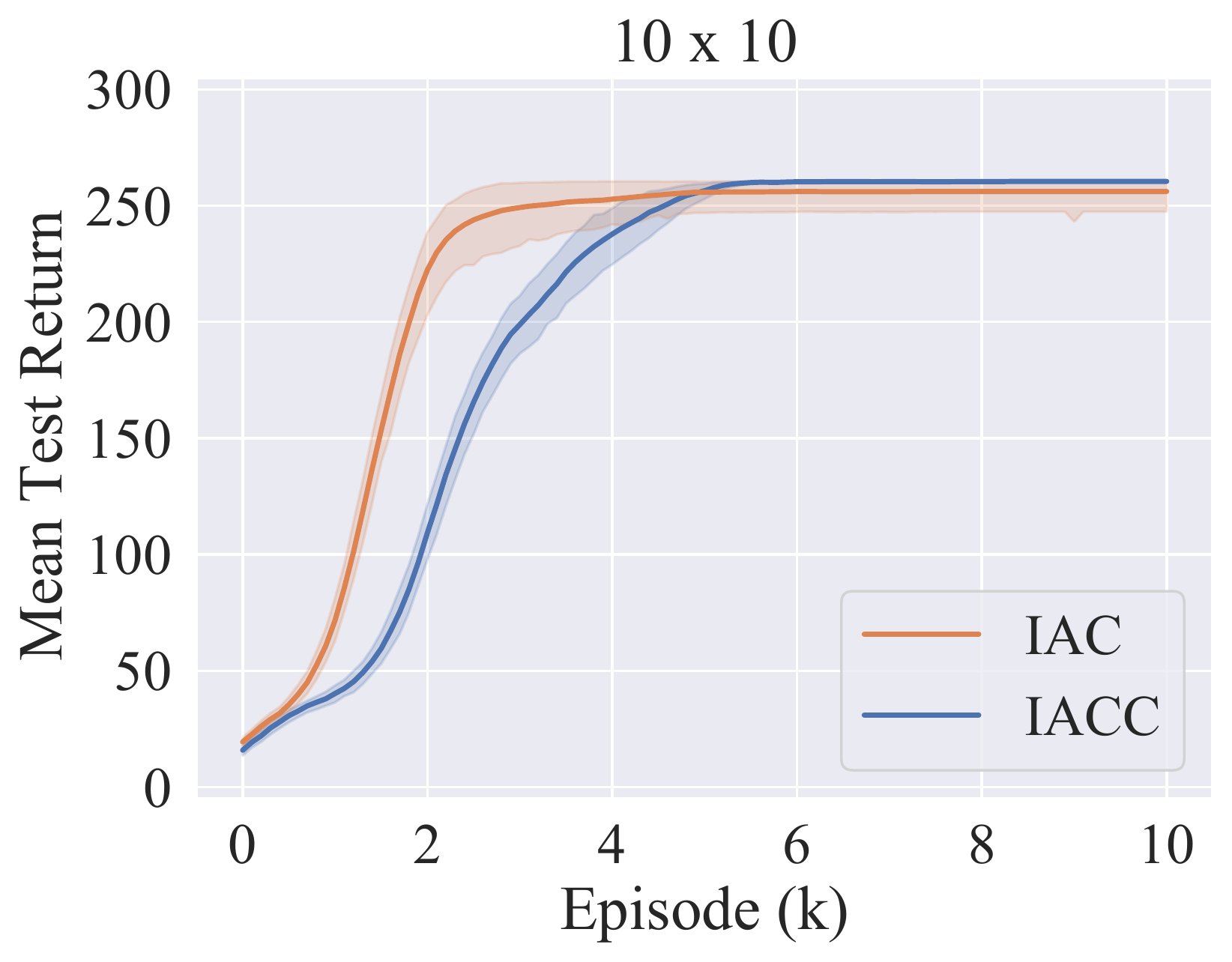}}
  ~
  \centering
  \subcaptionbox{}
      [0.31\linewidth]{\includegraphics[height=3.5cm]{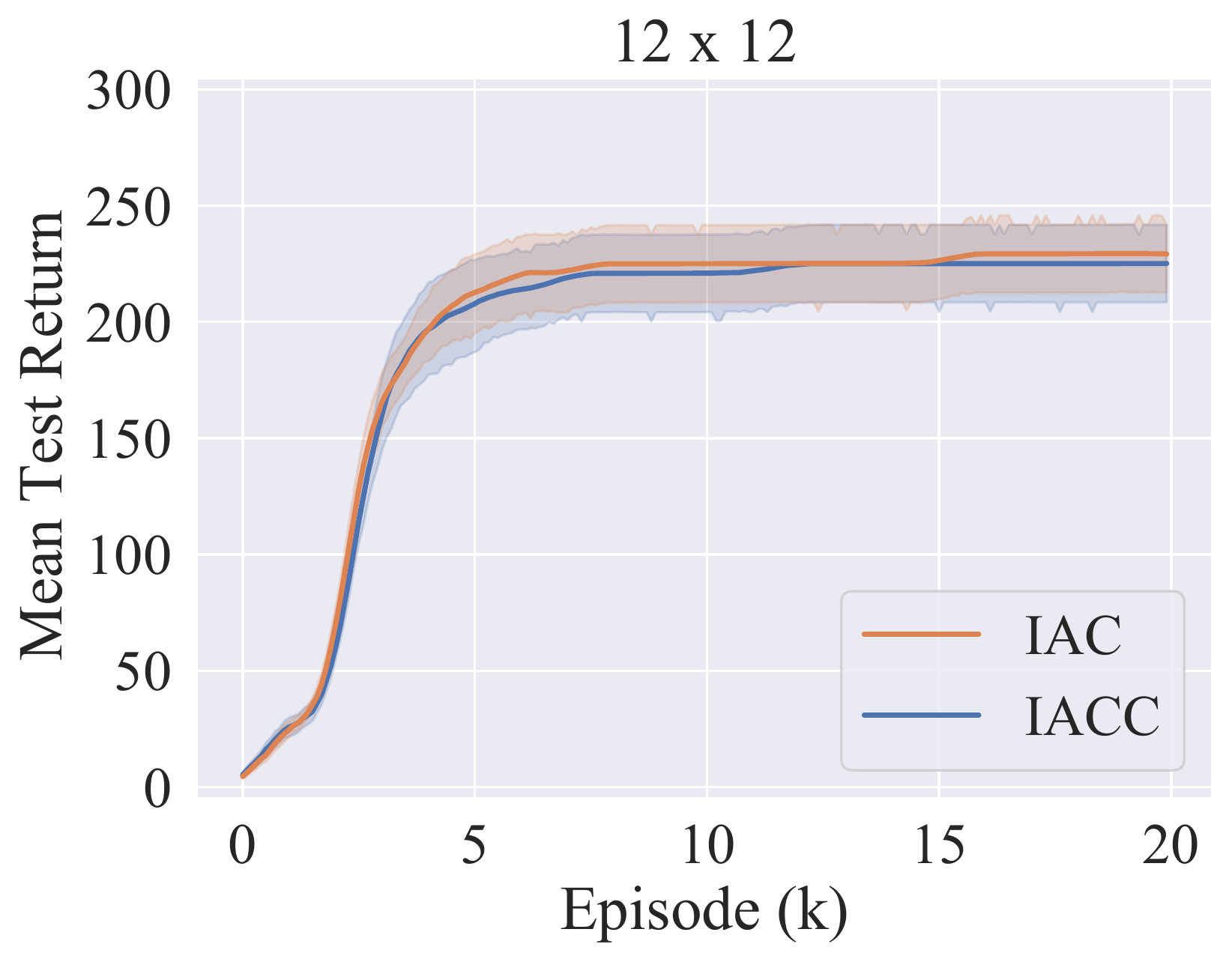}}

  \caption{Performance comparison on three-agent Small Box Pushing under various grid world sizes.}
  \label{3A_SBP2}
\end{figure}

\clearpage
\section{SMAC}\label{smac_envionrment}

We also investigated the performance of IAC and IACC on StarCraft II micromanagement tasks from the StarCraft Multi-Agent Challenge (SMAC) \cite{samvelyan19smac}.
We picked out three classical scenarios:
2 Stalkers vs 1 Spine Crawler (2s\_vs\_1sc), 3 Marines (3m) and 2 Stalkers and 3 Zealots (2s3z).
We used the default configurations recommended in SMAC.
Readers are referred to the source code\footnote{
    https://github.com/oxwhirl/smac
}
for details. 
We use simpler tasks in SMAC due to harder tasks poses a huge challenge as both method in their vanilla formulation .
\begin{figure}[ht!]

  \centering
  \captionsetup[subfigure]{labelformat=empty}
  \subcaptionbox{}
      [0.31\linewidth]{\includegraphics[height=3.5cm]{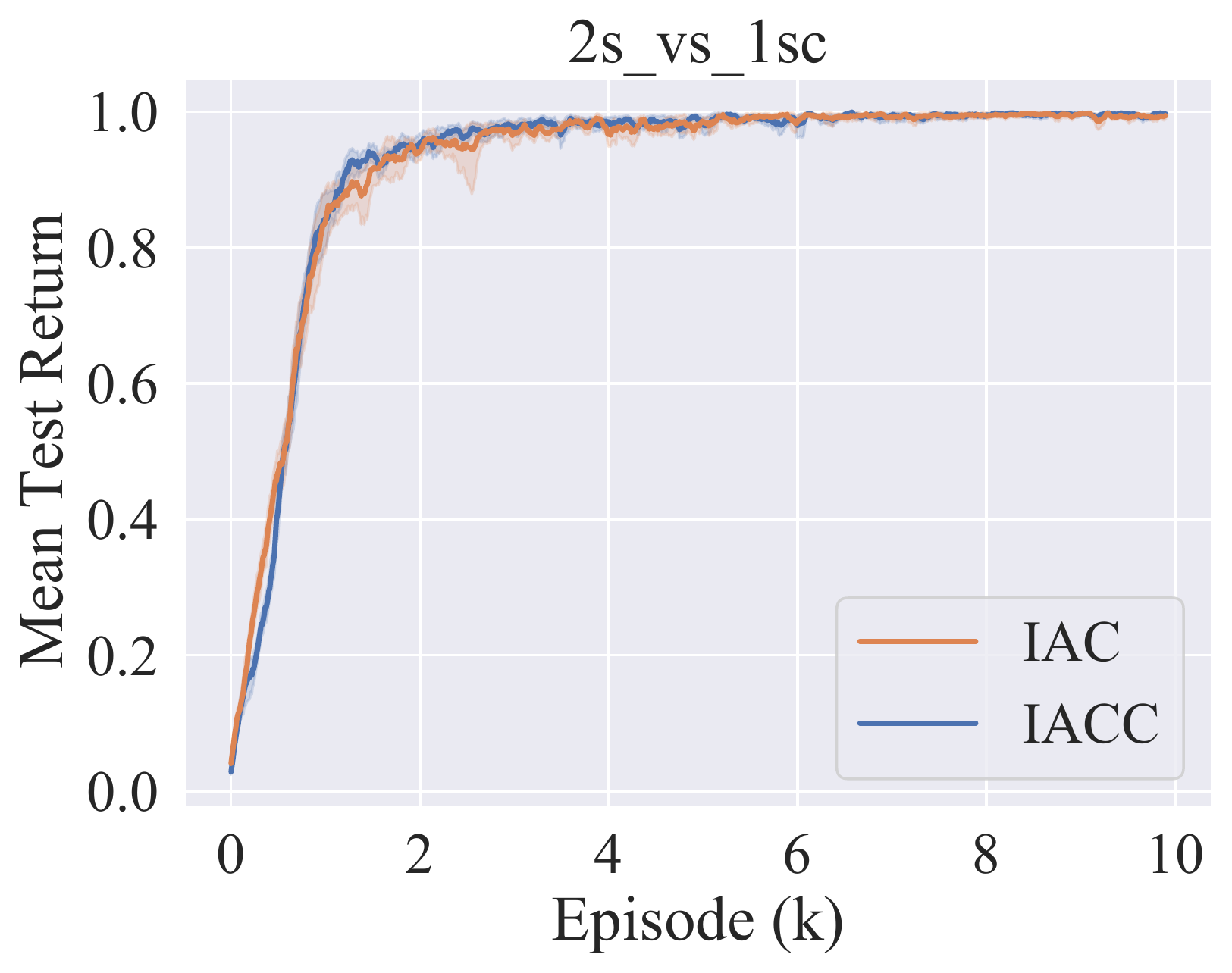}}
  ~
  \centering
  \subcaptionbox{}
      [0.31\linewidth]{\includegraphics[height=3.5cm]{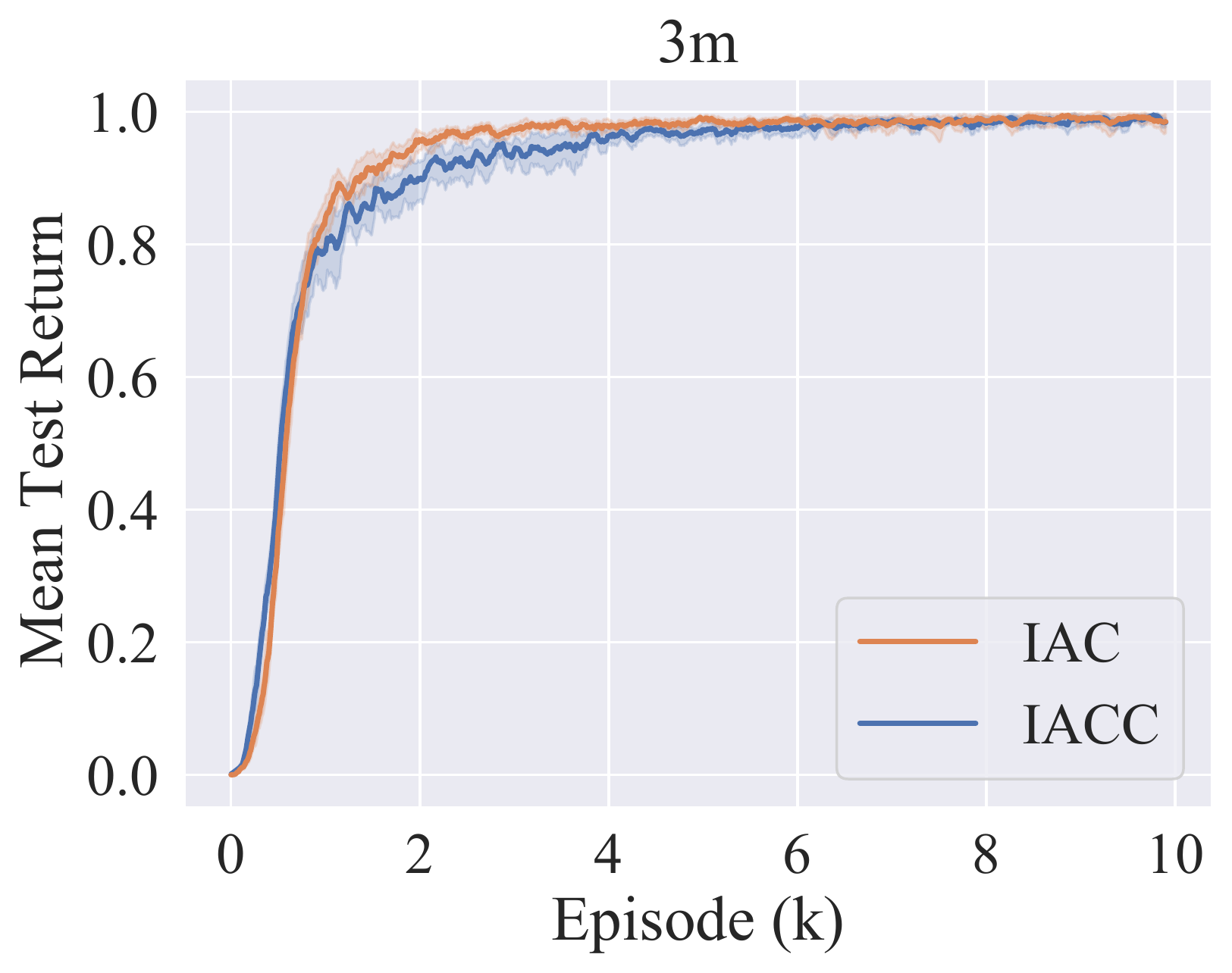}}
  ~
  \centering
  \subcaptionbox{}
      [0.31\linewidth]{\includegraphics[height=3.5cm]{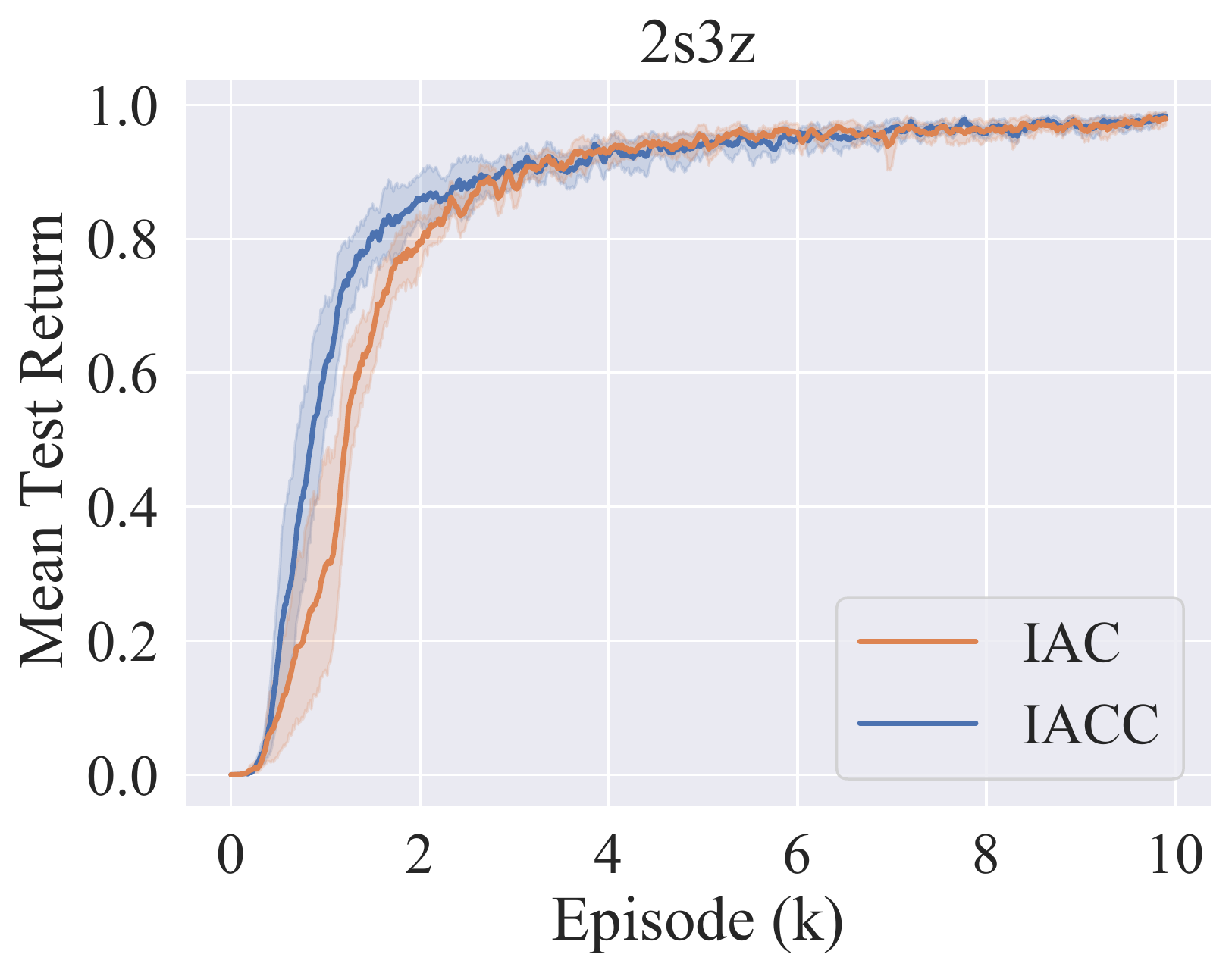}}
  \caption{Performance comparison on easy SMAC maps.}
  \label{smac_easy}
\end{figure}

\end{document}

%% file: math_commands.tex
\usepackage{amsmath,amsfonts,bm}

\def\eqref#1{equation~\ref{#1}}

\def\1{\bm{1}}

\DeclareMathAlphabet{\mathsfit}{\encodingdefault}{\sfdefault}{m}{sl}
\SetMathAlphabet{\mathsfit}{bold}{\encodingdefault}{\sfdefault}{bx}{n}

\newcommand{\E}{\mathbb{E}}

\newcommand{\Var}{\mathrm{Var}}

